%% file: main.tex
\documentclass[acmtog]{acmart}
\acmSubmissionID{790}

\usepackage{booktabs} %

\makeatletter
\newcommand{\removelatexerror}{\let\@latex@error\@gobble}
\makeatother

\usepackage[ruled]{algorithm2e} %

\SetAlFnt{\small}
\SetAlCapFnt{\small}
\SetAlCapNameFnt{\small}
\SetAlCapHSkip{0pt}

\acmJournal{TOG}
\acmVolume{1}
\acmNumber{1}
\acmArticle{1}
\acmYear{2023}
\acmMonth{1}
\acmPrice{}

\setcopyright{rightsretained}

\acmDOI{10.1145/3592450}

\usepackage{algorithmic}
\usepackage{cleveref}
\usepackage[export]{adjustbox}
\usepackage{rotating}
\usepackage{wrapfig}
\usepackage{float}
\usepackage{array}
\usepackage{subfig}
\usepackage{tabularx}
\usepackage{tikz}
\usetikzlibrary{spy}
\usepackage{enumitem}
\usepackage{colortbl}
\usepackage{gensymb}

\newlength{\ww}

\newcommand\blfootnote[1]{%
  \begingroup
  \renewcommand\thefootnote{}\footnote{#1}%
  \addtocounter{footnote}{-1}%
  \endgroup
}

\begin{document}
\title{Blended Latent Diffusion}

\author{Omri Avrahami}
\orcid{0000-0002-7628-7525}
\affiliation{
 \institution{The Hebrew University of Jerusalem}
 \city{Jerusalem}
 \country{Israel}}
\email{omri.avrahami@mail.huji.ac.il}

\author{Ohad Fried}
\orcid{0000-0001-7109-4006}
\affiliation{
 \institution{Reichman University}
 \city{Herzliya}
 \country{Israel}}
\email{ofried@idc.ac.il.}

\author{Dani Lischinski}
\orcid{0000-0002-6191-0361}
\affiliation{
 \institution{The Hebrew University of Jerusalem}
 \city{Jerusalem}
 \country{Israel}}
\email{danix@mail.huji.ac.il}

\renewcommand\shortauthors{Avrahami et al.}

\def\ShowNotes{}
\input{macros.tex}

\input{sections/abstract.tex}

  

\input{figures/teaser/fig.tex}

\maketitle

\blfootnote{Project page is available at: \href{https://omriavrahami.com/blended-latent-diffusion-page/}{https://omriavrahami.com/blended-latent-diffusion-page/}}
\input{sections/introduction.tex}
\input{sections/related_work.tex}
\input{sections/preliminaries.tex}
\input{sections/method.tex}

\input{sections/results.tex}
\input{sections/limitations.tex}

\begin{acks}
    This work was supported in part by the Israel Science Foundation (grants No. 1574/21, 2492/20, and 3611/21).
\end{acks}

\begin{appendix}
\input{appendices/additional_examples.tex}
\input{appendices/additional_comparisons.tex}
\input{appendices/implementation_details.tex}
\input{appendices/sensitivity_analysis.tex}
\input{sections/societal_impact.tex}
\end{appendix}

\clearpage
\bibliographystyle{ACM-Reference-Format}
\bibliography{egbib}

\end{document}

%% file: macros.tex
\newcommand{\ignorethis}[1]{}
\newcommand{\redund}[1]{#1}

\newcommand{\etal       }     {{et~al.}}
\newcommand{\apriori    }     {\textit{a~priori}}
\newcommand{\aposteriori}     {\textit{a~posteriori}}
\newcommand{\perse      }     {\textit{per~se}}
\newcommand{\eg         }     {{e.g.}}
\newcommand{\Eg         }     {{E.g.}}
\newcommand{\ie         }     {{i.e.}}
\newcommand{\naive      }     {{na\"{\i}ve}}

\newcommand{\Identity   }     {\mat{I}}
\newcommand{\Zero       }     {\mathbf{0}}
\newcommand{\Reals      }     {{\textrm{I\kern-0.18em R}}}
\newcommand{\isdefined  }     {\mbox{\hspace{0.5ex}:=\hspace{0.5ex}}}
\newcommand{\texthalf   }     {\ensuremath{\textstyle\frac{1}{2}}}
\newcommand{\half       }     {\ensuremath{\frac{1}{2}}}
\newcommand{\third      }     {\ensuremath{\frac{1}{3}}}
\newcommand{\fourth     }     {\ensuremath{\frac{1}{4}}}

\newcommand{\Lone} {\ensuremath{L_1}}
\newcommand{\Ltwo} {\ensuremath{L_2}}

\newcommand{\mat        } [1] {{\text{\boldmath $\mathbit{#1}$}}}
\newcommand{\Approx     } [1] {\widetilde{#1}}
\newcommand{\change     } [1] {\mbox{{\footnotesize $\Delta$} \kern-3pt}#1}

\newcommand{\Order      } [1] {O(#1)}
\newcommand{\set        } [1] {{\lbrace #1 \rbrace}}
\newcommand{\floor      } [1] {{\lfloor #1 \rfloor}}
\newcommand{\ceil       } [1] {{\lceil  #1 \rceil }}
\newcommand{\inverse    } [1] {{#1}^{-1}}
\newcommand{\transpose  } [1] {{#1}^\mathrm{T}}
\newcommand{\invtransp  } [1] {{#1}^{-\mathrm{T}}}
\newcommand{\relu       } [1] {{\lbrack #1 \rbrack_+}}

\newcommand{\abs        } [1] {{| #1 |}}
\newcommand{\Abs        } [1] {{\left| #1 \right|}}
\newcommand{\norm       } [1] {{\| #1 \|}}
\newcommand{\Norm       } [1] {{\left\| #1 \right\|}}
\newcommand{\pnorm      } [2] {\norm{#1}_{#2}}
\newcommand{\Pnorm      } [2] {\Norm{#1}_{#2}}
\newcommand{\inner      } [2] {{\langle {#1} \, | \, {#2} \rangle}}
\newcommand{\Inner      } [2] {{\left\langle \begin{array}{@{}c|c@{}}
                               \displaystyle {#1} & \displaystyle {#2}
                               \end{array} \right\rangle}}

\newcommand{\twopartdef}[4]
{
  \left\{
  \begin{array}{ll}
    #1 & \mbox{if } #2 \\
    #3 & \mbox{if } #4
  \end{array}
  \right.
}

\newcommand{\fourpartdef}[8]
{
  \left\{
  \begin{array}{ll}
    #1 & \mbox{if } #2 \\
    #3 & \mbox{if } #4 \\
    #5 & \mbox{if } #6 \\
    #7 & \mbox{if } #8
  \end{array}
  \right.
}

\newcommand{\len}[1]{\text{len}(#1)}

\newlength{\w}
\newlength{\h}
\newlength{\x}

\definecolor{darkred}{rgb}{0.7,0.1,0.1}
\definecolor{darkgreen}{rgb}{0.1,0.6,0.1}
\definecolor{cyan}{rgb}{0.7,0.0,0.7}
\definecolor{otherblue}{rgb}{0.1,0.4,0.8}
\definecolor{maroon}{rgb}{0.76,.13,.28}
\definecolor{burntorange}{rgb}{0.81,.33,0}

\ifdefined\ShowNotes
  \newcommand{\colornote}[3]{{\color{#1}\textbf{#2} #3\normalfont}}
\else
  \newcommand{\colornote}[3]{}
\fi

\newcommand {\todo}[1]{\colornote{cyan}{TODO}{#1}}
\newcommand {\ohad}[1]{\colornote{otherblue}{OF:}{#1}}
\newcommand {\dani}[1]{\colornote{darkgreen}{DL:}{#1}}
\newcommand {\omri}[1]{\colornote{burntorange}{OA:}{#1}}

\newcommand {\reqs}[1]{\colornote{red}{\tiny #1}}

\newcommand {\new}[1]{{\color{red}{#1}}}

\newcommand*\rot[1]{\rotatebox{90}{#1}}

\newcommand {\newstuff}[1]{#1}

\newcommand\todosilent[1]{}

\newcommand{\woBGmask}{{w/o~bg~\&~mask}}
\newcommand{\woMask}{{w/o~mask}}

\providecommand{\keywords}[1]
{
  \textbf{\textit{Keywords---}} #1
}

\newcommand{\GAN}{\textit{GAN}}
\newcommand{\data}{\mathit{data}}
\newcommand{\unionGAN}{\textsc{UnionGAN}\xspace}
\newcommand {\ganArrow}[2]{\ensuremath{\GAN_{{#1} \rightarrow {#2}}}}
\newcommand {\gan}[1]{\ensuremath{\GAN_{#1}}}
\newcommand{\DALLE}{{DALL$\cdot$E}}

%% file: sections/abstract.tex
\begin{abstract}
	The tremendous progress in neural image generation, coupled with the emergence of seemingly omnipotent vision-language models has finally enabled text-based interfaces for creating and editing images. Handling \emph{generic} images requires a diverse underlying generative model, hence the latest works utilize diffusion models, which were shown to surpass GANs in terms of diversity. One major drawback of diffusion models, however, is their relatively slow inference time. In this paper, we present an accelerated solution to the task of \emph{local} text-driven editing of generic images, where the desired edits are confined to a user-provided mask. Our solution leverages a text-to-image Latent Diffusion Model (LDM), which speeds up diffusion by operating in a lower-dimensional latent space and eliminating the need for resource-intensive CLIP gradient calculations at each diffusion step. We first enable LDM to perform local image edits by blending the latents at each step, similarly to Blended Diffusion. Next we propose an optimization-based solution for the inherent inability of LDM to accurately reconstruct images. Finally, we address the scenario of performing local edits using thin masks. We evaluate our method against the available baselines both qualitatively and quantitatively and demonstrate that in addition to being faster, it produces more precise results.
\end{abstract}

%% file: figures/teaser/fig.tex
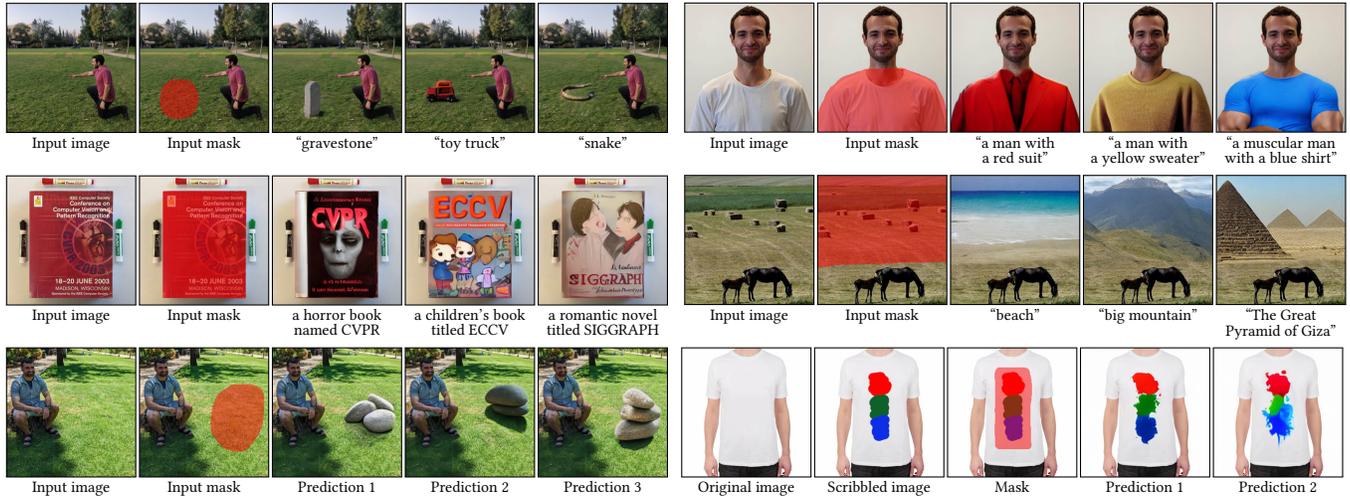
\begin{teaserfigure}
  \setlength{\tabcolsep}{-2.5pt}
  \begin{tabular}{cc}
    \input{figures/teaser/new_object.tex} & 
    \input{figures/teaser/alter_object.tex}
    \\[1cm]
    \input{figures/teaser/text_edit.tex} & 
    \input{figures/teaser/background_edit.tex}
    \\[1cm]
    \input{figures/teaser/multiple_predictions.tex} & 
    \input{figures/teaser/scribble_edit.tex}
    \\
  \end{tabular}
  \captionof{figure}{\textbf{Applications of our method:} (top left) adding a new object in a masked area, guided by a text prompt; (top right) altering a part within an existing object; (middle left) generation of text; (middle right) altering the background in the scene; (bottom left) generating multiple predictions for the same text prompt (``stones''); (bottom right) guiding the result by a combination of text (``paint splashes'') and scribbles.}
  \label{fig:teaser}
\end{teaserfigure}

%% file: figures/teaser/new_object.tex
\setlength{\tabcolsep}{0.5pt}
\renewcommand{\arraystretch}{0.5}
\setlength{\ww}{0.095\columnwidth}

\begin{tabular}{ccccc}
    \includegraphics[width=\ww,frame]{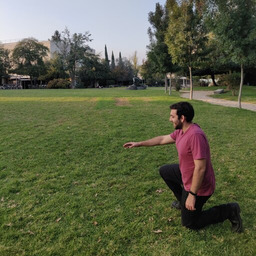} &
    \includegraphics[width=\ww,frame]{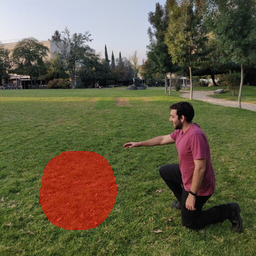} &
    \includegraphics[width=\ww,frame]{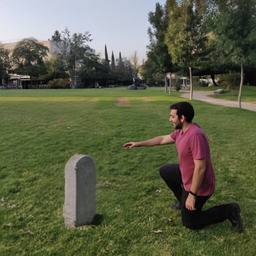} &
    \includegraphics[width=\ww,frame]{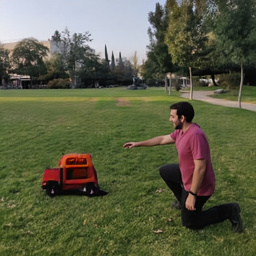} &
    \includegraphics[width=\ww,frame]{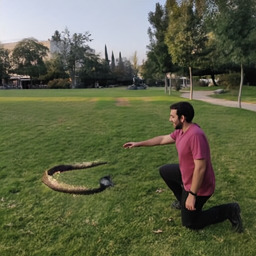}  \\

    \scriptsize{Input image} &
    \scriptsize{Input mask} &
    \scriptsize{``gravestone''} &
    \scriptsize{``toy truck''} &
    \scriptsize{``snake''}
    \\
    \\
    
\end{tabular}

%% file: figures/teaser/alter_object.tex
\setlength{\tabcolsep}{0.5pt}
\renewcommand{\arraystretch}{0.5}
\setlength{\ww}{0.095\columnwidth}

\begin{tabular}{ccccc}
    \includegraphics[width=\ww,frame]{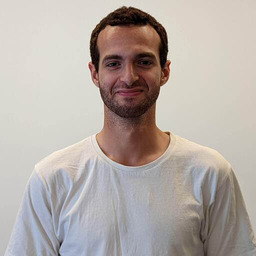} &
    \includegraphics[width=\ww,frame]{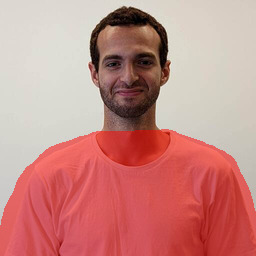} &
    \includegraphics[width=\ww,frame]{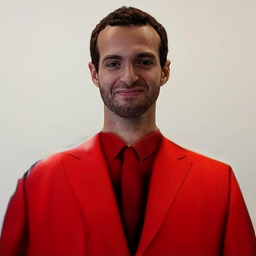} &
    \includegraphics[width=\ww,frame]{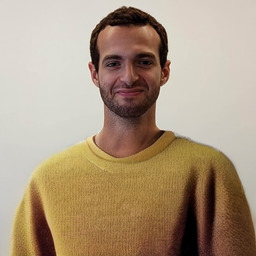} &
    \includegraphics[width=\ww,frame]{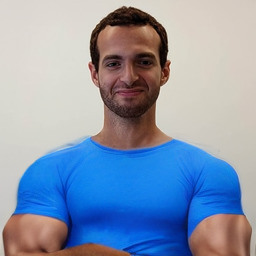}  \\

    \scriptsize{Input image} &
    \scriptsize{Input mask} &
    \scriptsize{``a man with} &
    \scriptsize{``a man with} &
    \scriptsize{``a muscular man}
    \\

    &&
    \scriptsize{a red suit''} &
    \scriptsize{a yellow sweater''} &
    \scriptsize{with a blue shirt''}
\end{tabular}

%% file: figures/teaser/text_edit.tex
\setlength{\tabcolsep}{0.5pt}
\renewcommand{\arraystretch}{0.5}
\setlength{\ww}{0.095\columnwidth}

\begin{tabular}{ccccc}
    \includegraphics[width=\ww,frame]{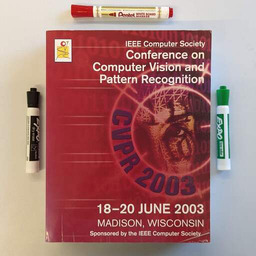} &
    \includegraphics[width=\ww,frame]{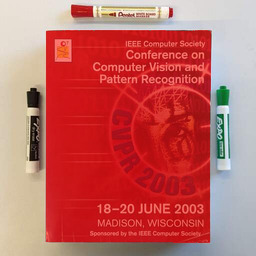} &
    \includegraphics[width=\ww,frame]{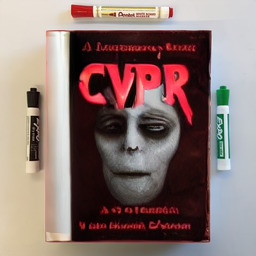} &
    \includegraphics[width=\ww,frame]{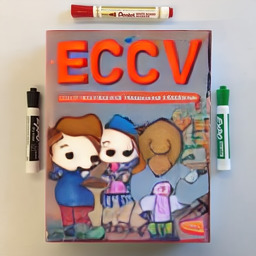} &
    \includegraphics[width=\ww,frame]{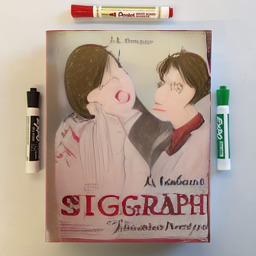}  \\

    \scriptsize{Input image} &
    \scriptsize{Input mask} &
    \scriptsize{a horror book} &
    \scriptsize{a children's book} &
    \scriptsize{a romantic novel}
    \\

    &&
    \scriptsize{named CVPR} &
    \scriptsize{titled ECCV} &
    \scriptsize{titled SIGGRAPH}
    \\
\end{tabular}

%% file: figures/teaser/background_edit.tex
\setlength{\tabcolsep}{0.5pt}
\renewcommand{\arraystretch}{0.5}
\setlength{\ww}{0.095\columnwidth}

\begin{tabular}{ccccc}
    \includegraphics[width=\ww,frame]{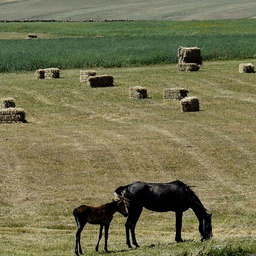} &
    \includegraphics[width=\ww,frame]{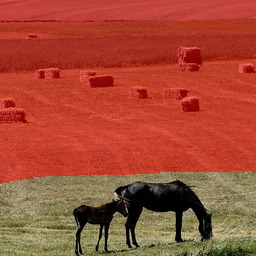} &
    \includegraphics[width=\ww,frame]{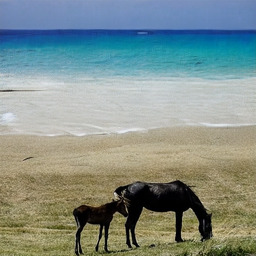} &
    \includegraphics[width=\ww,frame]{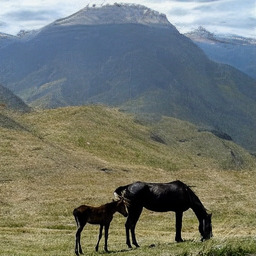} &
    \includegraphics[width=\ww,frame]{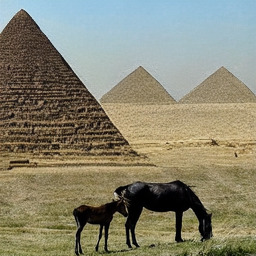}  \\

    \scriptsize{Input image} &
    \scriptsize{Input mask} &
    \scriptsize{``beach''} &
    \scriptsize{``big mountain''} &
    \scriptsize{``The Great}
    \\

    &&&&
    \scriptsize{Pyramid of Giza''}
    \\
\end{tabular}

%% file: figures/teaser/multiple_predictions.tex
\setlength{\tabcolsep}{0.5pt}
\renewcommand{\arraystretch}{0.5}
\setlength{\ww}{0.095\columnwidth}

\begin{tabular}{ccccc}
    \includegraphics[width=\ww,frame]{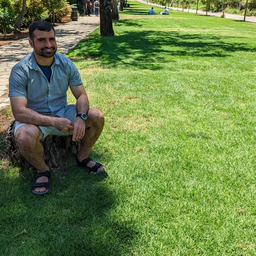} &
    \includegraphics[width=\ww,frame]{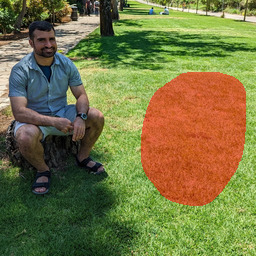} &
    \includegraphics[width=\ww,frame]{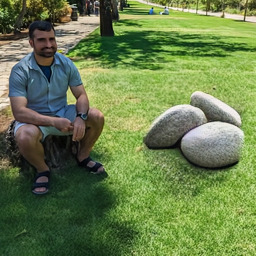} &
    \includegraphics[width=\ww,frame]{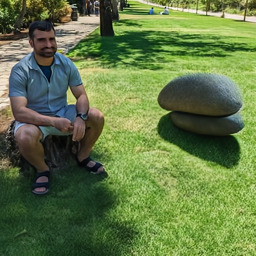} &
    \includegraphics[width=\ww,frame]{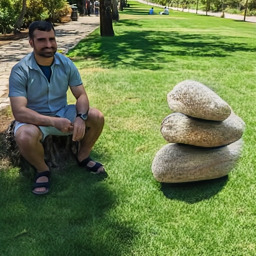}  \\

    \scriptsize{Input image} &
    \scriptsize{Input mask} &
    \scriptsize{Prediction 1} &
    \scriptsize{Prediction 2} &
    \scriptsize{Prediction 3}
    \\
\end{tabular}

%% file: figures/teaser/scribble_edit.tex
\setlength{\tabcolsep}{0.5pt}
\renewcommand{\arraystretch}{0.5}
\setlength{\ww}{0.095\columnwidth}

\begin{tabular}{ccccc}
    \includegraphics[width=\ww,frame]{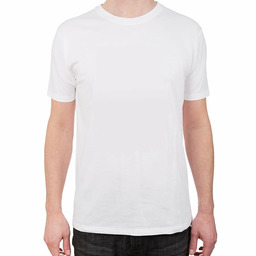} &
    \includegraphics[width=\ww,frame]{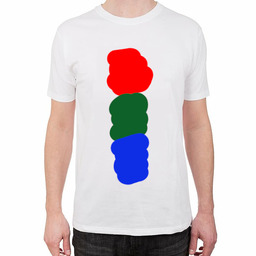} &
    \includegraphics[width=\ww,frame]{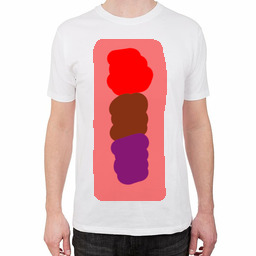} &
    \includegraphics[width=\ww,frame]{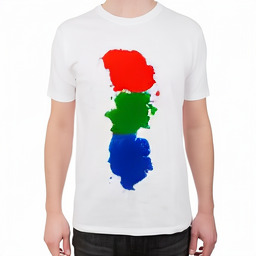} &
    \includegraphics[width=\ww,frame]{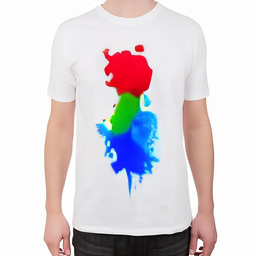}
    \\

    \scriptsize{Original image} &
    \scriptsize{Scribbled image} &
    \scriptsize{Mask} &
    \scriptsize{Prediction 1} &
    \scriptsize{Prediction 2}
    \\
\end{tabular}

%% file: sections/introduction.tex
\section{Introduction}

In recent years we have witnessed tremendous progress in realistic image synthesis and image manipulation with deep neural generative models.
GAN-based models were first to emerge \cite{goodfellow2014gans,brock2018large,karras2019style,karras2020analyzing}, soon followed by diffusion-based models \cite{sohl2015deep, ho2020denoising, nichol2021improved}.
In parallel, recent vision-language models, such as CLIP~\cite{radford2021learning}, have paved the way for generating and editing images using a fundamental form of human communication --- natural language.
The resulting text-guided image generation and manipulation approaches, e.g., \cite{patashnik2021styleclip,nichol2021glide,ramesh2022hierarchical, saharia2022photorealistic, yu2022scaling, ding2021cogview}, enable artists to simply convey their intent in natural language, potentially saving hours of painstaking manual work. \Cref{fig:teaser} shows a few examples.

However, the vast majority of text-guided approaches focus on generating images from scratch or on manipulating existing images \emph{globally}.
The \emph{local} editing scenario, where the artist is only interested in modifying a part of a \emph{generic} image, while preserving the remaining parts, has not received nearly as much attention, despite the ubiquity of this use case in practice.
We know of only three methods to date that \emph{explicitly} address the local editing scenario: Blended Diffusion \cite{avrahami2022blended}, GLIDE \cite{nichol2021glide} and \DALLE~2 \cite{ramesh2022hierarchical}.
Among these, only Blended Diffusion is publicly available in full.

All three local editing approaches above are based on diffusion models~\cite{sohl2015deep,nichol2021improved, ho2020denoising}.
While diffusion models have shown impressive results on generation, editing, and other tasks (\Cref{sec:related}), they suffer from long inference times, due to the iterative diffusion process that is applied at the pixel level to generate each result.
Some recent works \cite{rombach2021high, gu2021vector, esser2021imagebart, bond2021unleashing, hu2021global, vahdat2021score} have thus proposed to perform the diffusion in a latent space with lower dimensions and higher-level semantics, compared to pixels, yielding competitive performance on various tasks with much lower training and inference times. In particular, Latent Diffusion Models (LDM) \cite{rombach2021high} offer this appealing combination of competitive image quality with fast inference, however, this approach targets text-to-image generation from scratch, rather than global image manipulation, let alone local editing.

In this work, we harness the merits of LDM to the task of \emph{local} text-guided natural image editing, where the user provides the image to be edited, a natural language text prompt, and a mask indicating an area to which the edit should be confined. Our approach is ``zero-shot'', since it relies on available pretrained models, and requires no further training.
We first show how to adapt the Blended Diffusion approach of Avrahami \etal~\shortcite{avrahami2022blended} to work in the latent space of LDM, instead of working at the pixel level.

Next, we address the imperfect reconstruction inherent to LDM, due to the use of VAE-based lossy latent encodings.  
This is especially problematic when the original image contains areas to which human perception is particularly sensitive (e.g., faces or text) or other non-random high frequency details.
We present an approach that employs latent optimization to effectively mitigate this issue.

Then, we address the challenge of performing local edits inside thin masks. Such masks are essential when the desired edit is highly localized, but they present a difficulty when working in a latent space with lower spatial resolution.
To overcome this issue, we propose a solution that starts with a dilated mask, and gradually shrinks it as the diffusion process progresses. 

Finally, we evaluate our method against the baselines both qualitatively and quantitatively, using new metrics for text-driven editing methods that we propose: precision and diversity. We demonstrate that our method is not only faster than the baselines, but also achieves better precision.

In summary, the main contribution of this paper are: (1) Adapting the text-to-image LDM to the task of local text-guided image editing. (2) Addressing the inherent problem of inaccurate reconstruction in LDM, which severely limits the applicability of this method. (3) Addressing the case when the method is fed by a thin mask, based on our investigation of the diffusion dynamics. (4) Proposing new evaluation metrics for quantitative comparisons between local text-driven editing methods.

%% file: sections/related_work.tex
\section{Related Work}
\label{sec:related}

 \textbf{Text-to-image synthesis and global editing:} Text-to-image synthesis has advanced tremendously in recent years. Seminal works based on RNNs \cite{Mansimov2016GeneratingIF} and GANs \cite{reed2016generative, zhang2017stackgan, zhang2018stackgan++, xu2018attngan}, were later superseded by transformer-based  approaches \cite{vaswani2017attention}. \DALLE~\cite{ramesh2021zero} proposed a two-stage approach: first, train a discrete VAE \cite{Oord2017NeuralDR, razavi2019generating} to learn a rich semantic context, then train a transformer model to autoregressively model the joint distribution over the text and image tokens.

Another line of works is based on CLIP \cite{radford2021learning}, a vision-language model that learns a rich shared embedding space for images and text, by contrastive training on a dataset of 400 million (image, text) pairs collected from the internet. Some of them \cite{patashnik2021styleclip,crowson2022vqgan,clip_guided_diffusion, liu2021more, big_sleep, paiss2022no} combine a pretrained generative model \cite{brock2018large, esser2021taming, dhariwal2021diffusion} with a CLIP model to steer the generative model to perform text-to-image synthesis. Utilizing CLIP along with a generative model was also used for text-based domain adaptation \cite{gal2021stylegan} and text-to-image without training on text data \cite{zhou2021lafite, wang2022clip, ashual2022knn}. Make-a-scene \cite{gafni2022make} first predicts the segmentation mask, conditioned on the text, and then uses the generated mask along with the text to generate the predicted image. SpaText \cite{avrahami2022spatext} extends Make-a-scene to support free-form text prompt per segment. These works do not address our setting of \emph{local} text-guided image editing.

Diffusion models were also used for various global image-editing applications: ILVR \cite{choi2021ilvr} demonstrates how to condition a DDPM model on an input image for image translation tasks. Palette \cite{saharia2022palette} trains a designated diffusion model to perform four image-to-image translation tasks, namely colorization, inpainting, uncropping, and JPEG restoration. SDEdit \cite{meng2021sdedit} demonstrates stroke painting to image, image compositing, and stroke-based editing. RePaint 
\cite{lugmayr2022repaint} uses a diffusion model for free-form inpainting of images. None of the above methods tackle the problem of local text-driven image editing.

\textbf{Local text-guided image manipulation:} Paint By Word \cite{bau2021paint} was first to address the problem of zero-shot local text-guided image manipulation by combining BigGAN / StyleGAN with CLIP and editing only the part of the feature map that corresponds to the input mask. However, this method only operated on generated images as input, and used a separate generative model per input domain.
Later, Blended Diffusion \cite{avrahami2022blended} was proposed as the first solution for \emph{local} text-guided editing of real \emph{generic} images; this approach is further described in \Cref{sec:preliminaries}.

Text2LIVE \cite{bar2022text2live} enables editing the appearance of an object within an image, without relying on a pretrained generative model. They mainly focus on changing the colors/textures of an existing object or adding effects such as fire/smoke, and not on editing a general scene by removing objects or replacing them with new ones, as we do.

More related to our work are the recent GLIDE \cite{nichol2021glide} and \DALLE~2 \cite{ramesh2022hierarchical} works. GLIDE employs a two-stage diffusion-based approach for text-to-image synthesis: the first stage generates a low-resolution version of the image, while the second stage generates a higher resolution version of the image, conditioned on both the low-resolution version and the guiding text. In addition, they fine-tune their model specifically for the task of local editing by a guiding text prompt. Currently, only GLIDE-filtered, a smaller version of their model (300M parameters instead of 5B), which was trained on a smaller filtered dataset, has been released. As we demonstrate in \Cref{sec:results}, GLIDE-filtered often fails to obtain the desired edits.
\DALLE~2 performs text-to-image synthesis by mapping text prompts into CLIP image embeddings, followed by decoding such embeddings to images.
The \DALLE~2 website \cite{dalle2_website} shows some examples of local text-guided image editing; however, this is not discussed in the paper \cite{ramesh2022hierarchical}. Furthermore, neither of their two models has been released.  
The only available resource is their online demo \cite{dalle2_website} that is free for a small number of images, which we use for comparisons (\Cref{fig:baselines_comparison}).

The concurrent prompt-to-prompt work~\cite{hertz2022prompt} enables editing of \emph{generated} images without input masks, given a source text prompt and a target text prompt. In contrast, our method enables editing \emph{real} images, given only a target prompt and a mask.

In summary, at the time of this writing, the only publicly available models that address our setting are Blended Diffusion and GLIDE-filtered.

%% file: sections/preliminaries.tex
\input{figures/noise_artifact_comparison/fig.tex}

\section{Latent Diffusion and Blended Diffusion}
\label{sec:preliminaries}

Diffusion models are deep generative models that sample from the desired distribution by learning to reverse a gradual noising process. Starting from a standard normal distribution noise $x_T$, a series of less-noisy latents, $x_{T-1}, ..., x_0$, are produced. For more details, please refer to \cite{ho2020denoising,nichol2021improved}.

Traditional diffusion models operate directly in the pixel space, hence their optimization often consumes hundreds of GPU days and their inference times are long. To enable faster training and inference on limited computational resources, Rombach \etal~\shortcite{rombach2021high} proposed Latent Diffusion Models (LDMs). They first perform perceptual image compression, using an autoencoder (VAE \cite{kingma2013auto} or VQ-VAE \cite{razavi2019generating,van2017neural, esser2021taming}). Next, a diffusion model is used that operates on the lower-dimensional latent space. They also demonstrate the ability to train a conditional diffusion model on various modalities (e.g., semantic maps, images, or texts), s.t. when they combine it with the autoencoder they create image-to-image / semantic-map-to-image / text-to-image transitions.

Blended Diffusion~\cite{avrahami2022blended} addresses zero-shot text-guided local image editing. This approach utilizes a diffusion model trained on ImageNet \cite{deng2009imagenet}, which serves as a prior for the manifold of the natural images, and a CLIP model \cite{radford2021learning}, which navigates the diffusion model towards the desired text-specified outcome. In order to create a seamless result where only the masked region is modified to comply with the guiding text prompt, each of the noisy images progressively generated by the CLIP-guided process is spatially blended with the \emph{corresponding noisy version} of the input image. The main limitations of this method is its slow inference time (about 25 minutes using a GPU) and its pixel-level noise artifacts (see \Cref{fig:noise_artifact_comparison}).

In the next section, we leverage the trained LDM text-to-image model of Rombach \etal~\shortcite{rombach2021high} to offer a solution for zero-shot text-guided local image editing by incorporating the idea of blending the diffusion latents from Avrahami~\etal~\shortcite{avrahami2022blended} into the LDM latent space (\Cref{sec:incorporating_blended_diffusion}) and mitigating the artifacts inherent to working in that space (\Cref{sec:handling_innacurate_reconstruction,sec:handling_thin_masks}).

%% file: figures/noise_artifact_comparison/fig.tex
\begin{figure}[t]
    \centering
    \setlength{\tabcolsep}{-2pt}
    \renewcommand{\arraystretch}{0.5}
    \setlength{\ww}{0.24\columnwidth}
  
    \begin{tabular}{cccc}
        \begin{tikzpicture}[spy using outlines={}]
            \node {\includegraphics[width=\ww,frame]{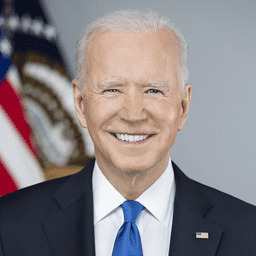}};
        \end{tikzpicture} &

        \begin{tikzpicture}[spy using outlines={}]
            \node {\includegraphics[width=\ww,frame]{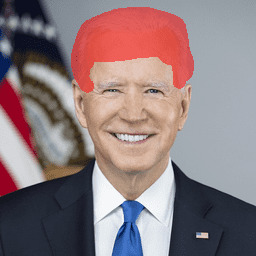}};
        \end{tikzpicture} &

        \begin{tikzpicture}[spy using outlines={circle,red,magnification=2.5,size=0.7cm, connect spies}]
            \node {\includegraphics[width=\ww,frame]{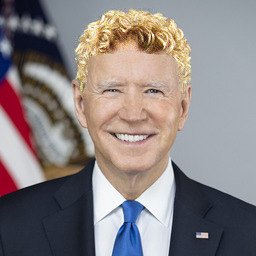}};
            \spy on (0.1,0.7) in node [left] at (0.96,-0.62);
        \end{tikzpicture} &

        \begin{tikzpicture}[spy using outlines={circle,red,magnification=2.5,size=0.7cm, connect spies}]
            \node {\includegraphics[width=\ww,frame]{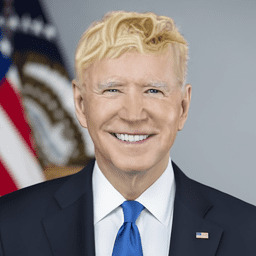}};
            \spy on (0.1,0.7) in node [left] at (0.96,-0.62);
        \end{tikzpicture}
        \\
        
        \scriptsize{(a) Input image} &
        \scriptsize{(b) Input mask} &
        \scriptsize{(c) Blended Diffusion} &
        \scriptsize{(d) Ours}
    \end{tabular}
    \caption{\textbf{Noise artifacts:} Given the input image (a) and mask (b) with the guiding text ``curly blond hair'', Blended Diffusion produces noticeable pixel-level noise artifacts (c), in contrast to our method (d).}
    \label{fig:noise_artifact_comparison}
\end{figure}

%% file: sections/method.tex
\input{figures/method_illustration/fig.tex}
\input{algorithms/blended_latent_diffusion.tex}

\input{figures/reconstruction_comparison/fig.tex}

\section{Method}
\label{sec:method}
Given an image $x$, a guiding text prompt $d$ and a binary mask $m$ that marks the region of interest in the image, our goal is to produce a modified image $\hat{x}$, s.t.~the content $\hat{x} \odot m$ is consistent with the text description $d$, while the complementary area remains close to the source image, i.e., $x \odot (1 - m) \approx \hat{x} \odot (1 - m)$, where $\odot$ is element-wise multiplication. Furthermore, the transition between the two areas of $\hat{x}$ should ideally appear seamless.

In \Cref{sec:incorporating_blended_diffusion} we start by incorporating Blended Diffusion \cite{avrahami2022blended} into Latent Diffusion \cite{rombach2021high} in order to achieve local text-driven editing. The resulting method fails to achieve satisfying results in some cases; specifically, the reconstruction of the complementary area is imperfect, and the method struggles when the input mask $m$ contains thin parts. We solve these issues in \Cref{sec:handling_innacurate_reconstruction,sec:handling_thin_masks}, respectively.

\subsection{Blended Latent Diffusion}
\label{sec:incorporating_blended_diffusion}

As explained in \Cref{sec:preliminaries}, Latent Diffusion \cite{rombach2021high} can generate an image from a given text (text-to-image LDM). However, this model 
lacks the capability of editing an existing image in a local fashion, hence we propose to incorporate Blended Diffusion \cite{avrahami2022blended} into text-to-image LDM.
Our approach is summarized in \Cref{alg:latent_blended_diffusion}, and depicted as a diagram in \Cref{fig:method_illusration}.

LDM performs text-guided denoising diffusion in the latent space learned by a variational auto-encoder $\textit{VAE} = (E(x), D(z))$, where $E(x)$ encodes an image $x$ to a latent representation $z$ and $D(z)$ decodes it back to the pixel space.
Referring to the part that we wish to modify as foreground (\emph{fg}) and to the remaining part as background (\emph{bg}), we follow the idea of Blended Diffusion and repeatedly blend the two parts in this latent space, as the diffusion progresses.
The input image $x$ is encoded into the latent space using the VAE encoder $z_{\textit{init}} \sim E(x)$. The latent space still has spatial dimensions (due to the convolutional nature of the VAE), however the width and the height are smaller than those of the input image (by a factor of 8). We therefore downsample the input mask $m$ to these spatial dimensions to obtain the latent space binary mask $m_{\textit{latent}}$, which will be used to perform the blending.

Now, we noise the initial latent $z_{\textit{init}}$ to the desired noise level (in a single step) and manipulate the denoising diffusion process in the following way: at each step, we first perform a latent denoising step, conditioned directly on the guiding text prompt $d$, to obtain a less noisy foreground latent denoted as $z_{\textit{fg}}$, while also noising the original latent $z_{\textit{init}}$ to the current noise level to obtain a noisy background latent $z_{\textit{bg}}$. The two latents are then blended using the resized mask, i.e. $z_{\textit{fg}} \odot m_{\textit{latent}} + z_{\textit{bg}} \odot (1 - m_{\textit{latent}})$, to yield the latent for the next latent diffusion step. Similarly to Blended Diffusion, at each denoising step the entire latent is modified, but the subsequent blending enforces the parts outside $m_{\textit{latent}}$ to remain the same. While the resulting blended latent is not guaranteed to be coherent, the next latent denoising step makes it so. Once the latent diffusion process terminates, we decode the resultant latent to the output image using the decoder $D(z)$. A visualization of the diffusion process is available in the supplementary material.

Operating on the latent level, in comparison to operating directly on pixels using a CLIP model, has the following main advantages:

\textbf{Faster inference}: The smaller dimension of the latent space makes the diffusion process much faster. In addition, there is no need to calculate the CLIP-loss gradients at each denoising step. Thus, the entire editing process is faster by an order of magnitude (see \Cref{sec:inference_time_comparison}).
    
\textbf{Avoiding pixel-level artifacts}: Pixel-level diffusion sometimes results in pixel values outside the valid range, producing noticeable clipping artifacts. Operating in the latent space avoids such artifacts (\Cref{fig:noise_artifact_comparison}).
    
\textbf{Avoiding adversarial examples}: Operating on the latent space with no pixel-level CLIP-loss gradients effectively eliminates the risk of adversarial examples, eliminating the need for the extending augmentations of Avrahami \etal~\shortcite{avrahami2022blended}.

\textbf{Better precision:} Our method achieves better precision than the baselines, both at the batch level and at the final prediction level (\Cref{sec:results}).

However, operating in latent space also introduces some drawbacks, which we will address later in this section:

\textbf{Imperfect reconstruction}: The VAE latent encoding is lossy; hence, the final results are upper-bounded by the decoder's reconstruction abilities. Even the initial reconstruction, before performing any diffusion, often visibly differs from the input. In images of human faces, or images with high frequencies, even such slight changes may be perceptible (see \Cref{fig:reconstruction_comparison}(b)).

\textbf{Thin masks}: When the input mask $m$ is relatively thin (and its downscaled version $m_{\textit{latent}}$ can become even thinner), the effect of the edit might be limited or non-existent (see \Cref{fig:thin_masks_ablation}).

\input{figures/reconstruction_ablation/fig.tex}

\subsection{Background Reconstruction}
\label{sec:handling_innacurate_reconstruction}

As discussed above, LDM's latent representation is obtained using a VAE \cite{kingma2013auto}, which is lossy.
As a result, the encoded image is not reconstructed exactly, even before any latent diffusion takes place (\Cref{fig:reconstruction_comparison}(a)). The imperfect reconstruction may thus be visible in areas outside the mask (\Cref{fig:reconstruction_comparison}(b)). 

A \naive{} way to deal with this problem is to stitch the original image and the edited result $\hat{x}$ at the pixel level, using the input mask $m$. However, because the unmasked areas were not generated by the decoder, there is no guarantee that the generated part will blend seamlessly with the surrounding background. Indeed, this \naive{} stitching produces visible seams, as demonstrated in \Cref{fig:reconstruction_comparison}(c).

Alternatively, one could perform seamless cloning between the edited region and the original, e.g., utilizing Poisson Image Editing \cite{perez2003poisson}, which uses gradient-domain reconstruction in pixel space. 
However, this often results in a noticeable color shift of the edited area, as demonstrated in \Cref{fig:reconstruction_comparison}(d).

In the GAN inversion literature \cite{abdal2019image2stylegan, abdal2020image2stylegan++,zhu2020domain,xia2021gan} it is standard practice to achieve image reconstruction via latent-space optimization. In theory, latent optimization can also be used to perform seamless cloning, as a post-process step:
given the input image $x$, the mask $m$, and the edited image $\hat{x}$, along with its corresponding latent vector $z_0$, one could use latent optimization to search for a better vector $z^*$, s.t. the masked area will be similar to the edited image $\hat{x}$ and the unmasked area will be similar to the input image $x$: 
\begin{equation}
    \label{eqn:latent_optimization}
    z^* = \operatornamewithlimits{argmin}\limits_{z} \| D(z) \odot m - \hat{x} \odot m \| \;+ \\ \lambda \| D(z) \odot (1 - m) - x \odot (1 - m) \|  
\end{equation}
using a standard distance metric, such as MSE. $\lambda$ is a hyperparameter that controls the importance of the background preservation, which we set to $\lambda = 100$ for all our results and comparisons. The optimization process is initialized with $z^* = z_0$. The final image is then inferred from $z^*$ using the decoder: $x^*=D(z^*).$ However, as we can see in \Cref{fig:reconstruction_comparison}(e), even though the resulting image is closer to the input image, it is over-smoothed.

The inability of latent space optimization to capture the high-frequency details suggests that the expressivity of the decoder $D(z)$ is limited. This leads us again to draw inspiration from GAN inversion literature --- it was shown \cite{bau2020semantic, pan2021exploiting,roich2021pivotal, tzaban2022stitch} that fine-tuning the GAN generator weights per image results in a better reconstruction. Inspired by this approach, we can achieve seamless cloning by fine-tuning the decoder's weights $\theta$ on a per-image basis:
\begin{equation}
    \label{eqn:weights_optimization}
    \theta^* = \operatornamewithlimits{argmin}\limits_{\theta} \| D_{\theta}(z_0) \odot m - \hat{x} \odot m \| \;+ \\ \lambda \| D_{\theta}(z_0) \odot (1 - m) - x \odot (1 - m) \|
\end{equation}
and use these weights to infer the result $x^*=D_{\theta^*}(z_0)$. As we can see in \Cref{fig:reconstruction_comparison}(f), this method yields the best result: the foreground region follows $\hat{x}$, while the background preserves the fine details from the input image $x$, and the blending appears seamless.

In contrast to Blended Diffusion \cite{avrahami2022blended}, in our method the background reconstruction is optional. Thus, it is only needed in cases where the unmasked area contains perceptually important fine-detail content, such as faces, text, structured textures, etc. A few reconstruction examples are shown in \Cref{fig:reconstruction_ablation}.

\subsection{Progressive Mask Shrinking}
\label{sec:handling_thin_masks}

\input{figures/thin_masks_progression/fig.tex}

When the input mask $m$ has thin parts, these parts may become even thinner in its downscaled version $m_{\textit{latent}}$, to the point that changing the latent values under $m_{\textit{latent}}$ by the text-driven diffusion process fails to produce a visible change in the reconstructed result. %
In order to pinpoint the root-cause, we visualize the diffusion process: given a noisy latent $z_t$ at timestep $t$, we can estimate $z_0$ using a single diffusion step with the closed form formula derived by Song \etal~\shortcite{song2020denoising}.
The corresponding image is then inferred using the VAE decoder $D(z_0)$.

Using the above visualization, \Cref{fig:thin_masks_progression} shows that during the denoising process, the earlier steps generate only rough colors and shapes, which are gradually refined to the final output. The top row shows that even though the guiding text ``fire'' is echoed in the latents early in the process, blending these latents with $z_{\textit{bg}}$ using a thin $m_{\textit{latent}}$ mask may cause the effect to disappear.

This understanding suggests the idea of \emph{progressive mask shrinking}: because the early noisy latents correspond to only the rough colors and shapes, we start with a rough, dilated version of $m_{\textit{latent}}$, and gradually shrink it as the diffusion process progresses, s.t. only the last denoising steps employ the thin $m_{\textit{latent}}$ mask when blending $z_{\textit{fg}}$ with $z_{\textit{bg}}$. The process is visualized in \Cref{fig:thin_masks_progression}.
For more implementation details and videos visualizing the process, please see the supplementary material. 

\Cref{fig:thin_masks_ablation} demonstrates the effectiveness of this method. Nevertheless, this technique struggles in generating fine details (e.g. the ``green bracelet'' example).

\input{figures/thin_masks_ablation/fig.tex}

\subsection{Prediction Ranking}
\label{sec:predictions_ranking}

Due to the stochastic nature of the diffusion process, we can generate multiple predictions for the same inputs, which is desirable because of the one-to-many nature of our problem. As in previous works \cite{razavi2019generating, ramesh2021zero,avrahami2022blended}, we found it beneficial to generate multiple predictions, rank them, and retrieve the best ones.
We rank the predictions by the normalized cosine distance between their CLIP embeddings and the CLIP embedding of the guiding prompt $d$. We also use the same ranking for all of the baselines that we compare our method against, except $\textit{PaintByWord++}$ \cite{bau2021paint, crowson2022vqgan}, as it produces a single output per input, and thus no ranking is required.

%% file: figures/method_illustration/fig.tex
\begin{figure*}
    \centering
    \setlength{\ww}{2\columnwidth}
    
    \includegraphics[width=\ww]{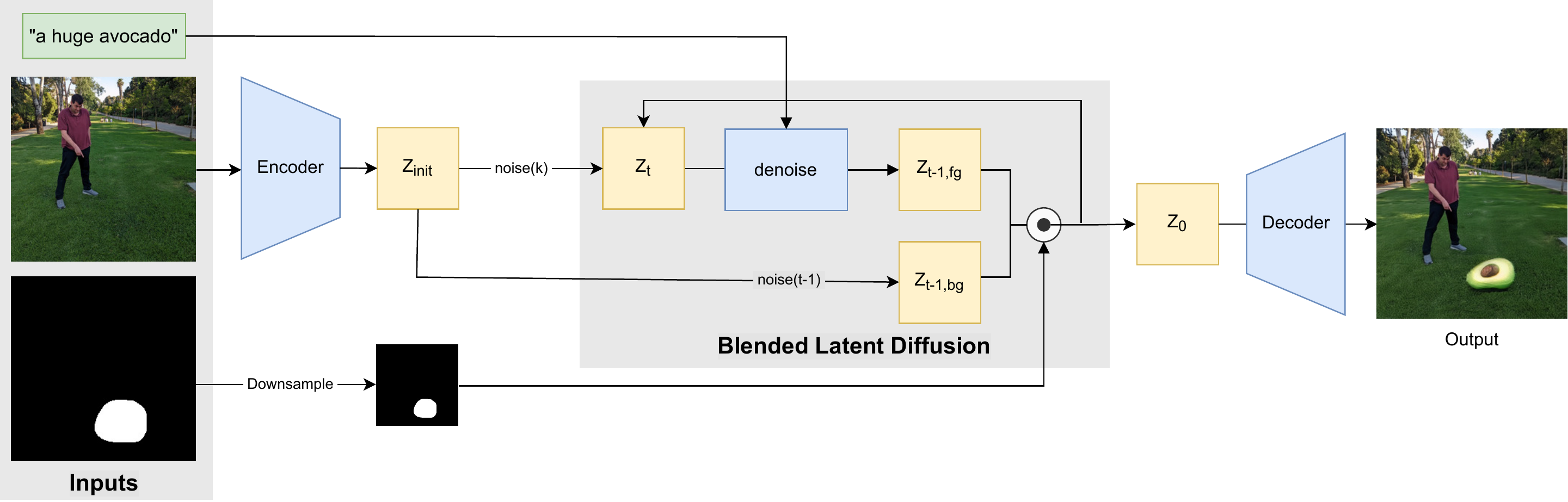}
    
    \caption{\textbf{Blended Latent Diffusion:} a diagram illustrating our method, as described in
    \Cref{alg:latent_blended_diffusion}.}
    \label{fig:method_illusration}
\end{figure*}

%% file: algorithms/blended_latent_diffusion.tex
\begin{figure}[t]
\removelatexerror
\begin{algorithm}[H]
    \caption{Latent Blended Diffusion: given a text-conditioned Latent Diffusion model $\{\textit{VAE} = (E(x), D(z)), \textit{DiffusionModel}=(\textit{noise}(z, t), \textit{denoise}(z, d, t))\}$}
    \label{alg:latent_blended_diffusion}
    \begin{algorithmic}
        \STATE \textbf{Input:} source image $x$, target text description $d$, input mask $m$, diffusion steps $k$.
        \STATE \textbf{Output:} edited image $\widehat{x}$ that differs from input image $x$ inside area $m$ according to text description $d$ 
        \STATE
        \STATE $m_{\textit{latent}} = \textit{downsample}(m)$
        \STATE $z_{\textit{init}} \sim E(x)$
        \STATE $z_{k} \sim \textit{noise}(z_{\textit{init}}, k)$
        \FORALL{$t$ from $k$ to 0}
            \STATE $z_{\textit{fg}} \sim \textit{denoise}(z_t, d, t)$
            \STATE $z_{\textit{bg}} \sim \textit{noise}(z_{\textit{init}}, t)$
            \STATE $z_{t} \gets z_{\textit{fg}} \odot m_{\textit{latent}} + z_{\textit{bg}} \odot (1 - m_{\textit{latent}})$
        \ENDFOR
        \STATE $\widehat{x} = D(z_{0})$
        \RETURN $\widehat{x}$
        \STATE
    \end{algorithmic}
\end{algorithm}
\end{figure}

%% file: figures/reconstruction_comparison/fig.tex
\begin{figure*}[ht]
    \centering
    \setlength{\tabcolsep}{-2pt}
    \renewcommand{\arraystretch}{0.5}
    \setlength{\ww}{0.25\columnwidth}
  
    \begin{tabular}{cccccccc}
        \begin{tikzpicture}[spy using outlines={}]
            \node {\includegraphics[width=\ww,frame]{figures/reconstruction_comparison/assets/img.jpg}};
        \end{tikzpicture} &

        \begin{tikzpicture}[spy using outlines={}]
            \node {\includegraphics[width=\ww,frame]{figures/reconstruction_comparison/assets/mask_overlay.jpg}};
        \end{tikzpicture} &

        \begin{tikzpicture}[spy using outlines={circle,yellow,magnification=2,size=0.6cm, connect spies}]
            \node {\includegraphics[width=\ww,frame]{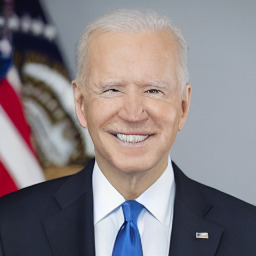}};
            \spy on (0.05,-0.12) in node [left] at (1.05,-0.7);
            \spy on (-0.13,0.31) in node [left] at (-0.35,-0.7);
        \end{tikzpicture} &

        \begin{tikzpicture}[spy using outlines={circle,yellow,magnification=2,size=0.6cm, connect spies}]
            \node {\includegraphics[width=\ww,frame]{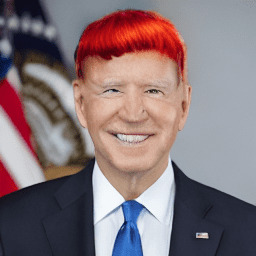}};
            \spy on (0.05,-0.12) in node [left] at (1.05,-0.7);
            \spy on (-0.13,0.31) in node [left] at (-0.35,-0.7);
        \end{tikzpicture} &

        \begin{tikzpicture}[spy using outlines={circle,yellow,magnification=2,size=0.6cm, connect spies}]
            \node {\includegraphics[width=\ww,frame]{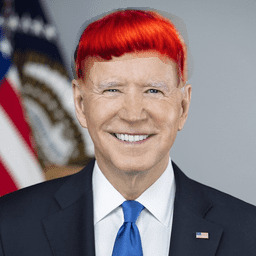}};
            \spy on (-0.2,0.6) in node [left] at (-0.35,-0.7);
        \end{tikzpicture} &

        \begin{tikzpicture}[spy using outlines={circle,yellow,magnification=2,size=0.6cm, connect spies}]
            \node {\includegraphics[width=\ww,frame]{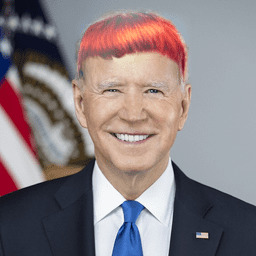}};
            \spy on (-0.2,0.6) in node [left] at (-0.35,-0.7);
        \end{tikzpicture} &

        \begin{tikzpicture}[spy using outlines={circle,yellow,magnification=2,size=0.6cm, connect spies}]
            \node {\includegraphics[width=\ww,frame]{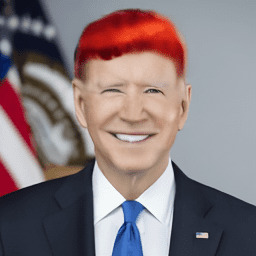}};
            \spy on (0.05,-0.12) in node [left] at (1.05,-0.7);
        \end{tikzpicture} &

        \begin{tikzpicture}[spy using outlines={circle,yellow,magnification=2,size=0.6cm, connect spies}]
            \node {\includegraphics[width=\ww,frame]{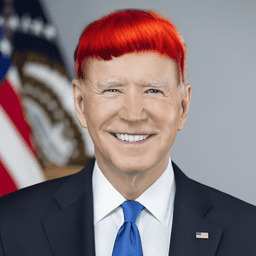}};
        \end{tikzpicture}
        \\
        
        \scriptsize{Input image} &
        \scriptsize{Input mask} &
        \scriptsize{(a) VAE reconstruction} &
        \scriptsize{(b) Edit result} &
        \scriptsize{(c) Pixel blending} &
        \scriptsize{(d) Poisson blending} &
        \scriptsize{(e) Latent optimization} &
        \scriptsize{(f) Weights optimization}
        \\
    \end{tabular}
    \caption{\textbf{Background reconstruction comparison:} Given the input image, mask, and guiding text prompt ``red hair'', the reconstruction does not preserve the unmasked area details (a,b). Pixel-level blending yields a result (c) with noticeable seams. Poisson seamless cloning (d) changes the colors of the edited area, while latent optimization (e) produces an over smoothed result. We propose per-sample weights optimization (f) which produces the best results.}
    \label{fig:reconstruction_comparison}
\end{figure*}

%% file: figures/reconstruction_ablation/fig.tex
\begin{figure}[t]
    \centering
    \setlength{\tabcolsep}{0.5pt}
    \renewcommand{\arraystretch}{0.5}
    \setlength{\ww}{0.23\columnwidth}
  
    \begin{tabular}{ccccc}
        \rotatebox{90}{\phantom{AAa}\scriptsize{text ``GAN''}} &
        \includegraphics[width=\ww,frame]{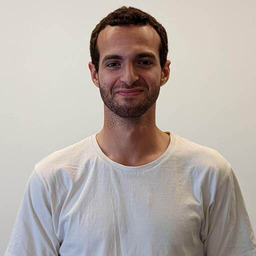} &
        \includegraphics[width=\ww,frame]{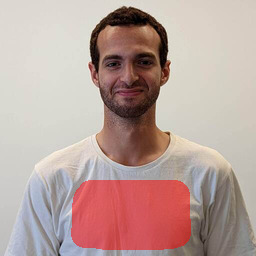} &
        \includegraphics[width=\ww,frame]{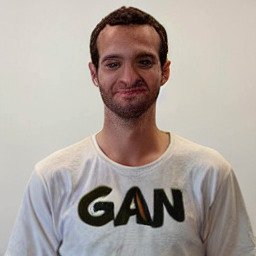} &
        \includegraphics[width=\ww,frame]{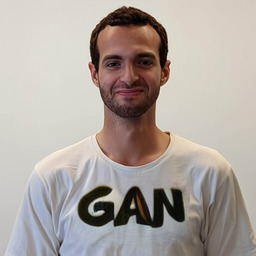}
        \\

        \rotatebox{90}{\phantom{AA} \scriptsize{``sculpture''}} &
        \includegraphics[width=\ww,frame]{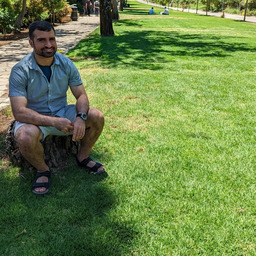} &
        \includegraphics[width=\ww,frame]{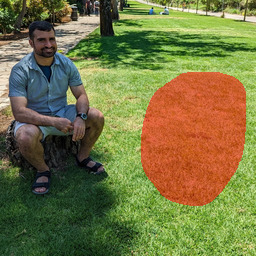} &
        \includegraphics[width=\ww,frame]{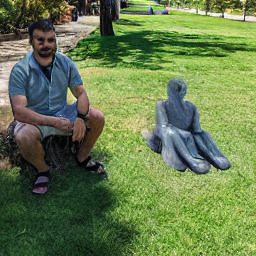} &
        \includegraphics[width=\ww,frame]{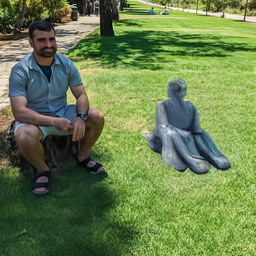}
        \\

        \rotatebox{90}{\phantom{AAA}\scriptsize{``cucumber''}} &
        \includegraphics[width=\ww,frame]{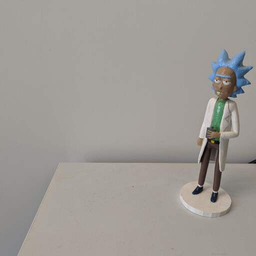} &
        \includegraphics[width=\ww,frame]{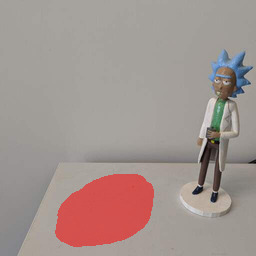} &
        \includegraphics[width=\ww,frame]{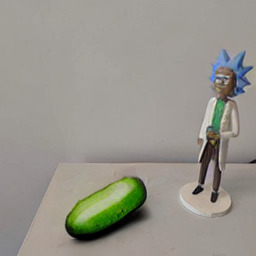} &
        \includegraphics[width=\ww,frame]{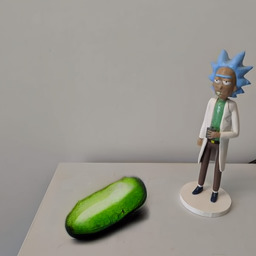}
        \\

        \rotatebox{90}{\phantom{a}\scriptsize{``heart wall painting''}} &
        \includegraphics[width=\ww,frame]{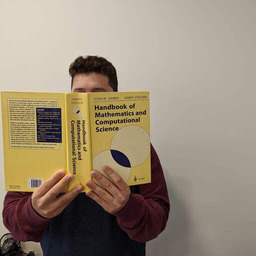} &
        \includegraphics[width=\ww,frame]{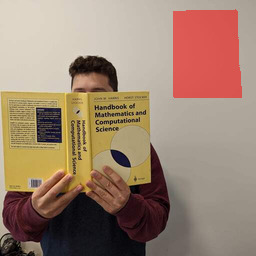} &
        \includegraphics[width=\ww,frame]{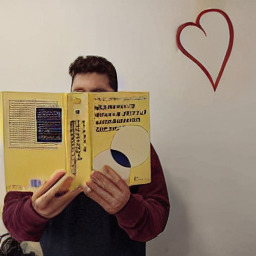} &
        \includegraphics[width=\ww,frame]{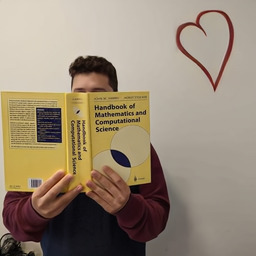}
        \\

        \rotatebox{90}{\phantom{Aa}\scriptsize{``corgi painting''}} &
        \includegraphics[width=\ww,frame]{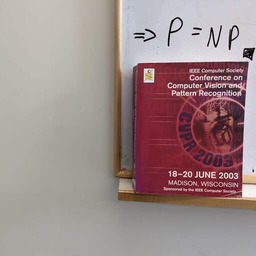} &
        \includegraphics[width=\ww,frame]{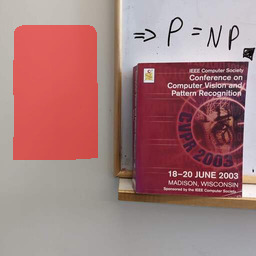} &
        \includegraphics[width=\ww,frame]{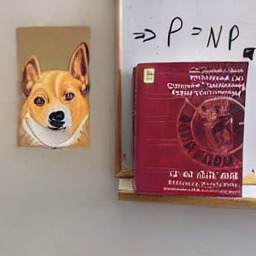} &
        \includegraphics[width=\ww,frame]{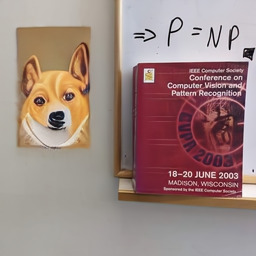}
        \\
        
        &
        \scriptsize{Input image} &
        \scriptsize{Input mask} &
        \scriptsize{Initial prediction} &
        \scriptsize{After reconstruction}
        \\
    \end{tabular}
    \caption{\textbf{Background reconstruction using decoder weights fine-tuning:} Note the bad initial prediction of the high-frequency background areas: the human face in the 1st and 2nd rows, the doll face in the 3rd row, and the text on the books on the 4th and 5th (zoom in for a better presentation).}
    \label{fig:reconstruction_ablation}
\end{figure}

%% file: figures/thin_masks_progression/fig.tex
\begin{figure}[t]
    \centering
    \setlength{\tabcolsep}{0.5pt}
    \renewcommand{\arraystretch}{0.5}
    \setlength{\ww}{0.19\columnwidth}
  
    \begin{tabular}{ccccccc}
        &
        \scriptsize{early steps} 
        &&&&
        \scriptsize{final steps}
        \\
        
        \rotatebox{90}{\phantom{A}\scriptsize{(1) Standard}} &
        \includegraphics[width=\ww,frame]{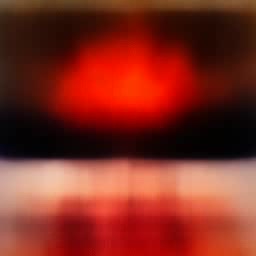} &
        \includegraphics[width=\ww,frame]{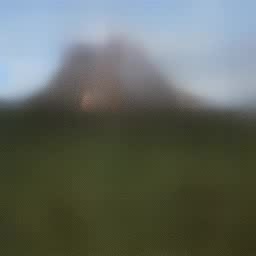} &
        \includegraphics[width=\ww,frame]{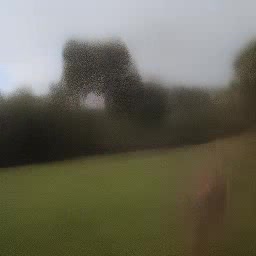} &
        \includegraphics[width=\ww,frame]{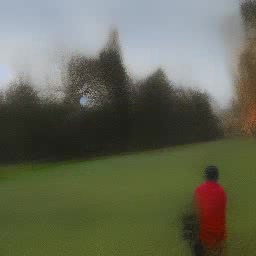} &
        \includegraphics[width=\ww,frame]{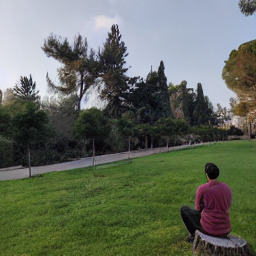}
        \\

        \rotatebox{90}{\scriptsize{(2) Pro. shrinking}} &
        \includegraphics[width=\ww,frame]{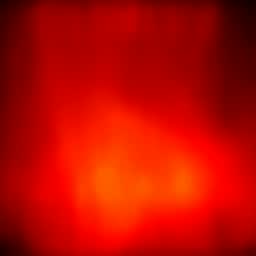} &
        \includegraphics[width=\ww,frame]{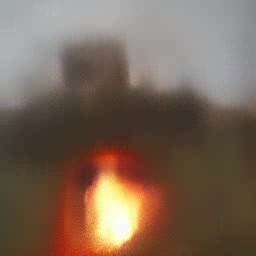} &
        \includegraphics[width=\ww,frame]{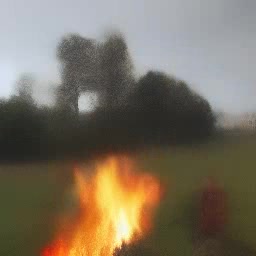} &
        \includegraphics[width=\ww,frame]{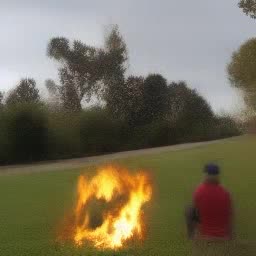} &
        \includegraphics[width=\ww,frame]{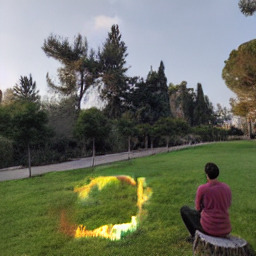}
        \\

        \rotatebox{90}{\phantom{Aa}\scriptsize{(3) Masks}} &
        \includegraphics[width=\ww,frame]{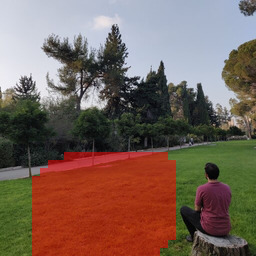} &
        \includegraphics[width=\ww,frame]{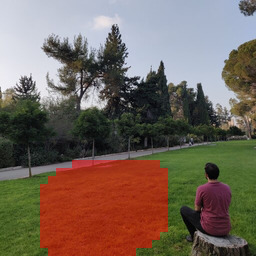} &
        \includegraphics[width=\ww,frame]{figures/thin_masks_progression/assets/mask1_overlay.jpg} &
        \includegraphics[width=\ww,frame]{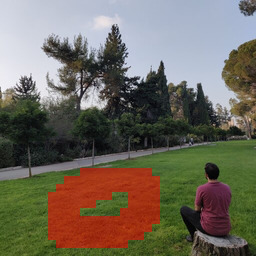} &
        \includegraphics[width=\ww,frame]{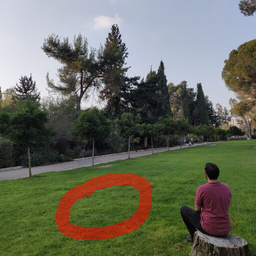}
        \\

        &
        \scriptsize{dilated mask}
        &&&&
        \scriptsize{original mask}
        \\
    \end{tabular}
    \caption{\textbf{Thin mask progression:} Given the input image, mask (bottom right corner), and guiding text ``fire'', in the standard case (1) only the initial stages correspond to the text (rough red colors), but later the blending overrides it. In contrast, using our progressively shrinking masks (3) the guiding text corresponds to all the images throughout the diffusion process (2). 
    }
    \label{fig:thin_masks_progression}
\end{figure}

%% file: figures/thin_masks_ablation/fig.tex
\begin{figure}[t]
    \centering
    \setlength{\tabcolsep}{0.5pt}
    \renewcommand{\arraystretch}{0.5}
    \setlength{\ww}{0.3\columnwidth}
  
    \begin{tabular}{cccc}
	
    &
	``white clouds'' & 
    ``green smoke'' & 
    ``green bracelet''
    \\
	
    \rotatebox{90}{{input image + mask}} &
	\includegraphics[width=\ww,frame]{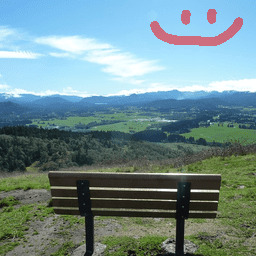} &
	\includegraphics[width=\ww,frame]{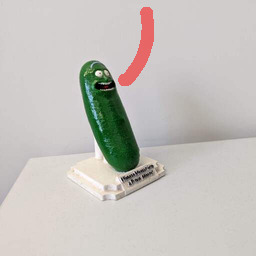} &	
    \includegraphics[width=\ww,frame]{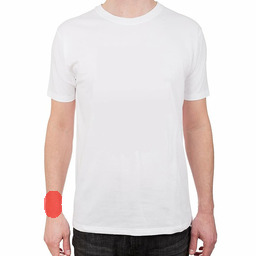} \\

    \rotatebox{90}{\phantom{a}{w/o prog. shrink}} &
    \includegraphics[width=\ww,frame]{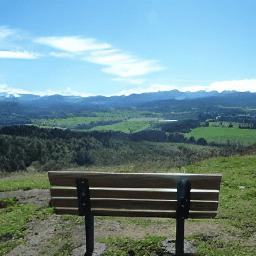} &
	\includegraphics[width=\ww,frame]{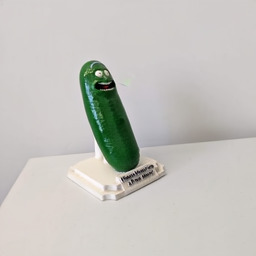} &	
    \includegraphics[width=\ww,frame]{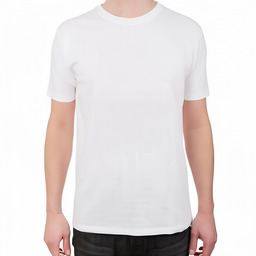} \\
	
    \rotatebox{90}{\phantom{a}{with prog. shrink}} &
	\includegraphics[width=\ww,frame]{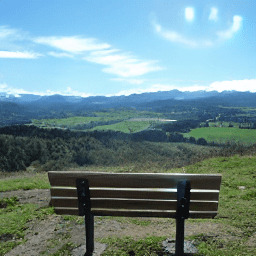} &
	\includegraphics[width=\ww,frame]{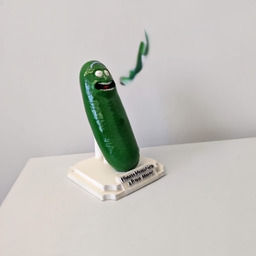} &
	\includegraphics[width=\ww,frame]{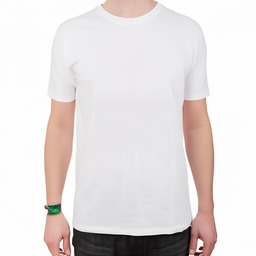}
	\\
\end{tabular}
\caption{\textbf{Progressive mask shrinking:} With the thin input masks in these examples (first row), the method described in \Cref{alg:latent_blended_diffusion} fails to alter the image according to the text (second row). This issue is mitigated using progressive mask shrinking (third row).}
\label{fig:thin_masks_ablation}
\end{figure}

%% file: sections/results.tex
\input{figures/comparisons/baselines/fig.tex}

\section{Results}
\label{sec:results}

We begin by comparing our method against previous methods, both qualitatively and quantitatively. Next, we demonstrate several of the use cases enabled by our method.

\subsection{Comparisons}
In \Cref{fig:baselines_comparison} we compare the zero-shot text-driven image editing results produced by our method against the following baselines: (1) Local CLIP-guided diffusion \cite{clip_guided_diffusion}, (2) $\textit{PaintByWord++}$ \cite{bau2021paint, crowson2022vqgan}, (3) Blended Diffusion \cite{avrahami2022blended}, (4) GLIDE \cite{nichol2021glide}, (5) GLIDE-masked \cite{nichol2021glide}, (6) GLIDE-filtered \cite{nichol2021glide}, and (7) \DALLE~2.
See Avrahami \etal~\shortcite{avrahami2022blended} for more details on baselines (1)--(3). The images for the baselines (1)--(5) were taken directly from the corresponding papers.
Note that Nichol \etal~\shortcite{nichol2021glide} only released GLIDE-filtered, a smaller version of GLIDE, which was trained on a filtered dataset, and this is the only public version of GLIDE. Because the (4) full GLIDE model and (5) GLIDE-masked are not available, we use the results from the paper \cite{nichol2021glide}. The images for (3)--(6) and our method required generating a batch of samples and taking the best one ranked by CLIP. The GLIDE model has about $\times 3$ the parameters vs.~our model. See the supplementary materials for more details.

\Cref{fig:baselines_comparison} demonstrates that baselines (1) and (2) do not always preserve the background of the input image. The edits by GLIDE-filtered (6) typically fail to follow the guiding text. So the comparable baselines are (3) Blended Diffusion, (4) GLIDE, (5) GLIDE-masked, and (7) \DALLE~2. As we can see, our method avoids the pixel-level noises of Blended Diffusion (e.g., the pizza example) and generates better colors and textures (e.g., the dog collar example). Comparing to GLIDE, we see that in some cases GLIDE generates better shadows than our method (e.g., the cat example), however it can add artifacts (e.g., the front right paw of the cat in GLIDE's prediction). Furthermore, GLIDE's generated results do not always follow the guiding text (e.g., the golden necklace and blooming tree examples), hence, the authors of GLIDE propose GLIDE-masked, a version of GLIDE that does not take into account the given image --- by fully masking the context. Using this approach, they manage to generate in the masked area, but it comes at the expense of the transition quality between the masked region and the background
(e.g., the plate in the pizza example and the bone in the dogs example). Our method is able to generate a result that corresponds to the text in all the examples, while being blended into the scene seamlessly.

Inspecting \DALLE~2 results, we see that most of the results either ignore the guiding text (e.g., the dog collar, dog bone, and pizza examples) or only partially follow it (e.g., ``golden necklace'' generates a regular necklace, ``blooming tree'' generates a flower, and ``blue short pants'' generates text on top of the pants). For more 
examples, please see the supplementary material.

\input{figures/experiment_results/fig_min.tex}

During our experiments, we noticed that the predictions of our method typically contain more results that comply with the guiding text prompt. In order to verify this quantitatively, we generated editing predictions for 50 random images, random masks, and text prompts randomly chosen from ImageNet classes. See \Cref{fig:experiment_results_min_example} for some examples. Then, batch precision was evaluated using an off-the-shelf ImageNet classifier. We refrained from using CLIP cosine similarity as the precision measure, because it was shown that CLIP operates badly as an evaluator for gradient-based solutions that use CLIP, due to adversarial attacks \cite{nichol2021glide}. We denote this measure as the ``precision'' of the model. For more details see 
\Cref{sec:precision_and_diversity_supp}
As reported in \Cref{tab:metrics_comparison}, our method indeed outperforms the baselines by a large margin. In addition, we ranked the results in the batch as described in \Cref{sec:predictions_ranking} and calculated the average accuracy by taking only the top image in each batch, to find that our method still outperforms the baselines.

We also assess the average batch diversity, by calculating the pairwise LPIPS \cite{zhang2018unreasonable} distances between all the masked predictions in the batch that were classified correctly by the classifier. As can be seen in \Cref{tab:metrics_comparison}, our method has the second-best diversity, but it is outperformed by Local CLIP-guided diffusion by a large margin, which we attribute to the fact that this method changes the entire image (does not preserve the background) and thus the content generated in the masked area is much less constrained.

In addition, we conducted a user study on Amazon Mechanical Turk (AMT) \cite{amt_website} to assess the visual quality and text-matching of our results. Each of the 50 random predictions that were used in the quantitative evaluation was presented to a human evaluator next to a result from one of the baselines. The evaluator was asked to choose which of the two results has a better (1) visual quality and (2) matches the text more closely. The evaluators could also indicate that neither image is better than the other.
As seen in \Cref{tab:metrics_comparison} (right), the majority ($\ge 50\%$) of evaluators prefer the visual quality and the text matching of our method over the other methods. A binomial statistical significance test, reported in Table 2 in the supplementary material, suggests that these results are statistically significant. The results of GLIDE-filtered \cite{nichol2021glide} were preferred in terms of visual quality, however these results typically fail to change the input image or make negligible changes: thus, although the result looks natural, it does not reflect the desired edit. See \Cref{fig:experiment_results_min_example} and the supplementary material for more examples and details. We chose to use a two-way question system in order to make the task clearer to the evaluators by providing only two images without the input image and mask.

\input{tables/metrics_comparison.tex}

\subsection{Inference Time Comparison}
\label{sec:inference_time_comparison}
\input{tables/inference_time_comparison.tex}
We compare the inference time of various methods on an A10 NVIDIA GPU in \Cref{tab:inference_time_comparison}. We show results for Blended Diffusion and GLIDE-filtered (the available smaller model, which is probably faster than the full unpublished model). Both of these methods require generating multiple predictions (batch) and taking the best one in order to achieve good results. The recommended batch size for Blended Diffusion is 64, whereas GLIDE-filtered and our method use a batch size of 24. %

Our method supports generation with or without optimizing for background preservation (\Cref{sec:handling_innacurate_reconstruction}), and we report both options in \Cref{tab:inference_time_comparison}. The background optimization introduces an additional inference time overhead, however, it is up to the user to decide whether this additional step is necessary (e.g., when editing images with human faces). Our method outperforms the baselines on the standard case of batch inference, even when accounting for the background preservation optimization. The acceleration in comparison to Blended Diffusion and Local CLIP-guided diffusion is $\times 10$ with equal batch sizes and $\times 20$ with the recommended batch sizes, which stems from the fact that our generation process is done in the lower dimensional latent space, and the background preservation optimization need only be done on the selected result. The acceleration in comparison to PaintByWord++ and GLIDE-filtered is $\times 1.47$ and $\times 1.23$, respectively.

\subsection{Use Cases}
Our method is applicable in a variety of editing scenarios with generic real-world images, several of which we demonstrate here.

\textbf{Text-driven object editing:} using our method one can easily add new objects (\Cref{fig:teaser}(top left)) or modify or replace existing ones (\Cref{fig:teaser}(top right)), guided by a text prompt. In addition, we have found that the method is capable of injecting visually plausible text into images, as demonstrated in \Cref{fig:teaser}(middle left).

\textbf{Background replacement:} rather than inserting or editing the foreground object, another important use case is text-guided background replacement, as demonstrated in \Cref{fig:teaser}(middle right).

\textbf{Scribble-guided editing:} %
The user can scribble a rough shape on a background image, provide a mask (covering the scribble) to indicate the area that is allowed to change, and provide a text prompt.
Our method transforms the scribble into a natural object while attempting to match the prompt, as demonstrated in \Cref{fig:teaser}(bottom left).

For all of the use cases mentioned above, our method is inherently capable of generating multiple predictions for the same input, as discussed in \Cref{sec:predictions_ranking} and demonstrated in \Cref{fig:teaser}(bottom right).
Due to the one-to-many nature of the task, we believe it is desirable to present the user with ranked (\Cref{sec:predictions_ranking}) multiple outcomes, from which they may chose the one that best suits their needs. Alternatively, the highest ranked result can be chosen automatically.  For more results, see 
Section A in the supplementary.

%% file: figures/comparisons/baselines/fig.tex
\begin{figure*}[h]
    \centering
    \setlength{\tabcolsep}{0.5pt}
    \renewcommand{\arraystretch}{0.5}
    \setlength{\ww}{0.243\columnwidth}
  
    \begin{tabular}{cccccccccc}
        \rotatebox{90}{\phantom{a}} &
        \rotatebox{90}{\phantom{A} Input + mask} &
        \includegraphics[width=\ww,frame]{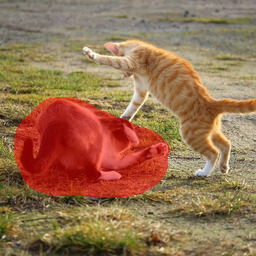} &
        \includegraphics[width=\ww,frame]{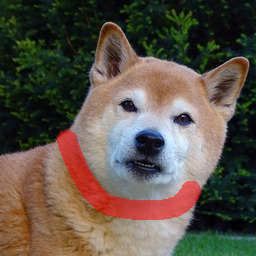} &
        \includegraphics[width=\ww,frame]{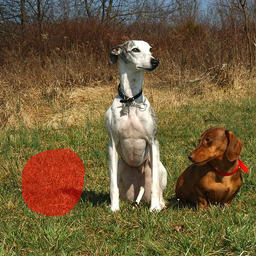} &
        \includegraphics[width=\ww,frame]{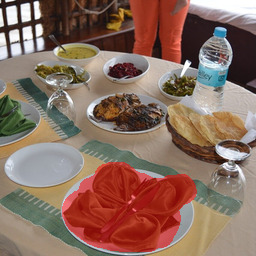} &
        \includegraphics[width=\ww,frame]{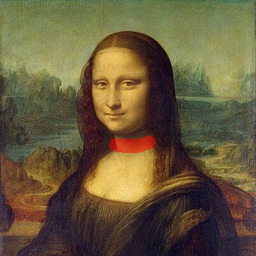} &
        \includegraphics[width=\ww,frame]{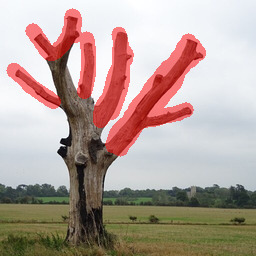} &
        \includegraphics[width=\ww,frame]{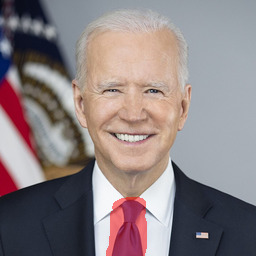} &
        \includegraphics[width=\ww,frame]{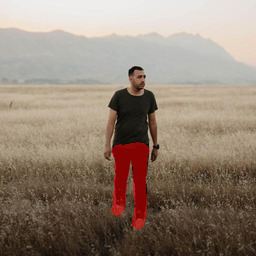} \\
        
        \rotatebox{90}{\phantom{AA}{Local CLIP-}} &
        \rotatebox{90}{guided diffusion} &
        \includegraphics[width=\ww,frame]{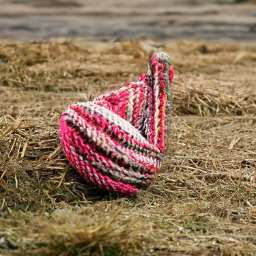} &
        \includegraphics[width=\ww,frame]{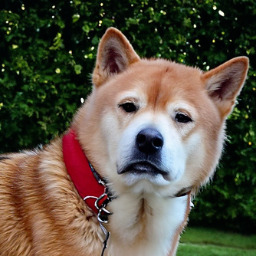} &
        \includegraphics[width=\ww,frame]{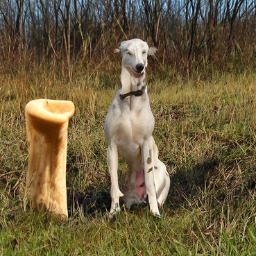} &
        \includegraphics[width=\ww,frame]{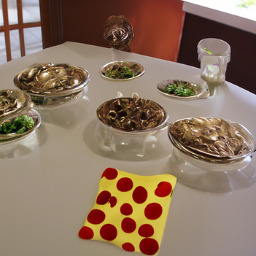} &
        \includegraphics[width=\ww,frame]{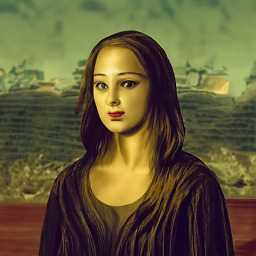} &
        \includegraphics[width=\ww,frame]{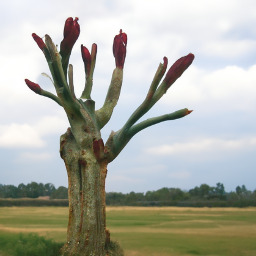} & 
        \includegraphics[width=\ww,frame]{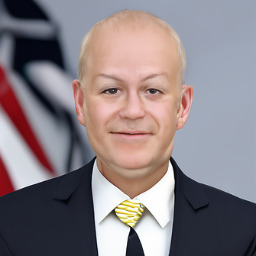} &
        \includegraphics[width=\ww,frame]{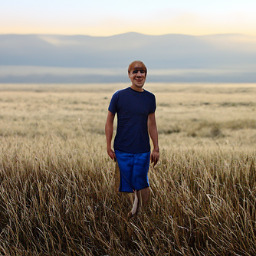} \\
        
        \rotatebox{90}{\phantom{A}} &
        \rotatebox{90}{$\textit{PaintByWord++}$} &
        \includegraphics[width=\ww,frame]{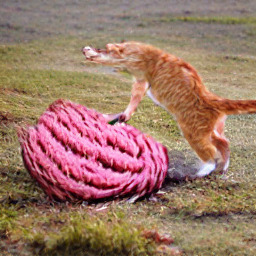} &
        \includegraphics[width=\ww,frame]{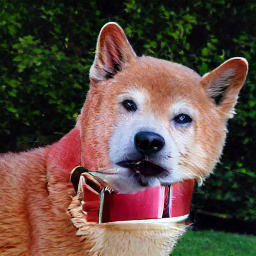} &
        \includegraphics[width=\ww,frame]{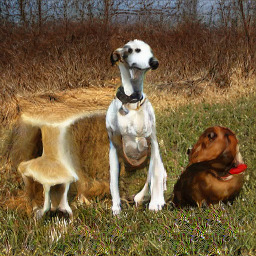} &
        \includegraphics[width=\ww,frame]{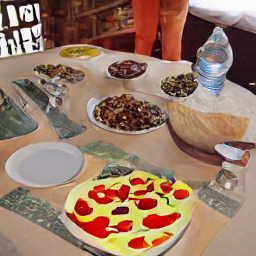} & 
        \includegraphics[width=\ww,frame]{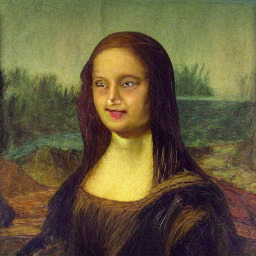} & 
        \includegraphics[width=\ww,frame]{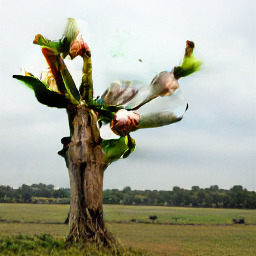} &
        \includegraphics[width=\ww,frame]{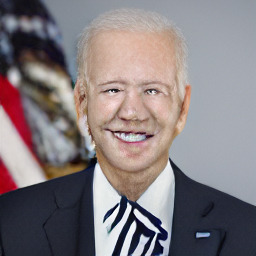} &
        \includegraphics[width=\ww,frame]{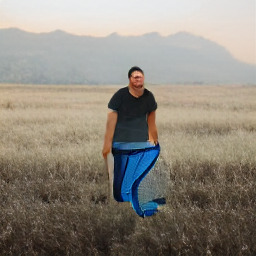} \\
        
        \rotatebox{90}{\phantom{AA}Blended} &
        \rotatebox{90}{\phantom{AA}Diffusion} &
        \includegraphics[width=\ww,frame]{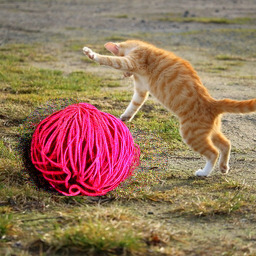} &
        \includegraphics[width=\ww,frame]{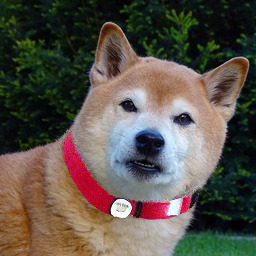} &
        \includegraphics[width=\ww,frame]{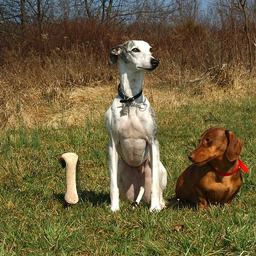} & 
        \includegraphics[width=\ww,frame]{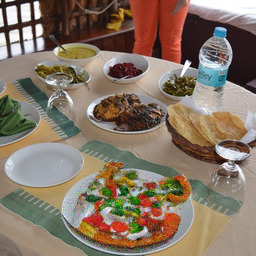} &
        \includegraphics[width=\ww,frame]{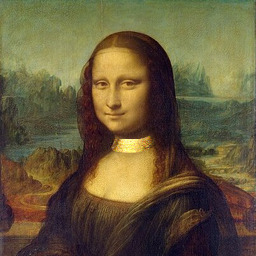} &
        \includegraphics[width=\ww,frame]{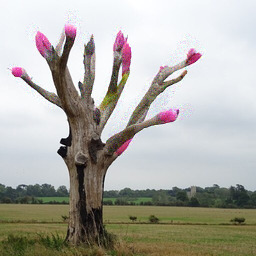} &
        \includegraphics[width=\ww,frame]{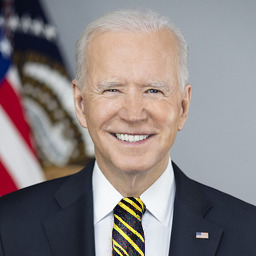} &
        \includegraphics[width=\ww,frame]{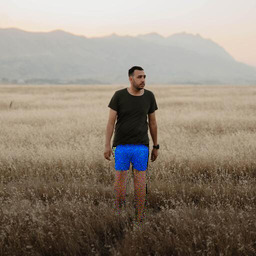} \\

        \rotatebox{90}{\phantom{A}} &
        \rotatebox{90}{\phantom{AAA}GLIDE} &
        \includegraphics[width=\ww,frame]{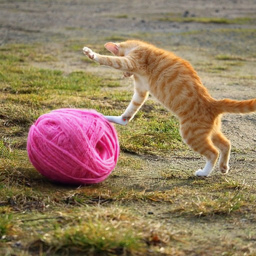} &
        \includegraphics[width=\ww,frame]{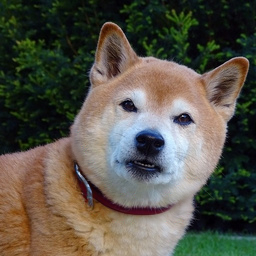} &
        \includegraphics[width=\ww,frame]{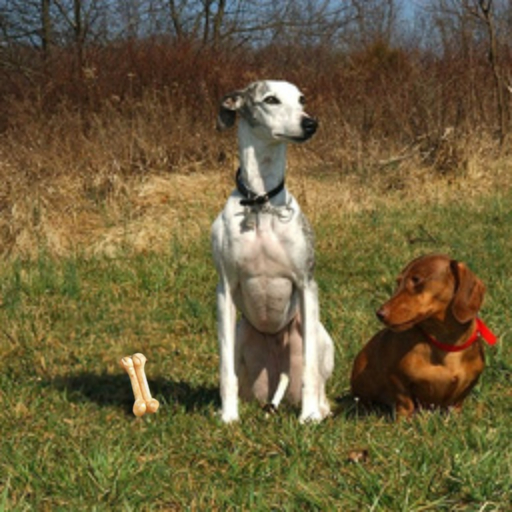} & 
        \includegraphics[width=\ww,frame]{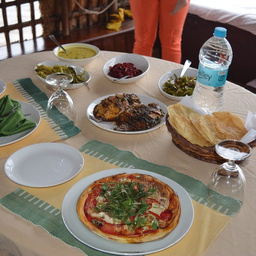} &
        \includegraphics[width=\ww,frame]{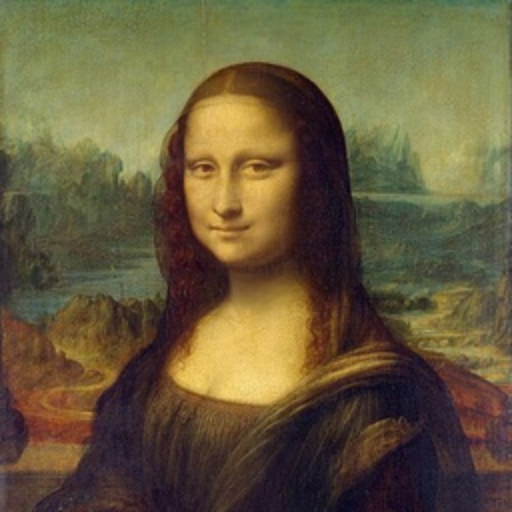} &
        \includegraphics[width=\ww,frame]{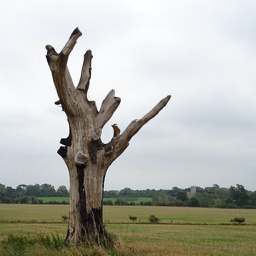} &
        \includegraphics[width=\ww,frame]{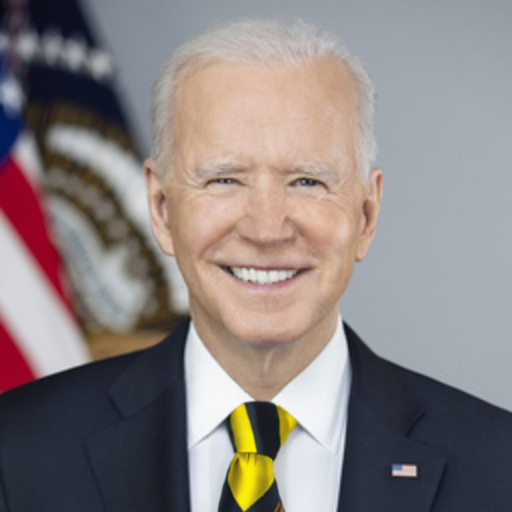} &
        \includegraphics[width=\ww,frame]{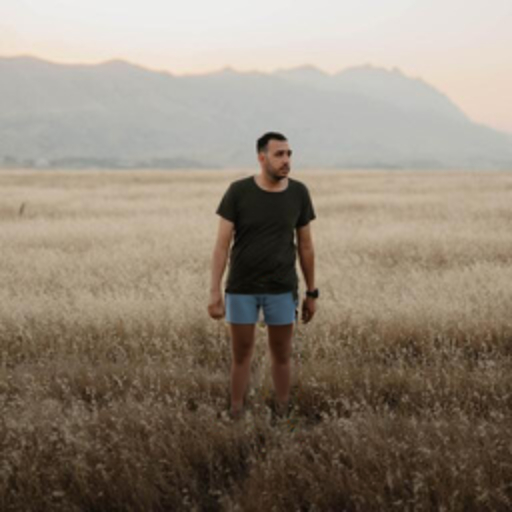} \\

        \rotatebox{90}{\phantom{A}} &
        \rotatebox{90}{\phantom{a}GLIDE-masked} &
        \includegraphics[width=\ww,frame]{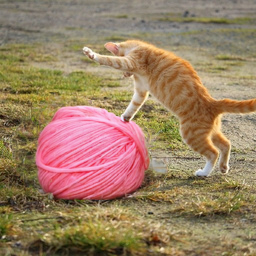} &
        \includegraphics[width=\ww,frame]{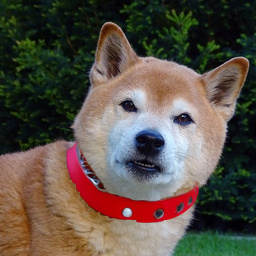} &
        \includegraphics[width=\ww,frame]{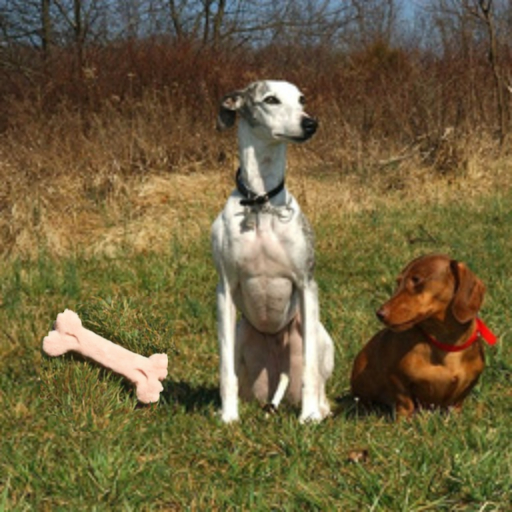} &
        \includegraphics[width=\ww,frame]{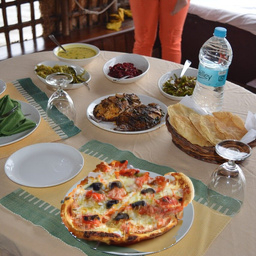} &
        \includegraphics[width=\ww,frame]{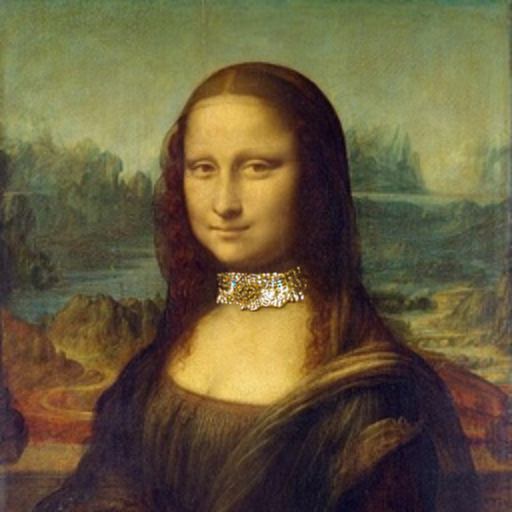} &
        \includegraphics[width=\ww,frame]{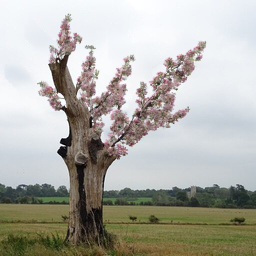} &
        \includegraphics[width=\ww,frame]{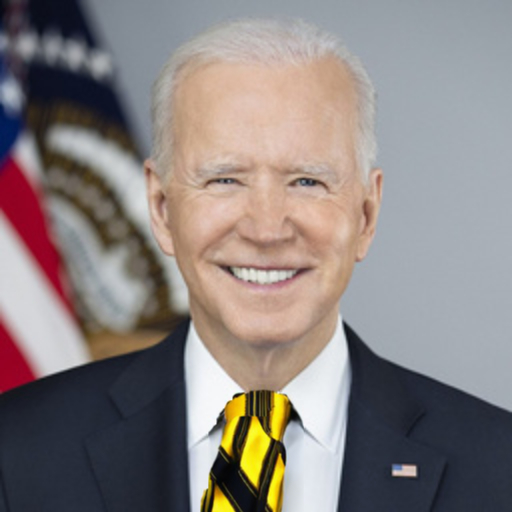} &
        \includegraphics[width=\ww,frame]{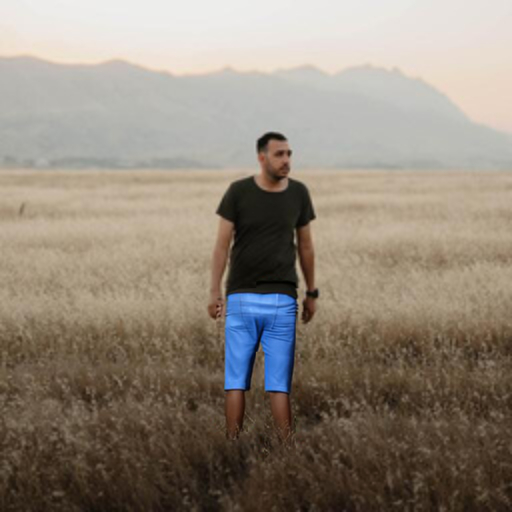} \\

        \rotatebox{90}{\phantom{A}} &
        \rotatebox{90}{\phantom{a}GLIDE-filtered} &
        \includegraphics[width=\ww,frame]{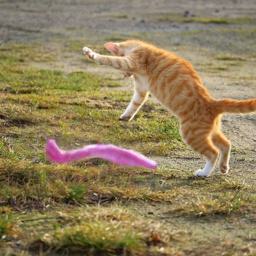} &
        \includegraphics[width=\ww,frame]{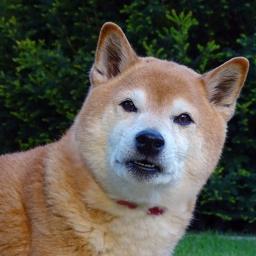} &
        \includegraphics[width=\ww,frame]{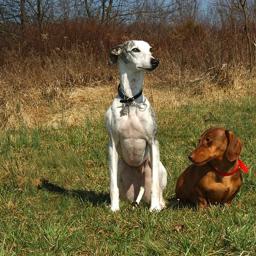} & 
        \includegraphics[width=\ww,frame]{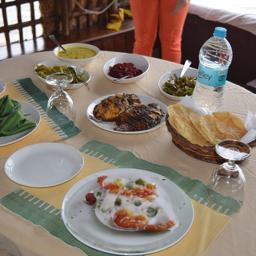} &
        \includegraphics[width=\ww,frame]{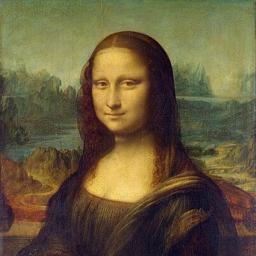} &
        \includegraphics[width=\ww,frame]{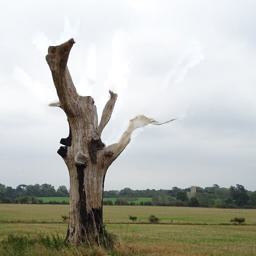} &
        \includegraphics[width=\ww,frame]{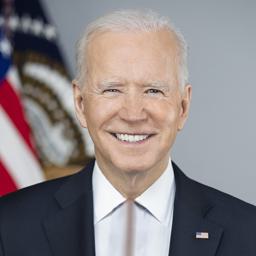} &
        \includegraphics[width=\ww,frame]{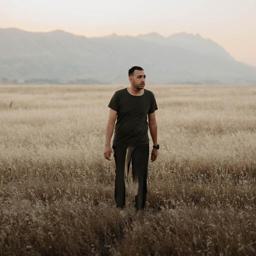} \\

        \rotatebox{90}{\phantom{A}} &
        \rotatebox{90}{\phantom{AA}\DALLE~2} &
        \includegraphics[width=\ww,frame]{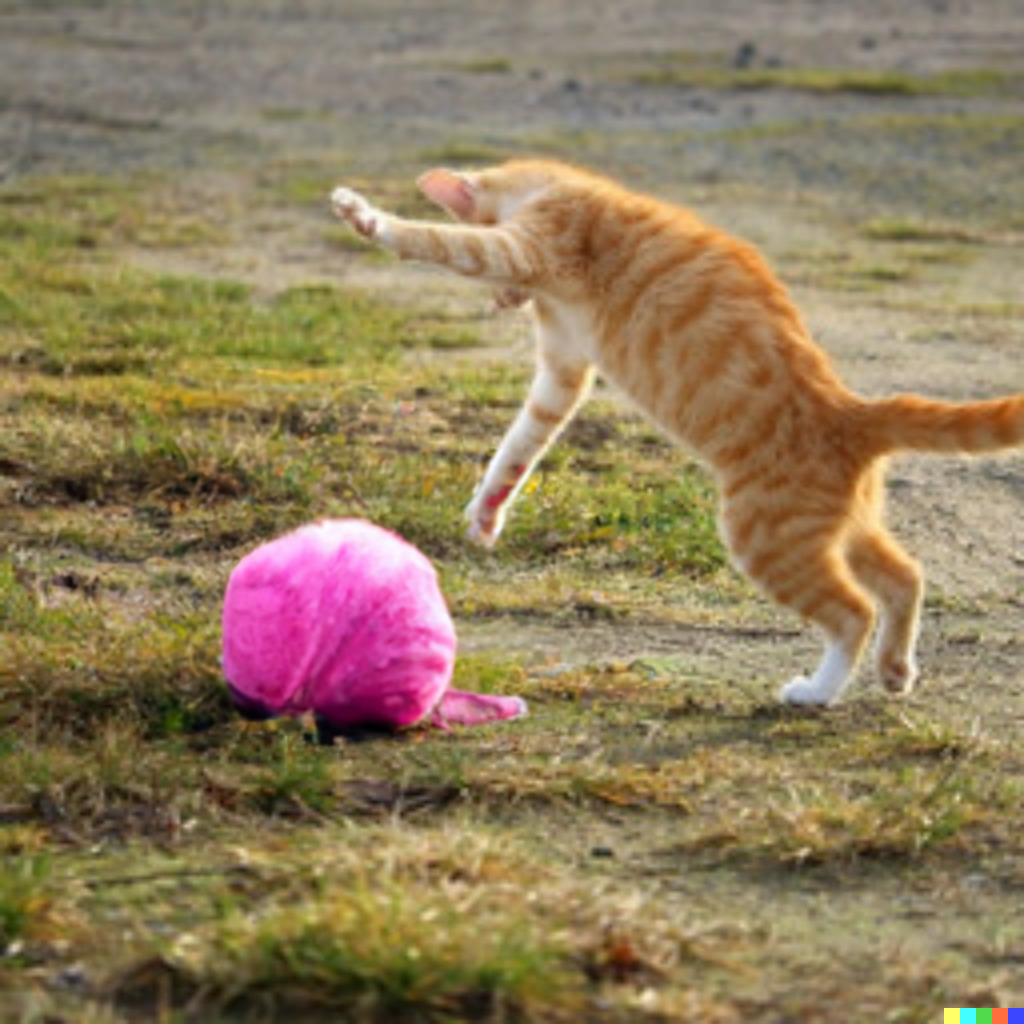} &
        \includegraphics[width=\ww,frame]{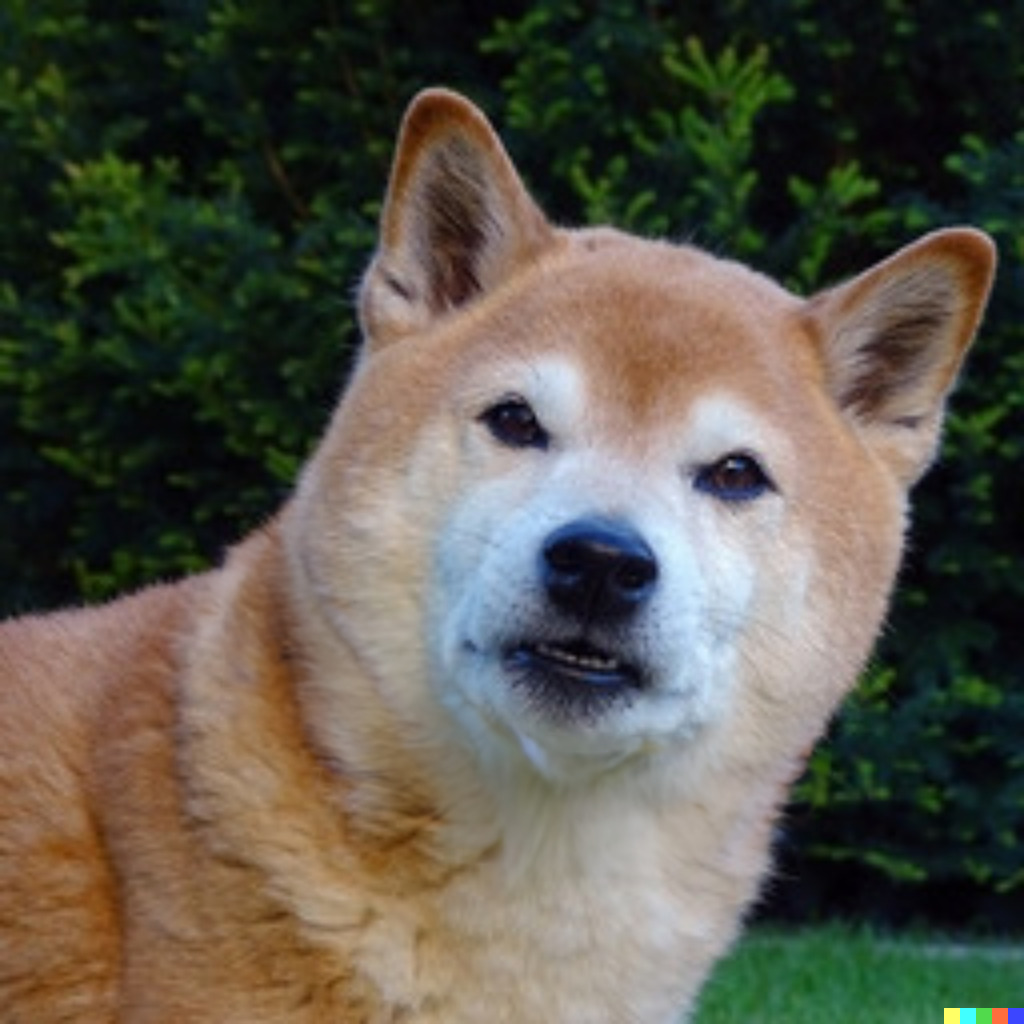} &
        \includegraphics[width=\ww,frame]{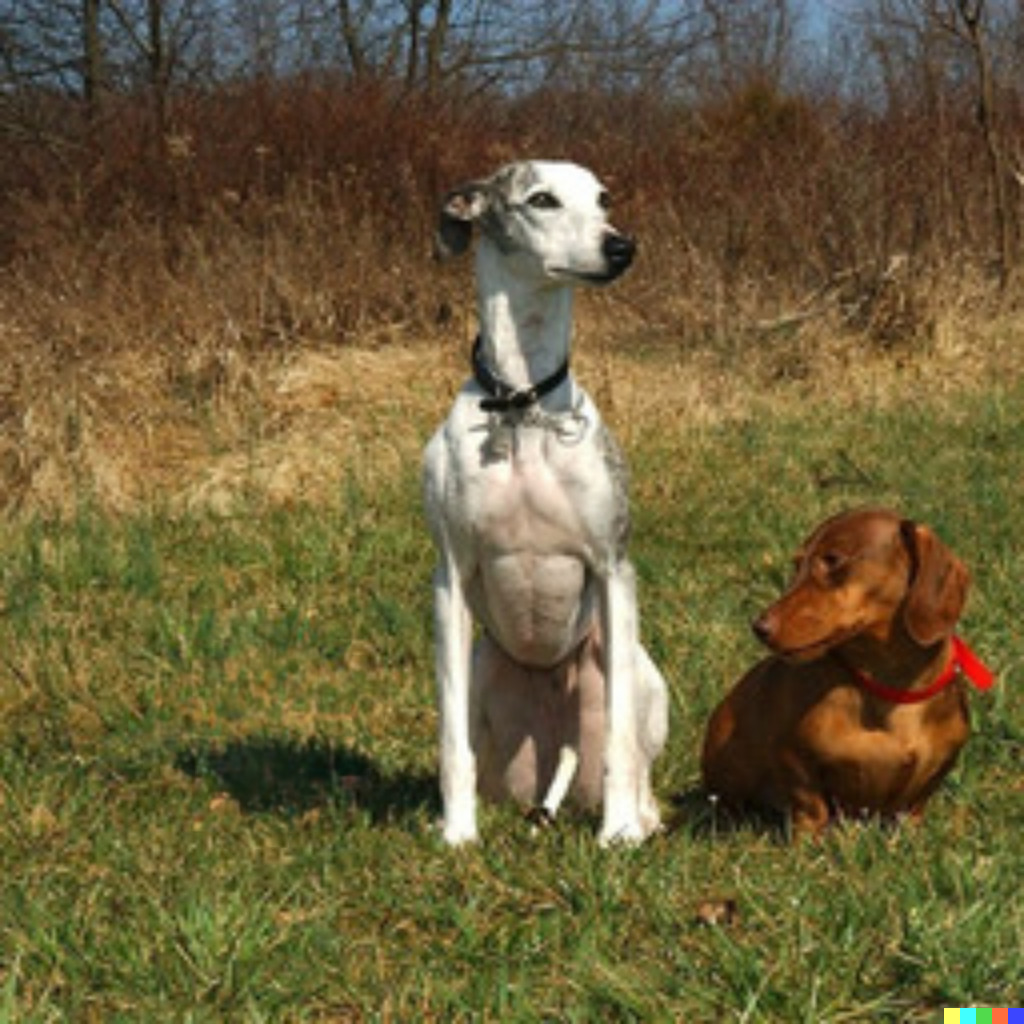} & 
        \includegraphics[width=\ww,frame]{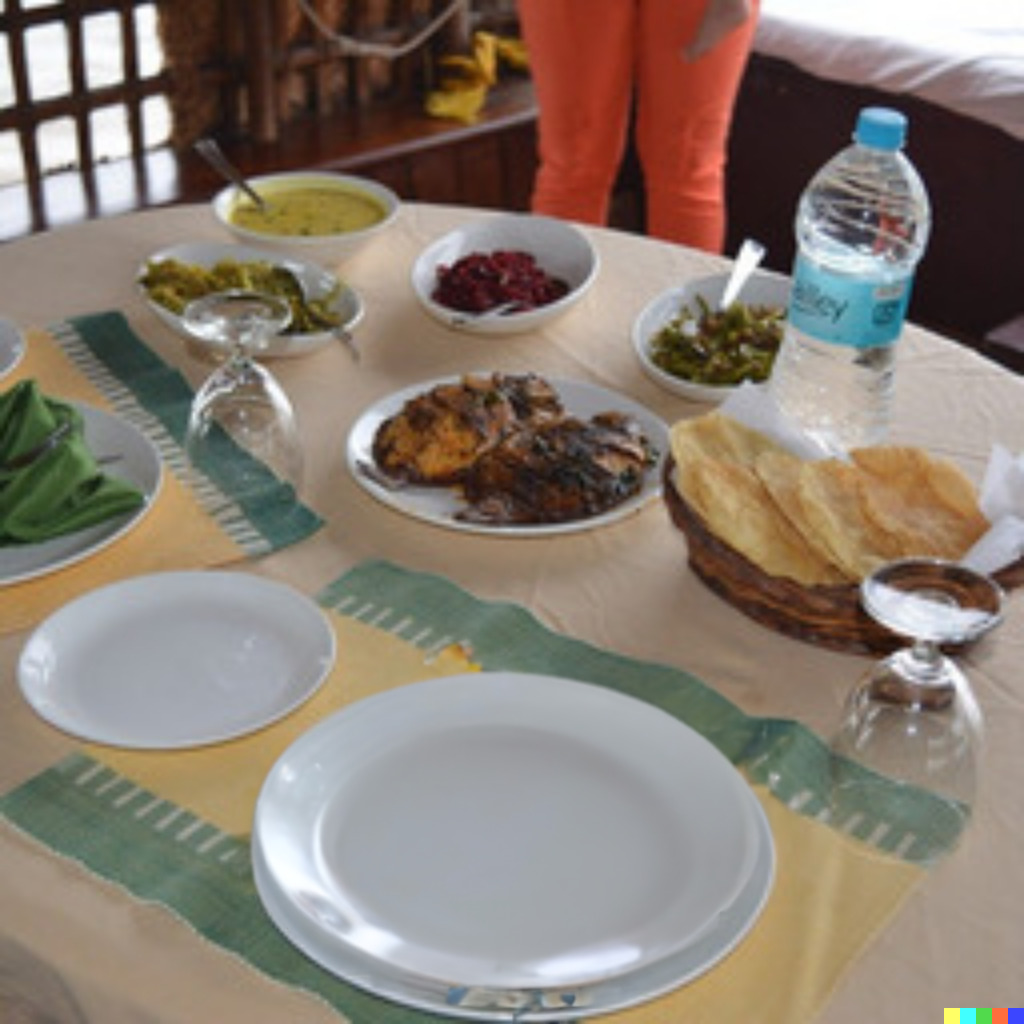} &
        \includegraphics[width=\ww,frame]{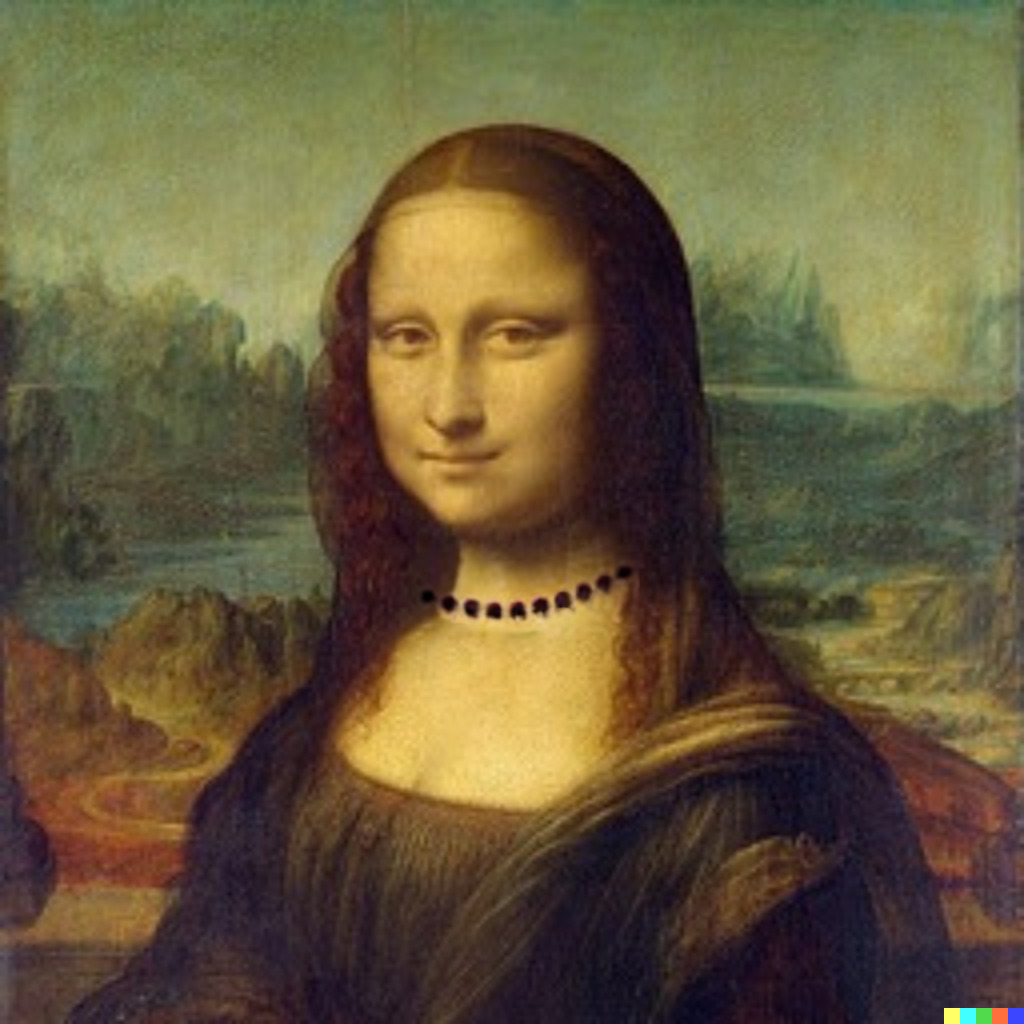} &
        \includegraphics[width=\ww,frame]{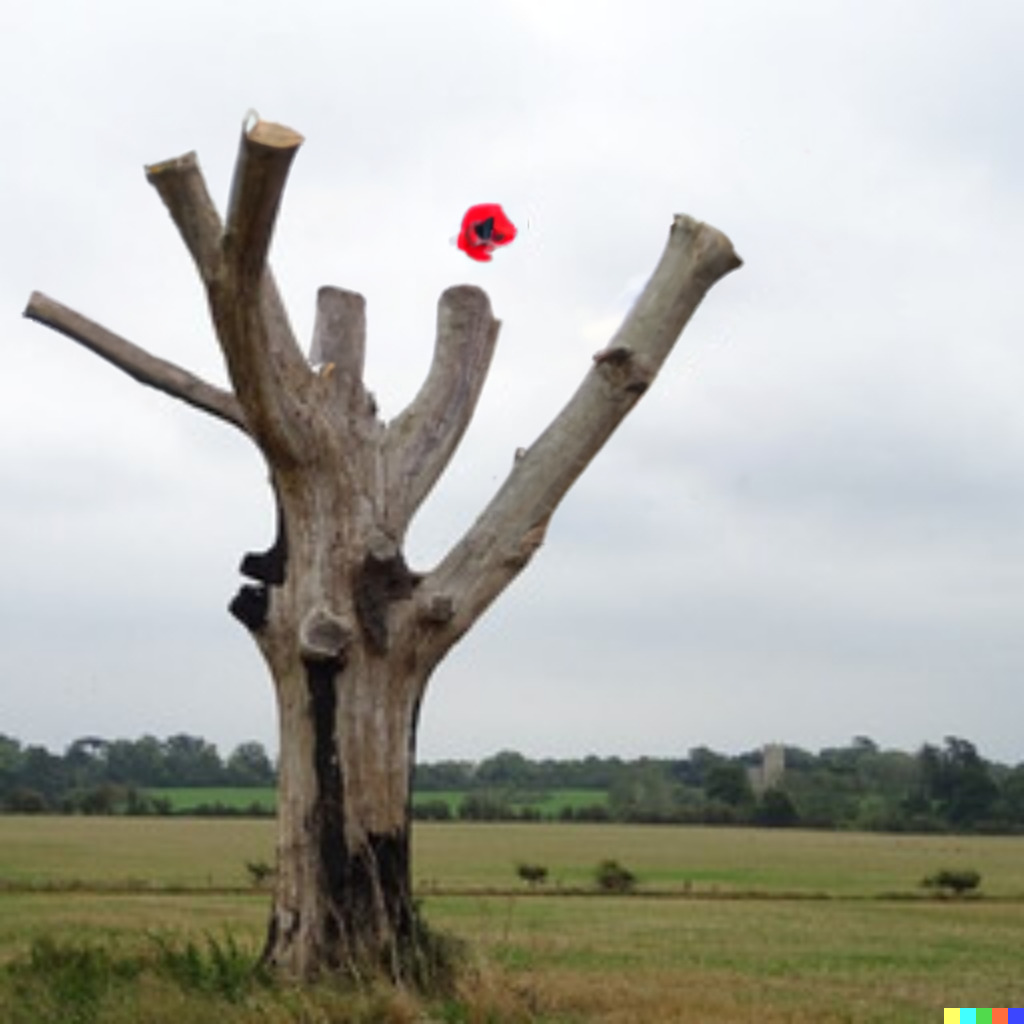} &
        \includegraphics[width=\ww,frame]{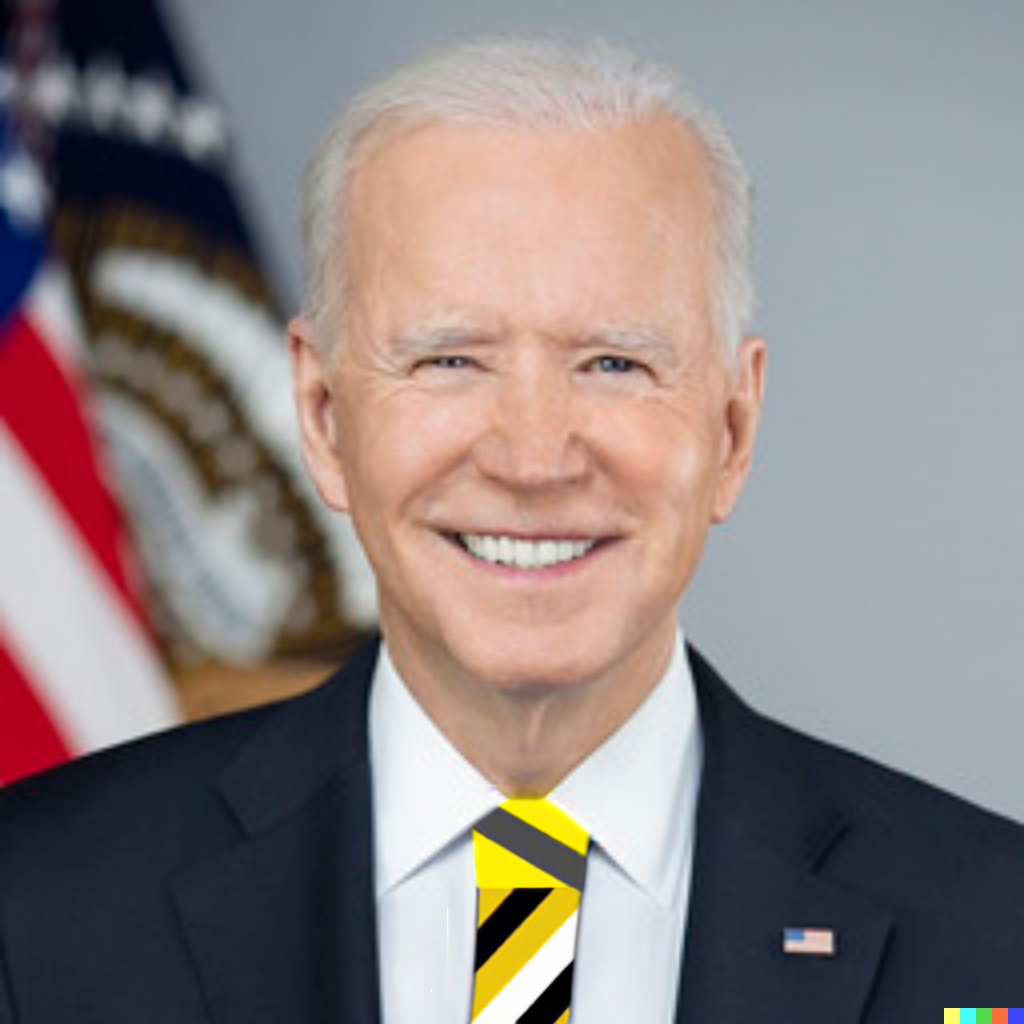} &
        \includegraphics[width=\ww,frame]{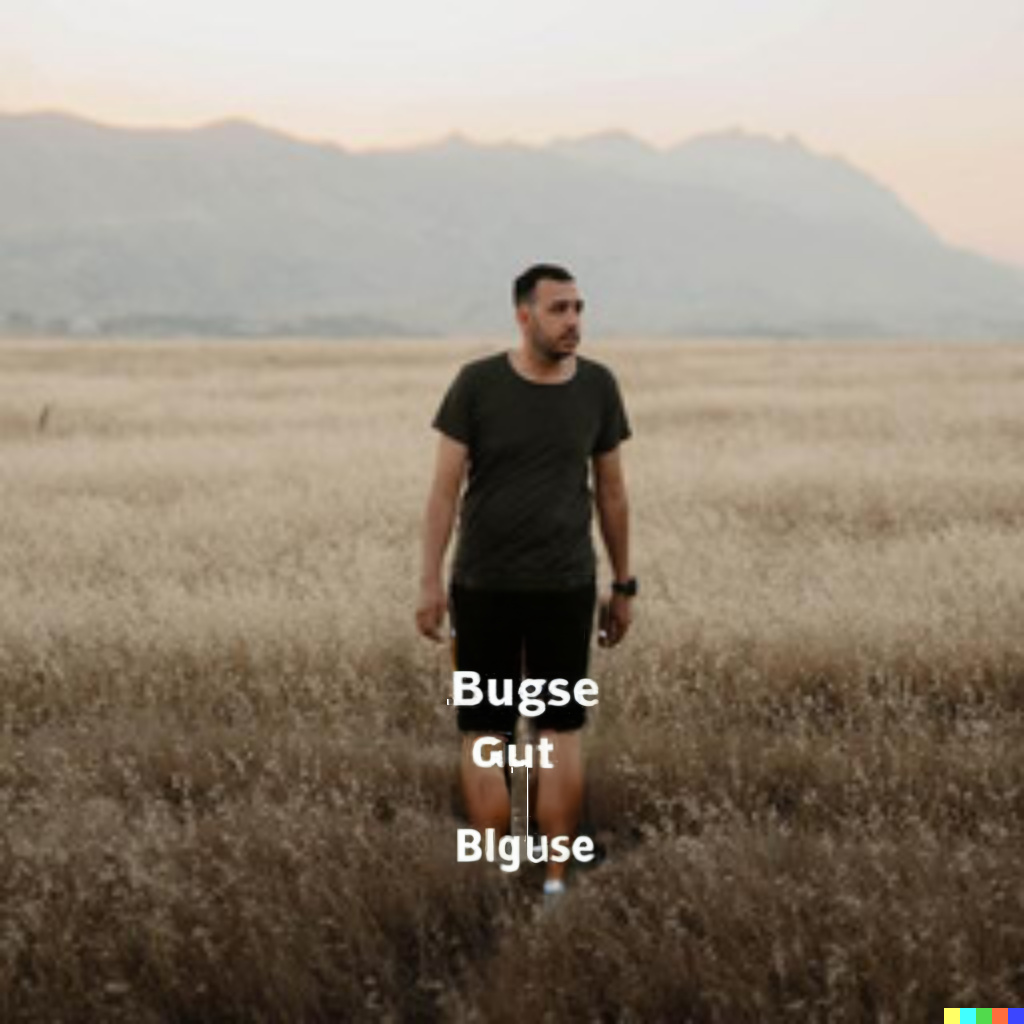} \\

        \rotatebox{90}{\phantom{A}} &
        \rotatebox{90}{\phantom{AAA}{Ours}} &
        \includegraphics[width=\ww,frame]{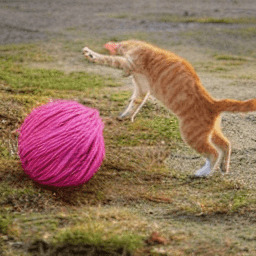} &
        \includegraphics[width=\ww,frame]{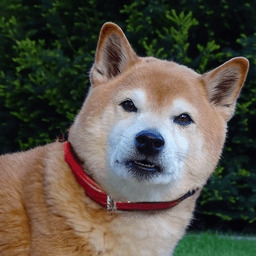} &
        \includegraphics[width=\ww,frame]{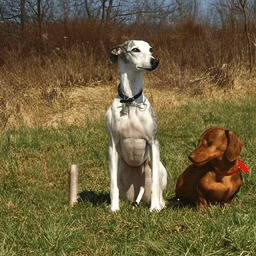} &
        \includegraphics[width=\ww,frame]{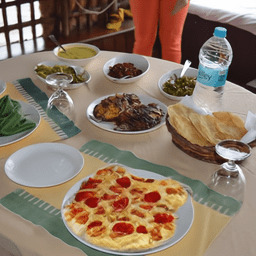} &
        \includegraphics[width=\ww,frame]{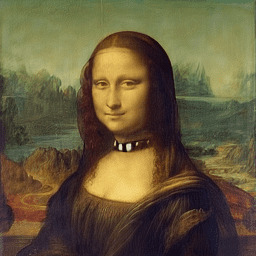} &
        \includegraphics[width=\ww,frame]{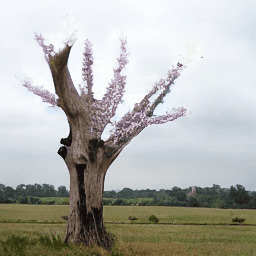} &
        \includegraphics[width=\ww,frame]{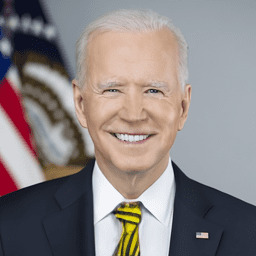} &
        \includegraphics[width=\ww,frame]{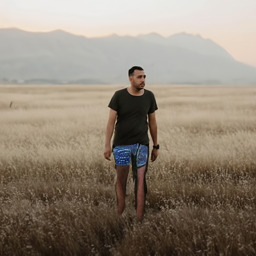}
        \\

        &&
        ``pink yarn ball'' &
        ``red dog collar'' &
        ``dog bone'' &
        ``pizza'' &
        ``golden &
        ``blooming tree'' &
        ``tie with black &
        ``blue short
         \\
        
        &&&&&& necklace''
        && and yellow 
        & pants''
        \\
        
        &&&&&&&& stripes'' & \\
    \end{tabular}
    \vspace{-3mm}
    \caption{\textbf{Comparison to baselines:} A comparison with Local CLIP-guided diffusion \cite{clip_guided_diffusion}, $\textit{PaintByWord++}$ \cite{bau2021paint, crowson2022vqgan}, Blended Diffusion \cite{avrahami2022blended}, GLIDE \cite{nichol2021glide}, GLIDE-masked \cite{nichol2021glide}, GLIDE-filtered \cite{nichol2021glide} and \DALLE~2 \cite{ramesh2022hierarchical}.}
    \label{fig:baselines_comparison}
    \vspace{-1.5em}
\end{figure*}

%% file: figures/experiment_results/fig_min.tex
\begin{figure*}[h]
    \centering
    \setlength{\tabcolsep}{0.5pt}
    \renewcommand{\arraystretch}{0.5}
    \setlength{\ww}{0.285\columnwidth}
  
    \begin{tabular}{cccccccc}
        \rotatebox{90}{\phantom{AAAa}``iPod''} &
        \includegraphics[width=\ww,frame]{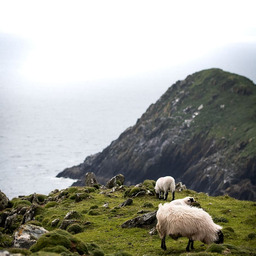} &
        \includegraphics[width=\ww,frame]{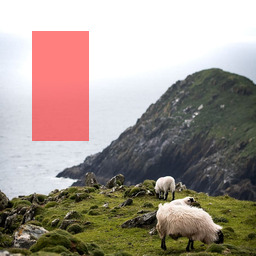} &
        \includegraphics[width=\ww,frame]{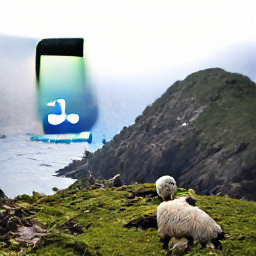} &
        \includegraphics[width=\ww,frame]{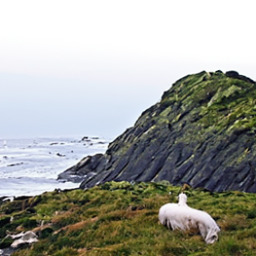} &
        \includegraphics[width=\ww,frame]{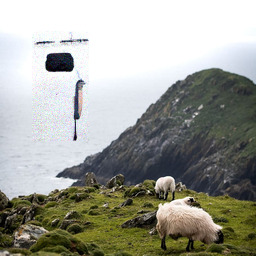} &
        \includegraphics[width=\ww,frame]{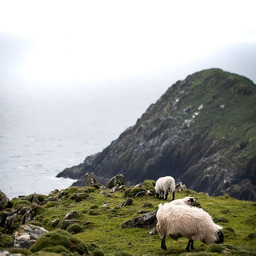} &
        \includegraphics[width=\ww,frame]{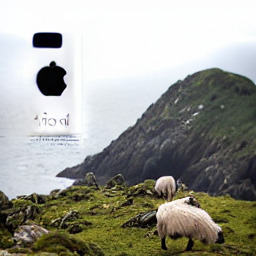}
        \\

        \rotatebox{90}{\phantom{AAa}``hourglass''} &
        \includegraphics[width=\ww,frame]{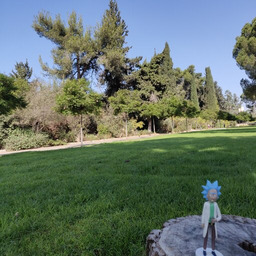} &
        \includegraphics[width=\ww,frame]{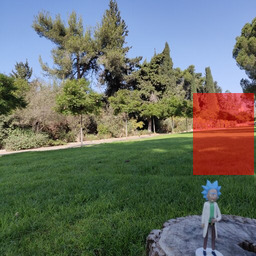} &
        \includegraphics[width=\ww,frame]{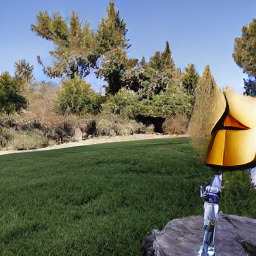} &
        \includegraphics[width=\ww,frame]{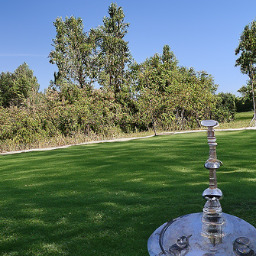} &
        \includegraphics[width=\ww,frame]{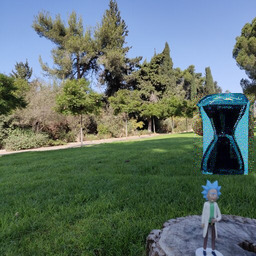} &
        \includegraphics[width=\ww,frame]{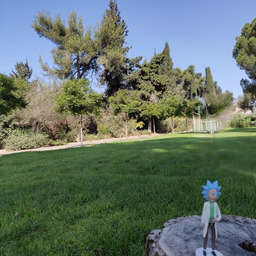} &
        \includegraphics[width=\ww,frame]{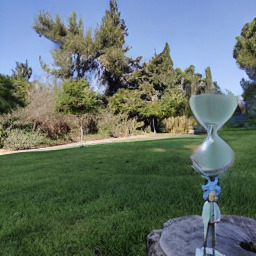}
        \\

        &
        Input image &
        Input mask &
        PaintByWord++ &
        Local CLIP-guided &
        Blended Diffusion &
        GLIDE-filtered &
        Ours
        \\

        &&&&
        diffusion
        \\
    \end{tabular}
    
    \caption{\textbf{Precision \& Diversity Experiment:} Two examples of random input images and masks, and the corresponding results of different methods, used in our quantitative evaluation. GLIDE-filtered \cite{nichol2021glide} typically fails to modify the image according to the guiding text prompt; hence, the typical result looks similar to the input image, and therefore looks natural. For more examples please refer to the supplementary material.}
    \label{fig:experiment_results_min_example}
\end{figure*}

%% file: tables/metrics_comparison.tex
\begin{table}
    \centering
    \caption{\textbf{Quantitative comparison.} In terms of precision, our method outperforms the baselines, both at the batch level and at the best result level. In terms of diversity, only the Local CLIP-guided diffusion baseline achieves a better score, due to its tendency to change the entire image significantly (lack of background preservation). The two rightmost columns report the percentage of human evaluators that preferred our method over the baseline. Our method outperforms the baselines in terms of visual quality and text matching except the visual quality of GLIDE-filtered, which mostly leaves the input untouched.}
    \begin{adjustbox}{width=1\columnwidth}
        \begin{tabular}{>{\columncolor[gray]{0.95}}lccc|cc}
            \toprule
            
            \textbf{Method} & 
            Batch &
            Batch & 
            Best Result &
            Human &
            Human
            \\
            
            & 
            Precision $\uparrow$ &
            Diversity $\uparrow$ & 
            Precision $\uparrow$ &
            Vis.~Quality &
            Text Matching
            \\
            
            \midrule

            Blended Diffusion& 
            10.4\% &
            0.106 & 
            36\% &
            64\% &
            55\%
            \\
            
            Local CLIP-guided diffusion & 
            10.49\% &
            \textbf{0.419} & 
            38\% &
            74\% &
            62\%
            \\
            
            PaintByWord++ & 
            - &
            - & 
            0\% &
            94\% &
            68\%
            \\

            GLIDE-filtered & 
            1.87\% &
            0.114 & 
            4\% &
            26\% &
            86\%
            \\
            
            Ours & 
            \textbf{28.66\%} &
            0.115 & 
            \textbf{54\%} &
            - &
            -
            \\
            
            \bottomrule
        \end{tabular}
    \end{adjustbox}
    \label{tab:metrics_comparison}
\end{table}

%% file: tables/inference_time_comparison.tex
\begin{table}
    \centering
    \caption{\textbf{Inference time comparison:} Our method outperforms all other methods when using batch processing. This stems from the fact that we perform diffusion in the latent space, and because our background preservation optimization is only required for the top-ranked result. 
    Batch sizes marked with $*$ are below the size recommended by the respective authors (lower batch precision), but are reported for comparison purposes.}
    \begin{adjustbox}{width=1.0\columnwidth}
        \begin{tabular}{>{\columncolor[gray]{0.95}}l>{\columncolor[gray]{0.95}}lccc}
            \toprule
            
            \textbf{Method} & 
            Batch &
            Single Image & 
            Full Batch &
            Per Image in
            \\
            
            & 
            Size &
            (sec) $\downarrow$ & 
            (sec) $\downarrow$ &
            Batch (sec) $\downarrow$
            \\
            
            \midrule

            Blended Diffusion& 
            64 &
            27 & 
            1472 &
            23
            \\

            Blended Diffusion & 
            $24^*$ &
            27 & 
            552 &
            23
            \\
            
            Local CLIP-guided diffusion & 
            64 &
            27 & 
            1472 &
            23
            \\

            Local CLIP-guided diffusion & 
            $24^*$ &
            27 & 
            552 &
            23
            \\
            
            PaintByWord++ & 
            - &
            78 & 
            - &
            -
            \\

            GLIDE-filtered & 
            24 &
            7 & 
            89 &
            3.7
            \\

            Ours (without background opt.) & 
            24 &
            6
            & 
            53 &
            \textbf{2.2}
            \\
            
            Ours (with background opt.) & 
            24 &
            25
            & 
            72 &
            \textbf{3}
            \\
            
            \bottomrule
        \end{tabular}
    \end{adjustbox}
    \label{tab:inference_time_comparison}
\end{table}
    

%% file: sections/limitations.tex
\section{Limitations \& Conclusions}
\label{sec:limitations}

\input{figures/limitations/fig.tex}
\input{figures/sensitivity_analysis/mini_fig.tex}

Although our method is significantly faster than prior works, it still takes over a minute on an A10 GPU to generate a ranked batch of predictions, due to the diffusion process. This limits the applicability of our method on lower-end devices. Hence, accelerating the inference time further is still an important research avenue.  

As in Blended Diffusion, the CLIP-based ranking only takes into account the generated masked area. Without a more holistic view of the image, this ranking ignores the overall realism of the output image, which may result in images where each area is realistic, but the image does not look realistic overall, e.g., \Cref{fig:limitations}(top). Thus, a better ranking system would prove useful.

Furthermore, we observe that LDM's amazing ability to generate texts is a double-edged sword: the guiding text may be interpreted by the model as a text generation task. For example, \Cref{fig:limitations}(bottom) demonstrates that instead of generating a big mountain, the model tries to generate a movie poster named ``big mountain''.

In addition, we found our method to be somewhat sensitive to its inputs. \Cref{fig:mini_sensitivity_analysis} demonstrates that small changes to the input prompt, to the input mask, or to the input image may result in small output changes. For more examples and details, please read Section D in the supplementary material.

Even without solving the aforementioned open problems, we have shown that our system can be used to locally edit images using text. Our results are realistic enough for real-world editing scenarios, and we are excited to see what users will create with the source code that we will release upon publication.

%% file: figures/limitations/fig.tex
\begin{figure}[t]
    \centering
    \setlength{\tabcolsep}{0.5pt}
    \renewcommand{\arraystretch}{0.5}
    \setlength{\ww}{0.182\columnwidth}
  
    \begin{tabular}{ccccccc}
        \rotatebox{90}{\phantom{a}\scriptsize{``a man with a}} &
        \rotatebox{90}{\phantom{aa}\scriptsize{red suit''}} &
        \includegraphics[width=\ww,frame]{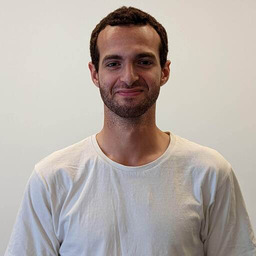} &
        \includegraphics[width=\ww,frame]{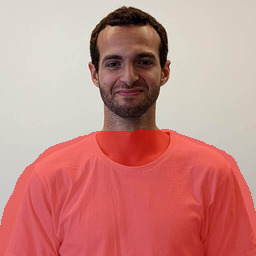} &
        \includegraphics[width=\ww,frame]{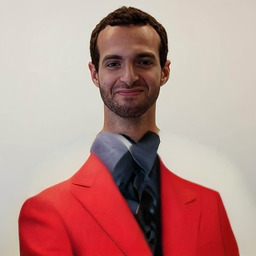} &
        \includegraphics[width=\ww,frame]{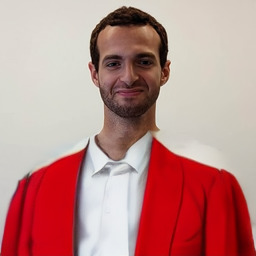} &
        \includegraphics[width=\ww,frame]{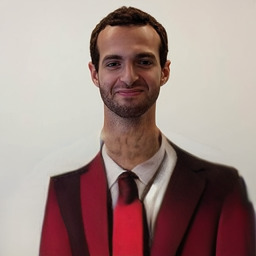}
        \\
        
        \rotatebox{90}{\phantom{a}} &
        \rotatebox{90}{\phantom{a}\scriptsize{``big mountain''}} &
        \includegraphics[width=\ww,frame]{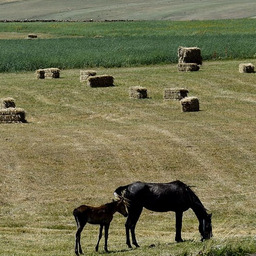} &
        \includegraphics[width=\ww,frame]{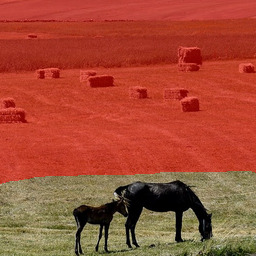} &
        \includegraphics[width=\ww,frame]{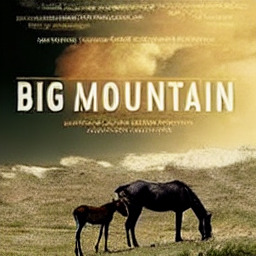} &
        \includegraphics[width=\ww,frame]{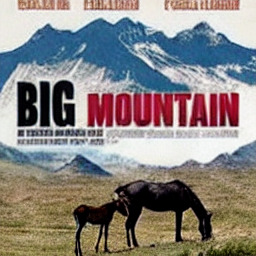} &
        \includegraphics[width=\ww,frame]{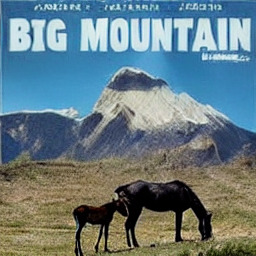}
        \\

        &&
        \scriptsize{Input image} &
        \scriptsize{Input mask}
        \\
    \end{tabular}
    \caption{\textbf{Limitations:} Top row: our CLIP-based ranking takes into account only the masked area, Thus, the results are sometimes only piece-wise realistic, and the image does not look realistic as a whole. Bottom row: the model has a text bias - it may try to create movie posters/book covers with text instead or in addition to generating the actual object.}
    \label{fig:limitations}
\end{figure}

%% file: figures/sensitivity_analysis/mini_fig.tex
\begin{figure}[ht]
    \centering
    \setlength{\tabcolsep}{0.5pt}
    \renewcommand{\arraystretch}{0.5}
    \setlength{\ww}{0.24\columnwidth}
    \begin{tabular}{ccccccc}
        \rotatebox[origin=c]{90}{Prompt sen.} &
        \includegraphics[valign=c, width=\ww,frame]{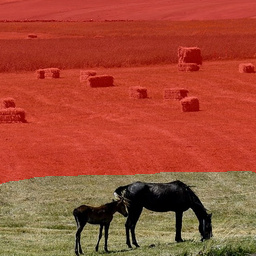} &
        \includegraphics[valign=c, width=\ww,frame]{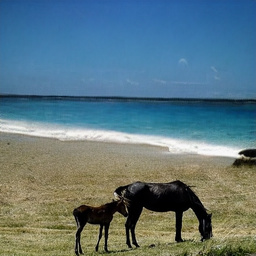} &
        \includegraphics[valign=c, width=\ww,frame]{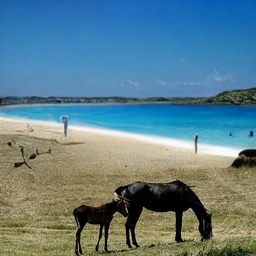} &
        \includegraphics[valign=c, width=\ww,frame]{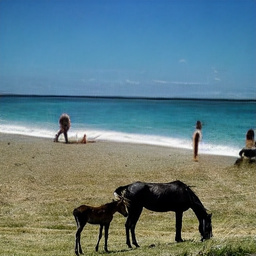}
        \\

        &
        input + mask &
        ``a beach!'' &
        ``beaches'' &
        ``a day at
        \\

        &&&&
        the beach''
        \\
        \\

        \rotatebox[origin=c]{90}{Mask sen.} &
        \includegraphics[valign=c, width=\ww,frame]{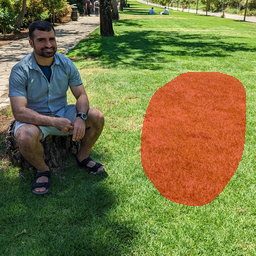} &
        \includegraphics[valign=c, width=\ww,frame]{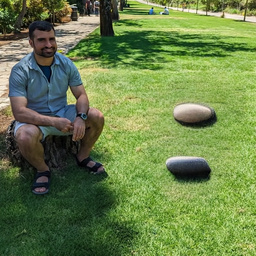} 
        &
        \includegraphics[valign=c, width=\ww,frame]{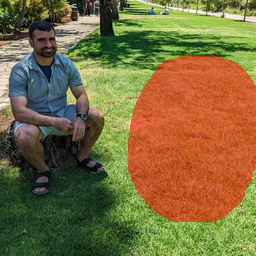} &
        \includegraphics[valign=c, width=\ww,frame]{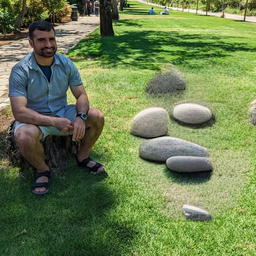}
        \\

        &
        small mask &
        ``stones'' &
        big mask &
        ``stones''
        \\
        \\

        \rotatebox[origin=c]{90}{Image sen.} &
        \includegraphics[valign=c, width=\ww,frame]{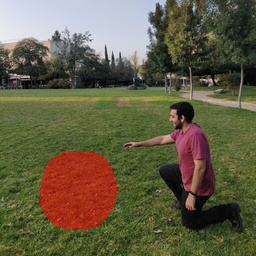} &
        \includegraphics[valign=c, width=\ww,frame]{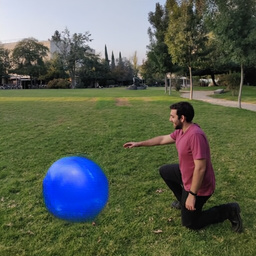} 
        &
        \includegraphics[valign=c, width=\ww,frame]{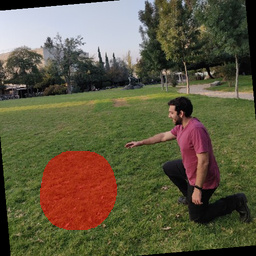} &
        \includegraphics[valign=c, width=\ww,frame]{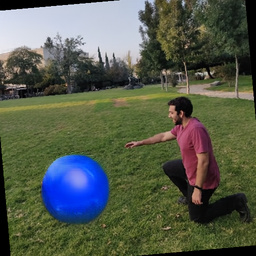} 
        \\

        &
        original image &
        ``a blue ball'' &
        rotated image &
        ``a blue ball''

    \end{tabular}
    
    \caption{\textbf{Sensitivity analysis:} we found our method to be somewhat sensitive to its inputs. Small changes to the input prompt (first row), to the input mask (second row), or to the input image (third row) may result in small output changes.}
    \label{fig:mini_sensitivity_analysis}
\end{figure}

%% file: appendices/additional_examples.tex
\section{Additional Examples}
\label{sec:additional_examples}
\input{figures/additional_applications/new_object/fig.tex}
\input{figures/additional_applications/multiple_predictions/fig.tex}
\input{figures/additional_applications/background_replacement/horses/fig.tex}

\input{figures/process_visualization/fig.tex}

\input{figures/additional_ablations/thin_masks/fig.tex}

In \Cref{fig:additional_application_new_object} we demonstrate more examples of adding a new object to a scene. In \Cref{fig:additional_multiple_predictions} we demonstrate the one-to-many generation ability of our model. In \Cref{fig:additional_application_replace_background_horses} we demonstrate more examples of background replacement. In addition, in \Cref{fig:process_visualization} we provide a visualization of the diffusion process on several examples.

\subsection{Interactive Editing}
\input{figures/editing_session/birka/fig.tex}

Because of the near-perfect background preservation of our method, the user is able to perform an interactive editing session: editing the image gradually, where at each stage of the editing session the user edits a different area within the image without changing the other parts of the image that were already edited. We show an interactive editing session in \Cref{fig:editing_session_birka}.

%% file: figures/additional_applications/new_object/fig.tex
\begin{figure*}[h]
    \centering
    \setlength{\tabcolsep}{0.5pt}
    \renewcommand{\arraystretch}{0.5}
    \setlength{\ww}{0.33\columnwidth}
  
    \begin{tabular}{cccccc}
        \includegraphics[width=\ww,frame]{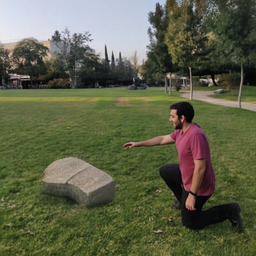} &
        \includegraphics[width=\ww,frame]{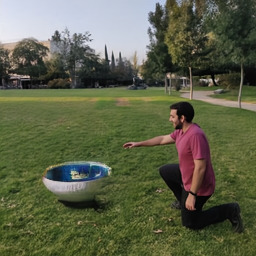} &
        \includegraphics[width=\ww,frame]{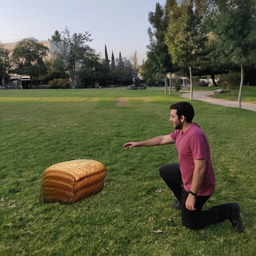} &
        \includegraphics[width=\ww,frame]{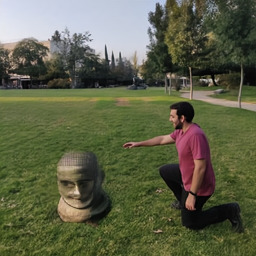} &
        \includegraphics[width=\ww,frame]{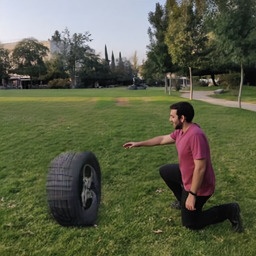} &
        \includegraphics[width=\ww,frame]{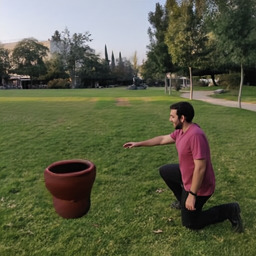}
        \\
        ``big stone'' &
        ``bowl of water'' &
        ``bread'' &
        ``Buddha'' &
        ``car tire'' &
        ``clay pot''
        \\
        \\

        \includegraphics[width=\ww,frame]{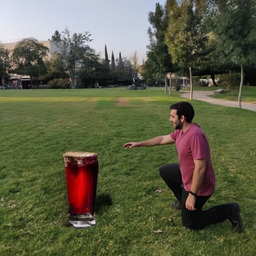} &
        \includegraphics[width=\ww,frame]{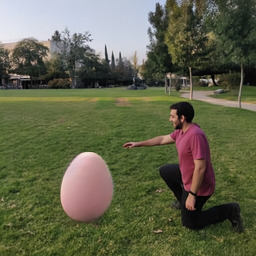} &
        \includegraphics[width=\ww,frame]{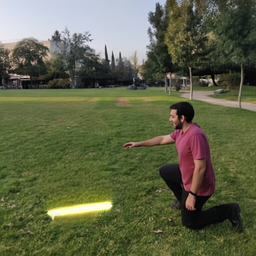} &
        \includegraphics[width=\ww,frame]{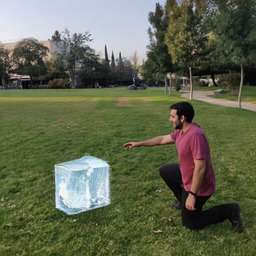} &
        \includegraphics[width=\ww,frame]{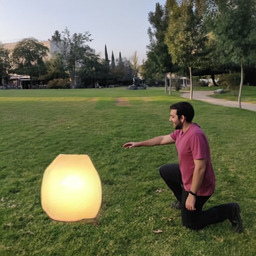} &
        \includegraphics[width=\ww,frame]{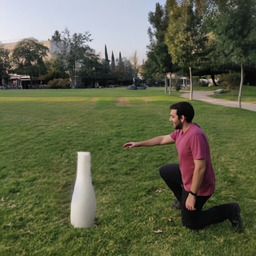}
        \\
        ``cola'' &
        ``egg'' &
        ``glow stick'' &
        ``ice cube'' &
        ``lamp'' &
        ``milk''
        \\
        \\

        \includegraphics[width=\ww,frame]{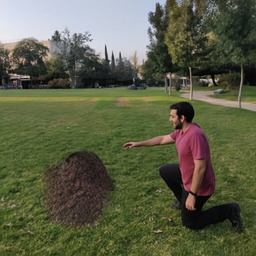} &
        \includegraphics[width=\ww,frame]{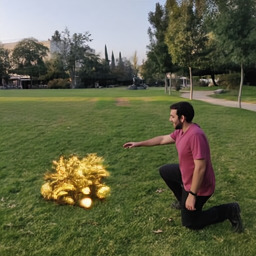} &
        \includegraphics[width=\ww,frame]{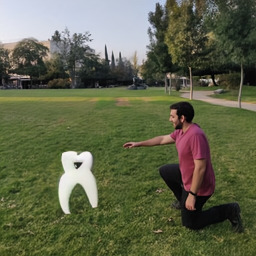} &
        \includegraphics[width=\ww,frame]{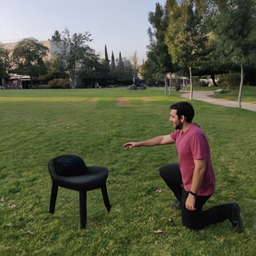} &
        \includegraphics[width=\ww,frame]{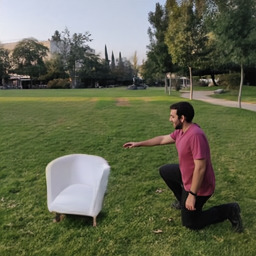} &
        \includegraphics[width=\ww,frame]{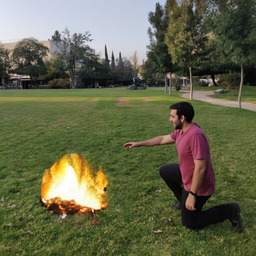}
        \\
        ``pile of dirt'' &
        ``pile of gold'' &
        ``tooth'' &
        ``black chair'' &
        ``white chair'' &
        ``bonfire''
        \\
        \\

        \includegraphics[width=\ww,frame]{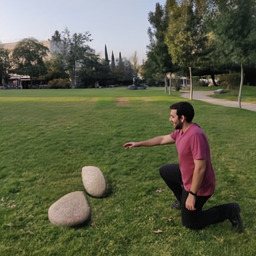} &
        \includegraphics[width=\ww,frame]{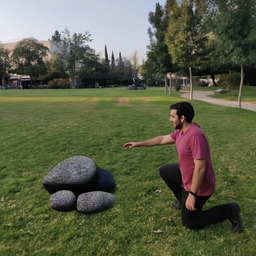} &
        \includegraphics[width=\ww,frame]{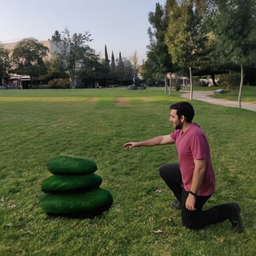} &
        \includegraphics[width=\ww,frame]{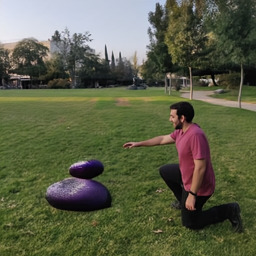} &
        \includegraphics[width=\ww,frame]{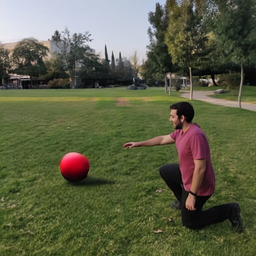} &
        \includegraphics[width=\ww,frame]{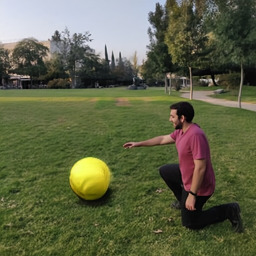}
        \\

        ``stones'' &
        ``black stones'' &
        ``green stones'' &
        ``purple stones'' &
        ``red ball'' &
        ``yellow ball''
        \\
        \\

        \includegraphics[width=\ww,frame]{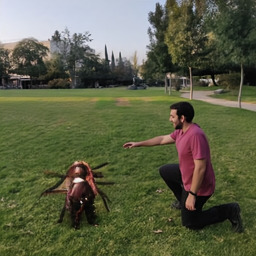} &
        \includegraphics[width=\ww,frame]{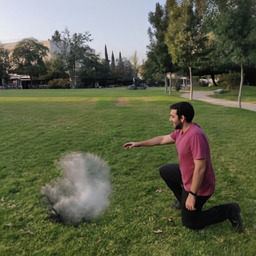} &
        \includegraphics[width=\ww,frame]{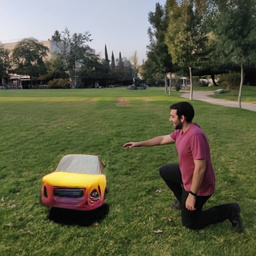} &
        \includegraphics[width=\ww,frame]{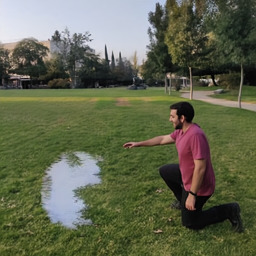} &
        \includegraphics[width=\ww,frame]{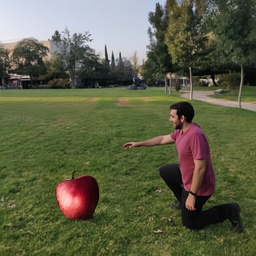} &
        \includegraphics[width=\ww,frame]{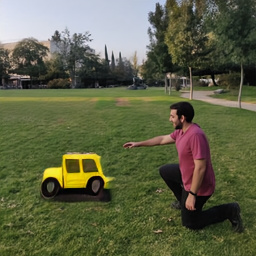}
        \\

        ``huge ant'' &
        ``smoke'' &
        ``toy car'' &
        ``water puddle'' &
        ``huge apple'' &
        ``yellow toy truck''
        \\
    \end{tabular}
    
    \caption{\textbf{Adding a new object:} Additional examples for adding a new object within a scene.}
    \label{fig:additional_application_new_object}
\end{figure*}

%% file: figures/additional_applications/multiple_predictions/fig.tex
\begin{figure*}[h]
    \centering
    \setlength{\tabcolsep}{0.5pt}
    \renewcommand{\arraystretch}{0.5}
    \setlength{\ww}{0.33\columnwidth}
  
    \begin{tabular}{ccccccc}
        \rotatebox{90}{\phantom{``N''}}
        \rotatebox{90}{\phantom{AAa} text ``GAN''} &
        \includegraphics[width=\ww,frame]{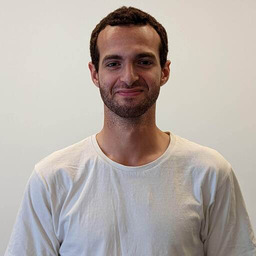} &
        \includegraphics[width=\ww,frame]{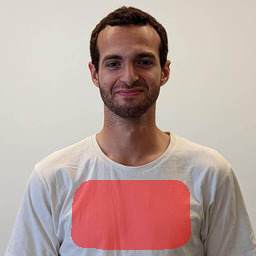} &
        \includegraphics[width=\ww,frame]{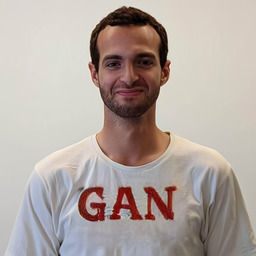} &
        \includegraphics[width=\ww,frame]{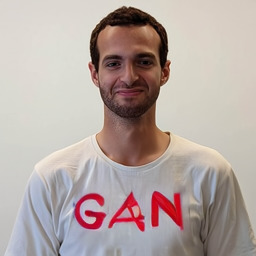} &
        \includegraphics[width=\ww,frame]{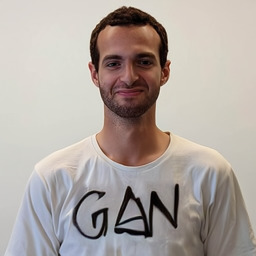} &
        \includegraphics[width=\ww,frame]{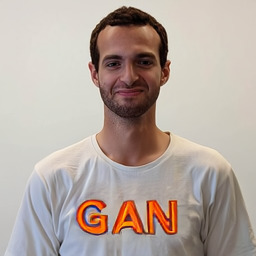}
        \\

        \rotatebox{90}{graffiti with the text}
        \rotatebox{90}{``No Free Lunch''} &
        \includegraphics[width=\ww,frame]{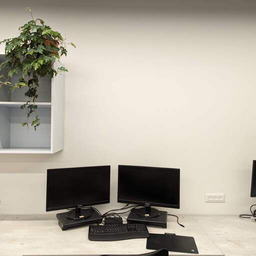} &
        \includegraphics[width=\ww,frame]{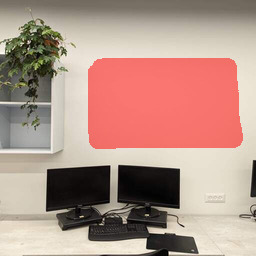} &
        \includegraphics[width=\ww,frame]{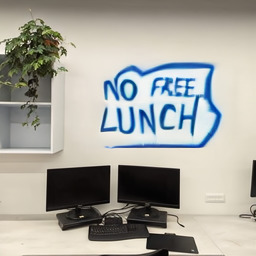} &
        \includegraphics[width=\ww,frame]{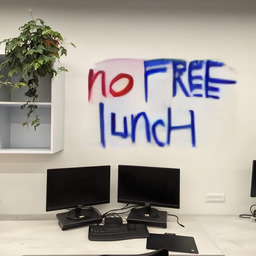} &
        \includegraphics[width=\ww,frame]{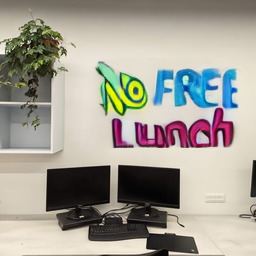} &
        \includegraphics[width=\ww,frame]{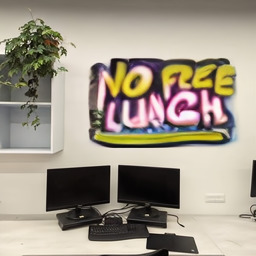}
        \\

        \rotatebox{90}{\phantom{``N''}}
        \rotatebox{90}{\phantom{A} ``big mountain''} &
        \includegraphics[width=\ww,frame]{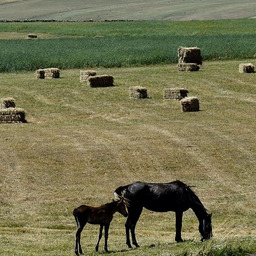} &
        \includegraphics[width=\ww,frame]{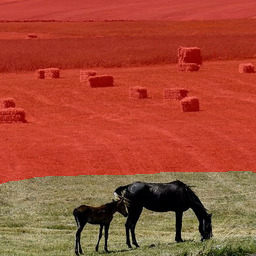} &
        \includegraphics[width=\ww,frame]{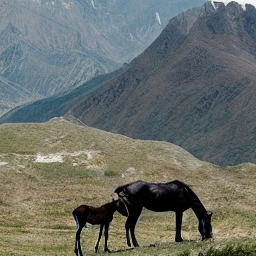} &
        \includegraphics[width=\ww,frame]{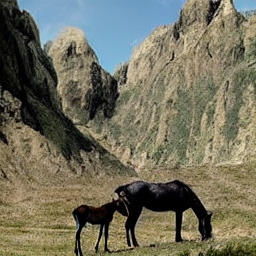} &
        \includegraphics[width=\ww,frame]{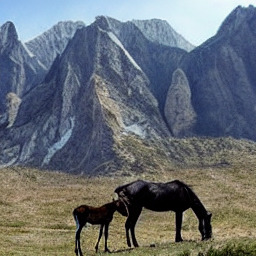} &
        \includegraphics[width=\ww,frame]{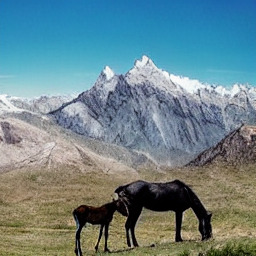}
        \\

        \rotatebox{90}{\phantom{``N''}}
        \rotatebox{90}{\phantom{A} ``colorful balls''} &
        \includegraphics[width=\ww,frame]{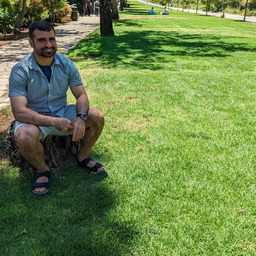} &
        \includegraphics[width=\ww,frame]{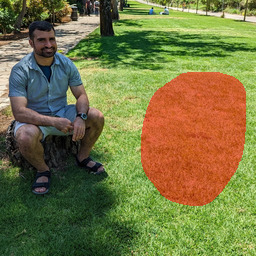} &
        \includegraphics[width=\ww,frame]{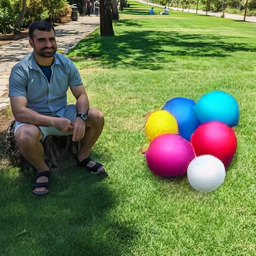} &
        \includegraphics[width=\ww,frame]{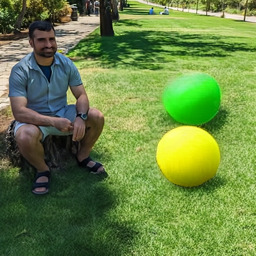} &
        \includegraphics[width=\ww,frame]{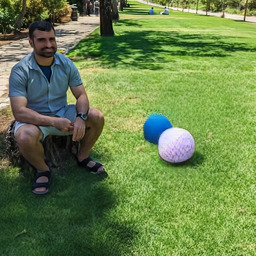} &
        \includegraphics[width=\ww,frame]{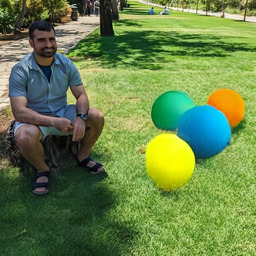}
        \\

        \rotatebox{90}{\phantom{``N''}}
        \rotatebox{90}{\phantom{AAa} ``ice cube''} &
        \includegraphics[width=\ww,frame]{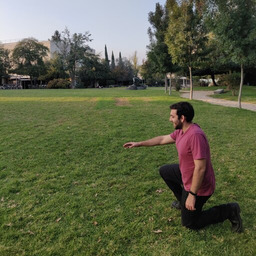} &
        \includegraphics[width=\ww,frame]{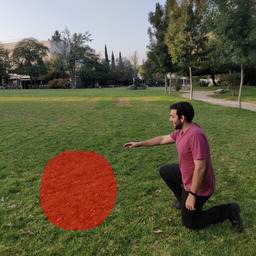} &
        \includegraphics[width=\ww,frame]{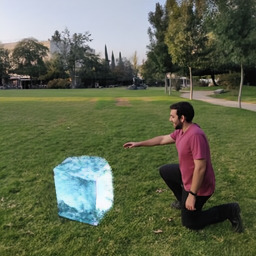} &
        \includegraphics[width=\ww,frame]{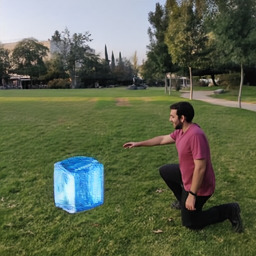} &
        \includegraphics[width=\ww,frame]{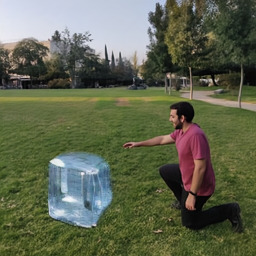} &
        \includegraphics[width=\ww,frame]{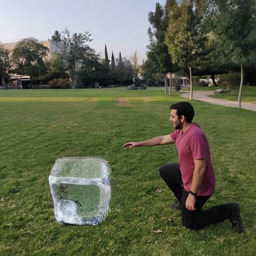}
        \\
        
        &
        Input image &
        Input mask &
        Prediction 1 &
        Prediction 2 &
        Prediction 3 &
        Prediction 4
        \\
    \end{tabular}
    
    \caption{\textbf{Multiple predictions:} Dealing with a one-to-many task, there is a need to generate multiple predictions.}
    \label{fig:additional_multiple_predictions}
\end{figure*}

%% file: figures/additional_applications/background_replacement/horses/fig.tex
\begin{figure*}[h]
    \centering
    \setlength{\tabcolsep}{0.5pt}
    \renewcommand{\arraystretch}{0.5}
    \setlength{\ww}{0.33\columnwidth}
  
    \begin{tabular}{cccccc}
        \includegraphics[width=\ww,frame]{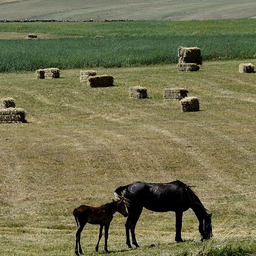} &
        \includegraphics[width=\ww,frame]{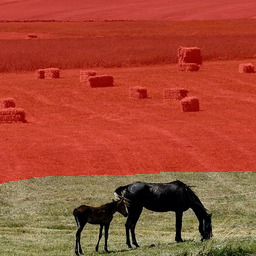} &
        \includegraphics[width=\ww,frame]{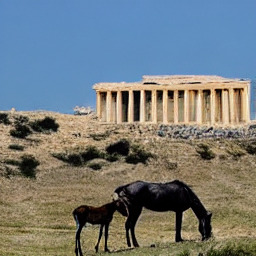} &
        \includegraphics[width=\ww,frame]{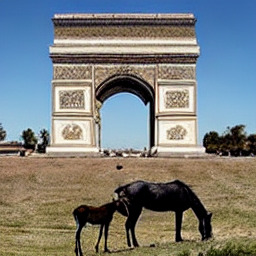} &
        \includegraphics[width=\ww,frame]{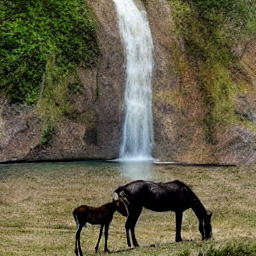} &
        \includegraphics[width=\ww,frame]{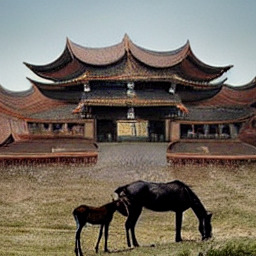}
        \\
        Input image &
        Input mask &
        ``Acropolis'' &
        ``Arc de Triomphe'' &
        ``big waterfall'' &
        ``China''
        \\
        \\

        \includegraphics[width=\ww,frame]{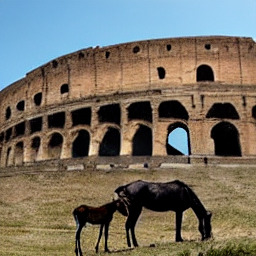} &
        \includegraphics[width=\ww,frame]{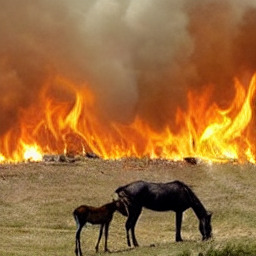} &
        \includegraphics[width=\ww,frame]{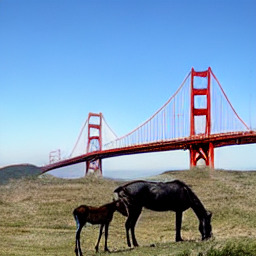} &
        \includegraphics[width=\ww,frame]{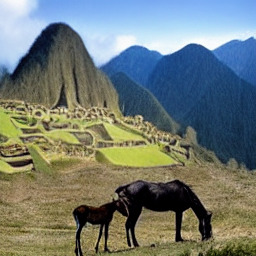} &
        \includegraphics[width=\ww,frame]{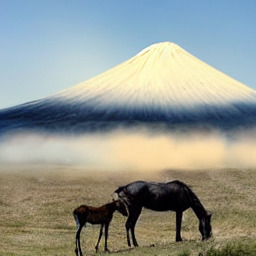} &
        \includegraphics[width=\ww,frame]{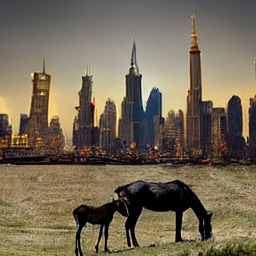}
        \\
        ``Colosseum'' &
        ``fire'' &
        ``Golden Gate Bridge'' &
        ``Machu Picchu'' &
        ``Mount Fuji'' &
        ``New York City''
        \\
        \\

        \includegraphics[width=\ww,frame]{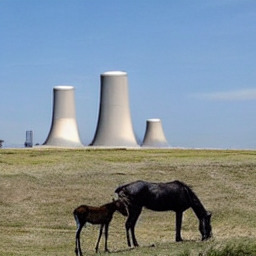} &
        \includegraphics[width=\ww,frame]{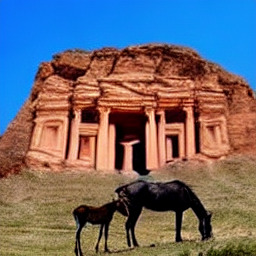} &
        \includegraphics[width=\ww,frame]{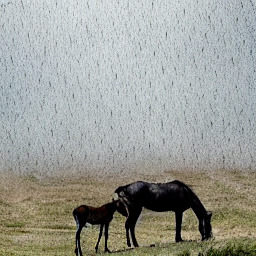} &
        \includegraphics[width=\ww,frame]{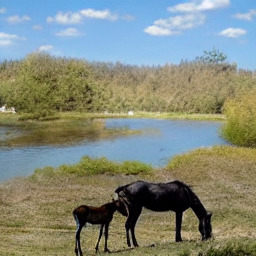} &
        \includegraphics[width=\ww,frame]{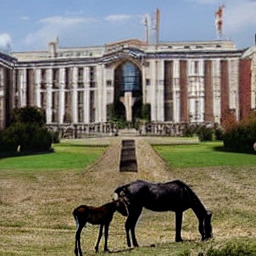} &
        \includegraphics[width=\ww,frame]{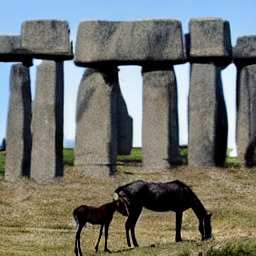}
        \\
        ``nuclear power plant'' &
        ``Petra'' &
        ``rainy'' &
        ``river'' &
        ``Stanford University'' &
        ``Stonehenge''
        \\
        \\

        \includegraphics[width=\ww,frame]{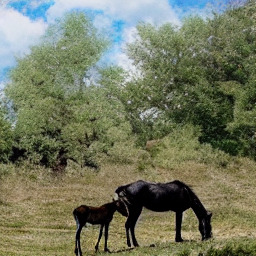} &
        \includegraphics[width=\ww,frame]{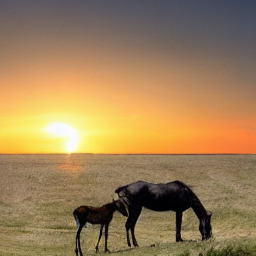} &
        \includegraphics[width=\ww,frame]{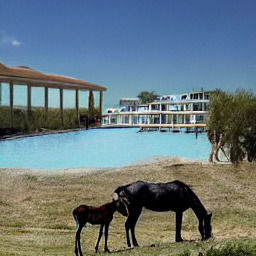} &
        \includegraphics[width=\ww,frame]{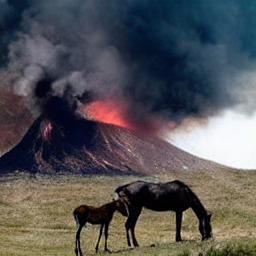} &
        \includegraphics[width=\ww,frame]{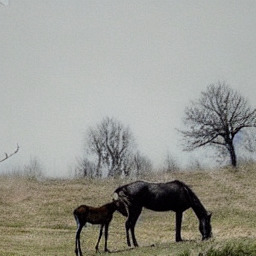} &
        \includegraphics[width=\ww,frame]{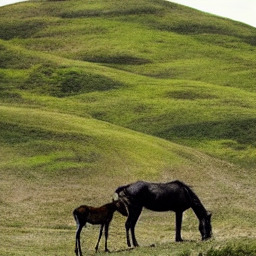}
        \\
        ``sunny'' &
        ``sunrise'' &
        ``swimming pool'' &
        ``volcanic eruption'' &
        ``winter'' &
        ``green hills''
        \\
        \\

        \includegraphics[width=\ww,frame]{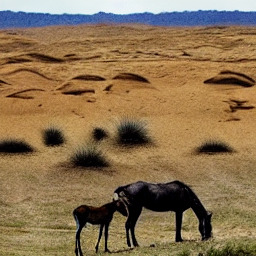} &
        \includegraphics[width=\ww,frame]{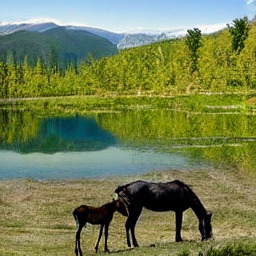} &
        \includegraphics[width=\ww,frame]{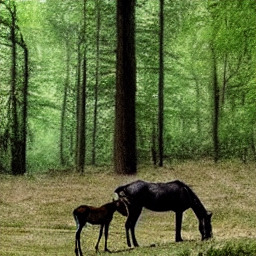} &
        \includegraphics[width=\ww,frame]{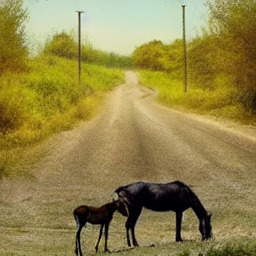} &
        \includegraphics[width=\ww,frame]{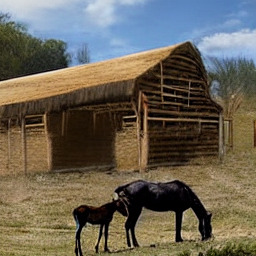} &
        \includegraphics[width=\ww,frame]{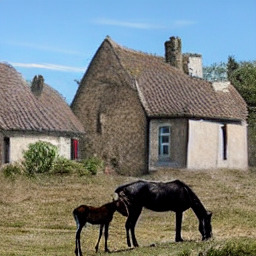}
        \\
        ``desert'' &
        ``big lake'' &
        ``forest'' &
        ``dusty road'' &
        ``horses stable'' &
        ``houses''
        \\

    \end{tabular}
    
    \caption{\textbf{Replacing the background:} Additional examples for the background replacement capability of our model.}
    \label{fig:additional_application_replace_background_horses}
\end{figure*}

%% file: figures/process_visualization/fig.tex
\begin{figure*}[ht]
    \centering
    \setlength{\tabcolsep}{0.5pt}
    \renewcommand{\arraystretch}{0.5}
    \setlength{\ww}{0.325\columnwidth}
  
    \begin{tabular}{ccccccccc}
        &
        \rotatebox{90}{\phantom{AAAA}``stones''} &
        \includegraphics[width=\ww,frame]{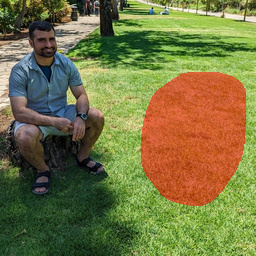} &
        \phantom{A} &
        \includegraphics[width=\ww,frame]{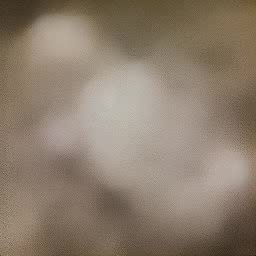} &
        \includegraphics[width=\ww,frame]{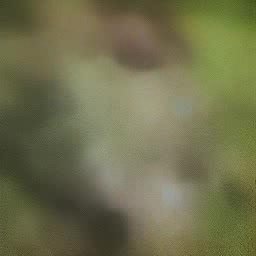} &
        \includegraphics[width=\ww,frame]{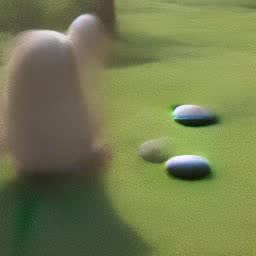} &
        \includegraphics[width=\ww,frame]{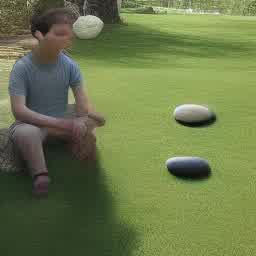} &
        \includegraphics[width=\ww,frame]{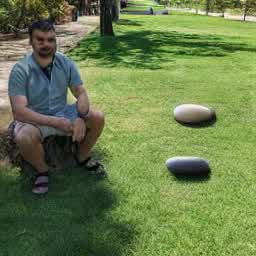}
        \\

        &
        \rotatebox{90}{\phantom{AAa}``a blue ball''} &
        \includegraphics[width=\ww,frame]{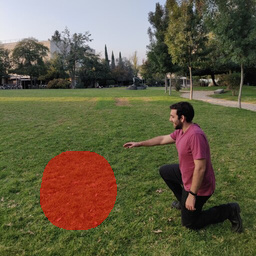} &
        \phantom{A} &
        \includegraphics[width=\ww,frame]{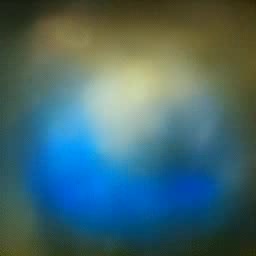} &
        \includegraphics[width=\ww,frame]{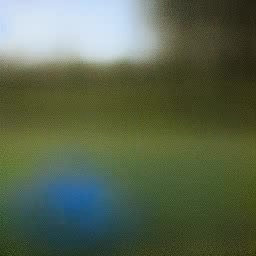} &
        \includegraphics[width=\ww,frame]{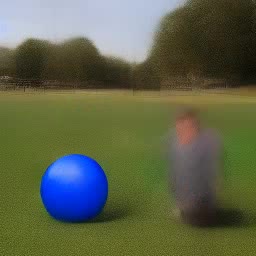} &
        \includegraphics[width=\ww,frame]{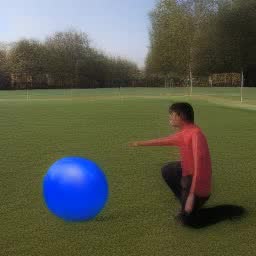} &
        \includegraphics[width=\ww,frame]{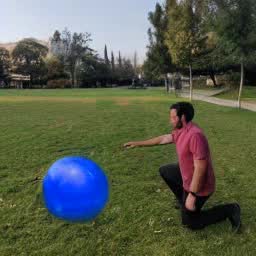}
        \\

        &
        \rotatebox{90}{\phantom{AAA}``a beach''} &
        \includegraphics[width=\ww,frame]{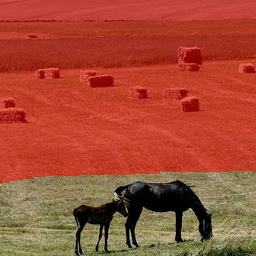} &
        \phantom{A} &
        \includegraphics[width=\ww,frame]{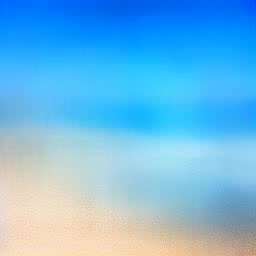} &
        \includegraphics[width=\ww,frame]{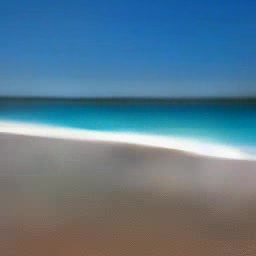} &
        \includegraphics[width=\ww,frame]{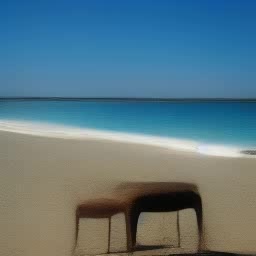} &
        \includegraphics[width=\ww,frame]{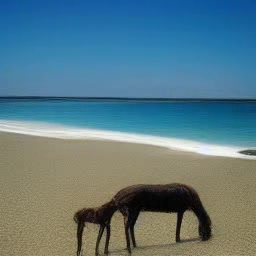} &
        \includegraphics[width=\ww,frame]{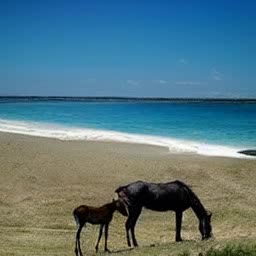}
        \\

        \rotatebox{90}{\phantom{AA}``a man with a} &
        \rotatebox{90}{\phantom{AAA}green shirt''} &
        \includegraphics[width=\ww,frame]{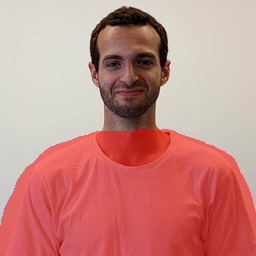} &
        \phantom{A} &
        \includegraphics[width=\ww,frame]{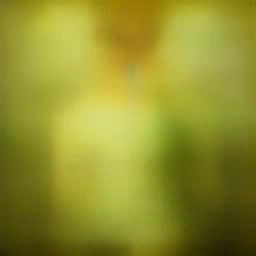} &
        \includegraphics[width=\ww,frame]{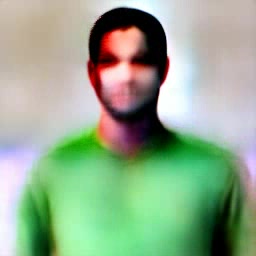} &
        \includegraphics[width=\ww,frame]{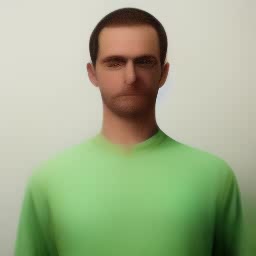} &
        \includegraphics[width=\ww,frame]{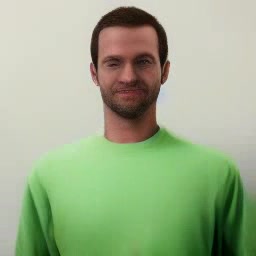} &
        \includegraphics[width=\ww,frame]{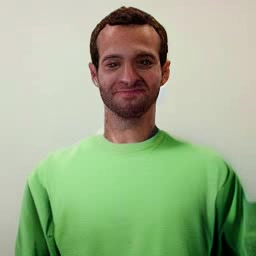}
        \\

        \rotatebox{90}{\phantom{AAA}``an avocado} &
        \rotatebox{90}{\phantom{AAAA}painting''} &
        \includegraphics[width=\ww,frame]{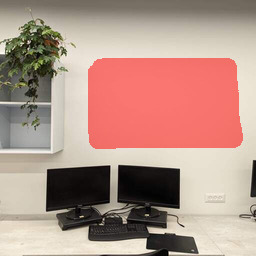} &
        \phantom{A} &
        \includegraphics[width=\ww,frame]{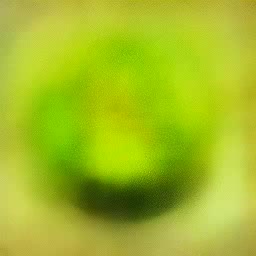} &
        \includegraphics[width=\ww,frame]{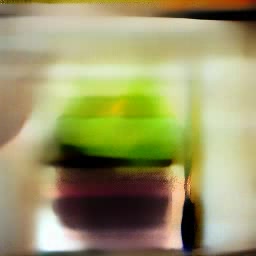} &
        \includegraphics[width=\ww,frame]{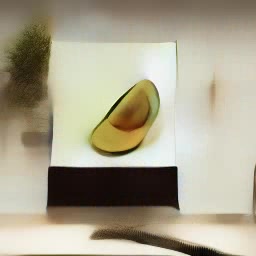} &
        \includegraphics[width=\ww,frame]{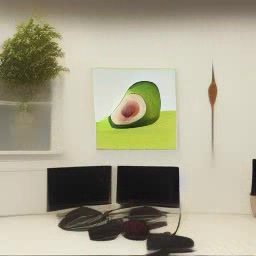} &
        \includegraphics[width=\ww,frame]{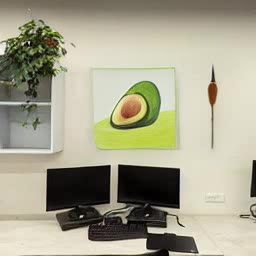}
        \\

        &
        &
        input + mask &
        &
        diffusion start &
        &
        &
        &
        diffusion end
    \end{tabular}
    
    \caption{\textbf{Process visualization:} a visualization of the diffusion process for various inputs, without the background reconstruction step that is explained in section 4.2 in the main paper.}
    \label{fig:process_visualization}
\end{figure*}

%% file: figures/additional_ablations/thin_masks/fig.tex
\begin{figure*}[h]
    \centering
    \setlength{\tabcolsep}{0.5pt}
    \renewcommand{\arraystretch}{0.5}
    \setlength{\ww}{0.5\columnwidth}
  
    \begin{tabular}{ccccc}
        \rotatebox{90}{\phantom{AAAA} ``white clouds''} &
        \includegraphics[width=\ww,frame]{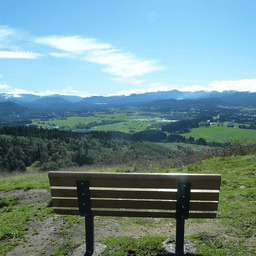} &
        \includegraphics[width=\ww,frame]{figures/thin_masks_ablation/assets/bench_mask_overlay.jpg} &
        \includegraphics[width=\ww,frame]{figures/thin_masks_ablation/assets/bench_standard_pred.jpg} &
        \includegraphics[width=\ww,frame]{figures/thin_masks_ablation/assets/bench_pred1.jpg}
        \\

        \rotatebox{90}{\phantom{AAAA} ``green smoke''} &
        \includegraphics[width=\ww,frame]{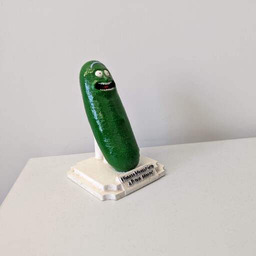} &
        \includegraphics[width=\ww,frame]{figures/thin_masks_ablation/assets/smoke_mask_overlay.jpg} &
        \includegraphics[width=\ww,frame]{figures/thin_masks_ablation/assets/smoke_standard_pred.jpg} &
        \includegraphics[width=\ww,frame]{figures/thin_masks_ablation/assets/smoke_pred1.jpg}
        \\

        \rotatebox{90}{\phantom{AAAAAAAA} ``fire''} &
        \includegraphics[width=\ww,frame]{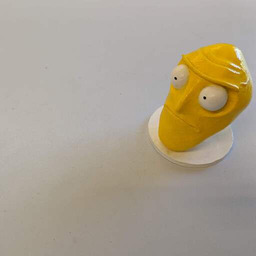} &
        \includegraphics[width=\ww,frame]{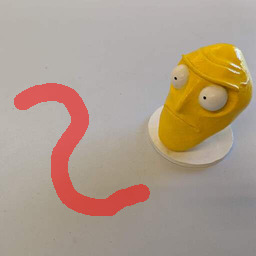} &
        \includegraphics[width=\ww,frame]{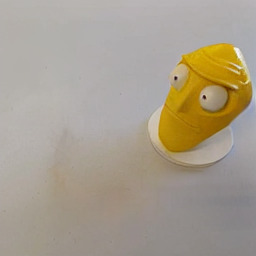} &
        \includegraphics[width=\ww,frame]{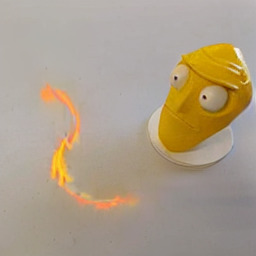}
        \\

        \rotatebox{90}{\phantom{AAAAA} ``green bracelet''} &
        \includegraphics[width=\ww,frame]{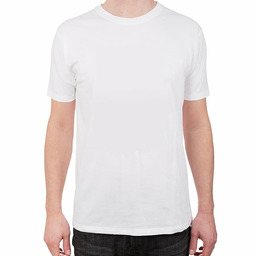} &
        \includegraphics[width=\ww,frame]{figures/thin_masks_ablation/assets/bracelet_mask_overlay.jpg} &
        \includegraphics[width=\ww,frame]{figures/thin_masks_ablation/assets/bracelet_standard_pred.jpg} &
        \includegraphics[width=\ww,frame]{figures/thin_masks_ablation/assets/bracelet_pred1.jpg}
        \\

        &
        Input image &
        Input mask &
        Standard process &
        Progressive mask shrinking
        \\
    \end{tabular}
    
    \caption{\textbf{Thin masks:} An expanded version 
    of Figure 6 from the main paper
    .}
    \label{fig:additional_ablation_thin masks}
\end{figure*}

%% file: figures/editing_session/birka/fig.tex
\begin{figure*}[h]
    \centering
    \setlength{\tabcolsep}{2pt}
    \renewcommand{\arraystretch}{1.5}
    \setlength{\ww}{0.6\columnwidth}
  
    \begin{tabular}{cccccc}
        \rotatebox{90}{Step 1: ``a man with a green sweater''}
        \includegraphics[width=\ww,frame]{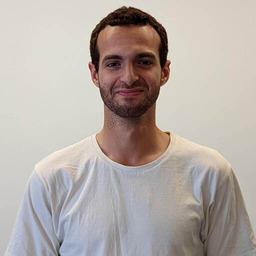} &
        \includegraphics[width=\ww,frame]{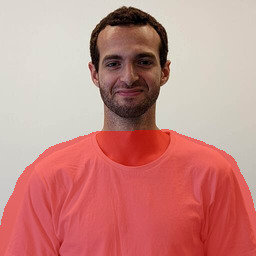} &
        \includegraphics[width=\ww,frame]{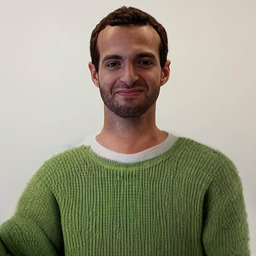} &
        \\

        \rotatebox{90}{\phantom{AAAAAA}Step 2: ``a straw hat''}
        \includegraphics[width=\ww,frame]{figures/editing_session/birka/assets/img2.jpg} &
        \includegraphics[width=\ww,frame]{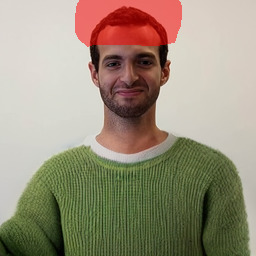} &
        \includegraphics[width=\ww,frame]{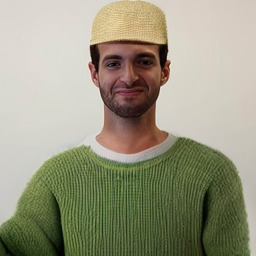} &
        \\

        \rotatebox{90}{\phantom{AAAAAA}Step 3: ``a window''}
        \includegraphics[width=\ww,frame]{figures/editing_session/birka/assets/img3.jpg} &
        \includegraphics[width=\ww,frame]{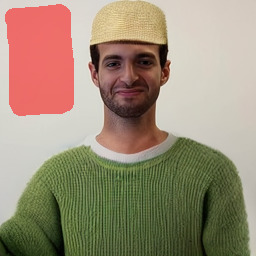} &
        \includegraphics[width=\ww,frame]{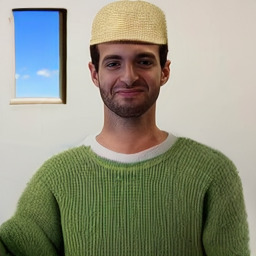} &
        \\

        Input image &
        Input mask &
        Prediction
        \\
    \end{tabular}
    
    \caption{\textbf{Editing session:} The user is able to perform several edit operations consecutively. First, the user provides the input image, mask, and text prompt ``a man with a green sweater'' to get the first result, then, he masks the head area and provides the text prompt ``a straw hat'', finally, he masks an area on the wall and provides the text ``a window'' to get the final result.}
    \label{fig:editing_session_birka}
\end{figure*}

%% file: appendices/additional_comparisons.tex
\section{Additional Comparisons}
In this section we start by comparing the number of parameters of our model against the baselines, discuss pixel-level artifacts of Blended Diffusion, show additional visual comparisons to the baselines, and compare against a variant of the background reconstruction loss.

\subsection{Parameters Comparison}
\label{sec:parameters_comparison}

In \Cref{tab:parameters_comparison} we compare the number of parameters in our model to that of the following baselines: (1) Local CLIP-guided diffusion \cite{clip_guided_diffusion} (for more details see Avrahami~et~al.~\shortcite{avrahami2022blended}), (2) $\textit{PaintByWord++}$ \cite{bau2021paint, crowson2022vqgan} (for more details see Avrahami~et~al.~\shortcite{avrahami2022blended}), (3) Blended Diffusion \cite{avrahami2022blended}, (4) GLIDE \cite{nichol2021glide} and (5) GLIDE-filtered \cite{nichol2021glide}.

\input{tables/parameters_comparison.tex}

\subsection{Pixel-level Artifacts Comparison}
\label{subsec:pixel-level-artifacts}
As described in Section 4 of the main paper, the latent space diffusion used by our method is not only faster than pixel-based diffusion, but also mitigates the pixel-level artifacts in Blended Diffusion \cite{avrahami2022blended}. We provide additional comparisons in \Cref{fig:noise_artifact_comparison_extended}.

\input{figures/noise_artifact_comparison_extended/fig.tex}

\subsection{Additional Comparison Against the Baselines}
In Figure 7 in the main paper, we compared our method against the baselines qualitatively on the set of images provided by Blended Diffusion \cite{avrahami2022blended}. In addition, we compare our method against the freely-available models in \Cref{fig:baselines_comparison_extended}. 

As we can see, baselines (1) Local CLIP-guided diffusion and (2) $\textit{PaintByWord++}$ fail to preserve the background of the input image. Baseline (4) GLIDE-filtered does not follow the guiding text, whereas (5) \DALLE~2 only partially corresponds to the guiding text (in the corgi and the yellow sweater examples). While (3) Blended Diffusion does preserve the background and follows all of the input guiding texts (except for the graffiti example), it suffers from noise-level artifacts as described in 
\Cref{subsec:pixel-level-artifacts}.

\input{figures/comparisons/baselines_extended/fig.tex}

\subsection{Background Reconstruction Loss Comparison}
As described in 
\Cref{sec:handling_innacurate_reconstruction}
we handled the background reconstruction by optimizing the decoder's weights $\theta$ on a per-image basis:
\begin{equation}
    \label{eqn:weights_optimization_appdx}
    \theta^* = \operatornamewithlimits{argmin}\limits_{\theta} \| D_{\theta}(z_0) \odot m - \hat{x} \odot m \| \;+ \\ \lambda \| D_{\theta}(z_0) \odot (1 - m) - x \odot (1 - m) \|
\end{equation}
where $D_{\theta}$ is the decoder, $m$ is the input mask, $x$ is the input image and $\hat{x}$ is the predicted image. Because our goal is to preserve the background, we set most of the weight to the background term (by setting $\lambda = 100$). It raises the question of what is the effect of dropping the foreground term completely. As demonstrated in \Cref{fig:reconstruction_loss_lambda_ablation}, doing so makes the colors of the edited area less vivid.

\input{figures/reconstruction_loss_lambada_ablation/fig.tex}

%% file: tables/parameters_comparison.tex
\begin{table}
    \caption{\textbf{Parameters comparison:} A comparison between the number of parameters of the different models. We used the same CLIP model for all the base models which has 0.15B parameters.}
    \begin{tabular}{lccc}
        \toprule
        
        \textbf{Method} & 
        \# Parameters  %
        &
        \\
        \midrule

        Local CLIP-guided diffusion &
        0.55B + 0.15B = 0.70B
        \\

        PaintByWord++ & 
        0.09B + 0.15B = 0.24B
        \\

        Blended Diffusion & 
        0.55B + 0.15B = 0.70B
        \\
        
        GLIDE & 
        5.00B + 0.15B = 5.15B
        \\

        GLIDE-filtered & 
        0.30B + 0.15B = 0.45B
        \\

        Ours & 
        1.45B + 0.15B = 1.60B
        \\
        \bottomrule
    \end{tabular}
    \label{tab:parameters_comparison}
\end{table}

%% file: figures/noise_artifact_comparison_extended/fig.tex
\begin{figure*}[ht]
    \centering
    \setlength{\tabcolsep}{-0.5pt}
    \renewcommand{\arraystretch}{-0.5}
    \setlength{\ww}{0.38\columnwidth}
  
    \begin{tabular}{cccccc}
        \rotatebox{90}{\phantom{A}} &
        \rotatebox{90}{\phantom{AAAA}``blue ball''} &

        \begin{tikzpicture}[spy using outlines={}]
            \node {\includegraphics[width=\ww,frame]{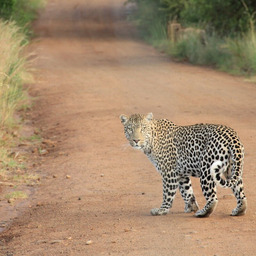}};
        \end{tikzpicture} &

        \begin{tikzpicture}[spy using outlines={}]
            \node {\includegraphics[width=\ww,frame]{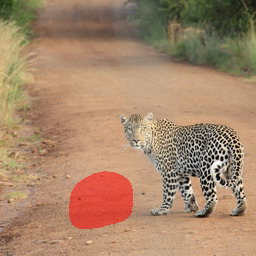}};
        \end{tikzpicture} &

        \begin{tikzpicture}[spy using outlines={circle,red,magnification=2.5,size=1cm, connect spies}]
            \node {\includegraphics[width=\ww,frame]{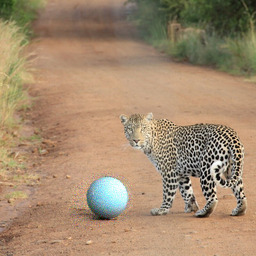}};
            \spy on (-0.5,-1.1) in node [left] at (-0.6,1.1);
        \end{tikzpicture} &

        \begin{tikzpicture}[spy using outlines={circle,red,magnification=2.5,size=1cm, connect spies}]
            \node {\includegraphics[width=\ww,frame]{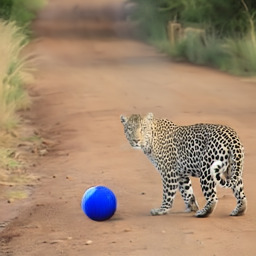}};
            \spy on (-0.5,-1.1) in node [left] at (-0.6,1.1);
        \end{tikzpicture}
        \\

        \rotatebox{90}{\phantom{A}} &
        \rotatebox{90}{\phantom{AAAA}``knit beanie''} &

        \begin{tikzpicture}[spy using outlines={}]
            \node {\includegraphics[width=\ww,frame]{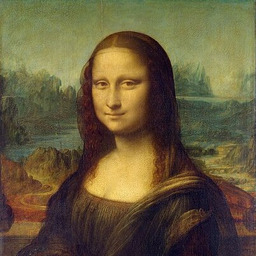}};
        \end{tikzpicture} &

        \begin{tikzpicture}[spy using outlines={}]
            \node {\includegraphics[width=\ww,frame]{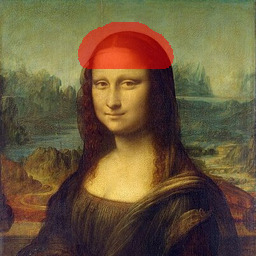}};
        \end{tikzpicture} &

        \begin{tikzpicture}[spy using outlines={circle,red,magnification=2.5,size=1cm, connect spies}]
            \node {\includegraphics[width=\ww,frame]{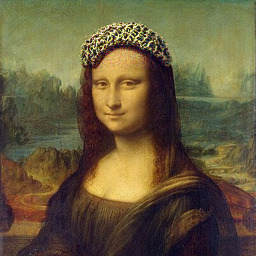}};
            \spy on (-0.1,1) in node [right] at (0.6,-1.1);
        \end{tikzpicture} &

        \begin{tikzpicture}[spy using outlines={circle,red,magnification=2.5,size=1cm, connect spies}]
            \node {\includegraphics[width=\ww,frame]{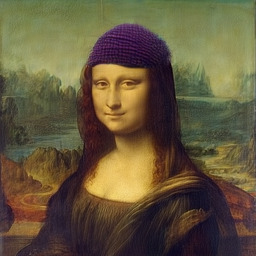}};
            \spy on (-0.1,1) in node [right] at (0.6,-1.1);
        \end{tikzpicture}
        \\

        \rotatebox{90}{\phantom{A}} &
        \rotatebox{90}{\phantom{AAAAAA}``pizza''} &

        \begin{tikzpicture}[spy using outlines={}]
            \node {\includegraphics[width=\ww,frame]{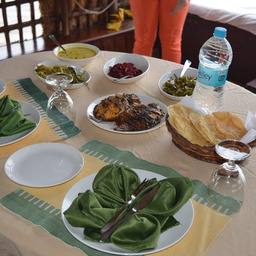}};
        \end{tikzpicture} &

        \begin{tikzpicture}[spy using outlines={}]
            \node {\includegraphics[width=\ww,frame]{figures/comparisons/baselines/assets/mask_overlay4.jpg}};
        \end{tikzpicture} &

        \begin{tikzpicture}[spy using outlines={circle,red,magnification=2.5,size=1cm, connect spies}]
            \node {\includegraphics[width=\ww,frame]{figures/comparisons/baselines/assets/blended_diffusion4.jpg}};
            \spy on (-0.8,-1.1) in node [left] at (-0.6,1.1);
        \end{tikzpicture} &

        \begin{tikzpicture}[spy using outlines={circle,red,magnification=2.5,size=1cm, connect spies}]
            \node {\includegraphics[width=\ww,frame]{figures/comparisons/baselines/assets/ours/dinner/pred_recon1.jpg}};
            \spy on (-0.8,-1.1) in node [left] at (-0.6,1.1);
        \end{tikzpicture}
        \\

        \rotatebox{90}{\phantom{A}} &
        \rotatebox{90}{\phantom{AAA}``corgi painting''} &

        \begin{tikzpicture}[spy using outlines={}]
            \node {\includegraphics[width=\ww,frame]{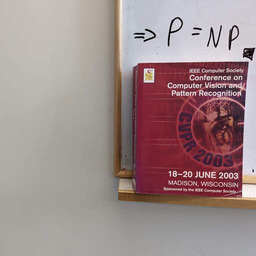}};
        \end{tikzpicture} &

        \begin{tikzpicture}[spy using outlines={}]
            \node {\includegraphics[width=\ww,frame]{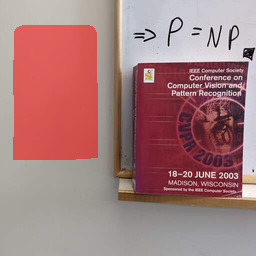}};
        \end{tikzpicture} &

        \begin{tikzpicture}[spy using outlines={circle,red,magnification=2.5,size=1cm, connect spies}]
            \node {\includegraphics[width=\ww,frame]{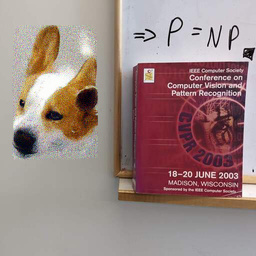}};
            \spy on (-1.3,1.2) in node [right] at (0.6,-1.1);
        \end{tikzpicture} &

        \begin{tikzpicture}[spy using outlines={circle,red,magnification=2.5,size=1cm, connect spies}]
            \node {\includegraphics[width=\ww,frame]{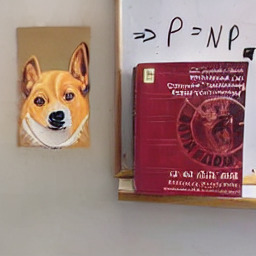}};
            \spy on (-1.3,1.2) in node [right] at (0.6,-1.1);
        \end{tikzpicture}
        \\

        \rotatebox{90}{\phantom{A}} &
        \rotatebox{90}{\phantom{AAAAA}``cucumber''} &

        \begin{tikzpicture}[spy using outlines={}]
            \node {\includegraphics[width=\ww,frame]{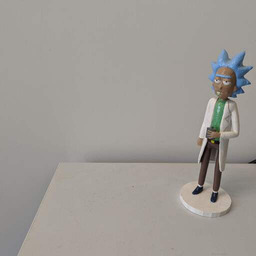}};
        \end{tikzpicture} &

        \begin{tikzpicture}[spy using outlines={}]
            \node {\includegraphics[width=\ww,frame]{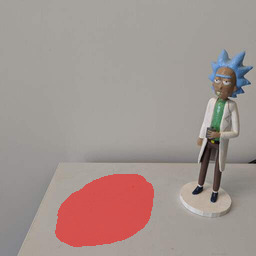}};
        \end{tikzpicture} &

        \begin{tikzpicture}[spy using outlines={circle,red,magnification=2.5,size=1cm, connect spies}]
            \node {\includegraphics[width=\ww,frame]{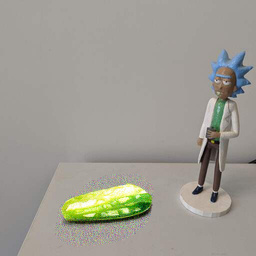}};
            \spy on (-0.6,-1.4) in node [left] at (-0.6,1.1);
        \end{tikzpicture} &

        \begin{tikzpicture}[spy using outlines={circle,red,magnification=2.5,size=1cm, connect spies}]
            \node {\includegraphics[width=\ww,frame]{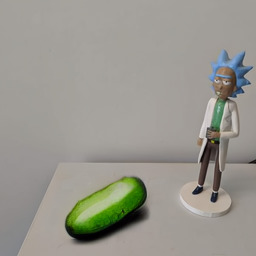}};
            \spy on (-0.6,-1.4) in node [left] at (-0.6,1.1);
        \end{tikzpicture}
        \\

        \rotatebox{90}{\phantom{AAAA}``a man with } &
        \rotatebox{90}{\phantom{AAA}a yellow sweater''} &

        \begin{tikzpicture}[spy using outlines={}]
            \node {\includegraphics[width=\ww,frame]{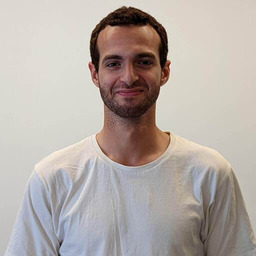}};
        \end{tikzpicture} &

        \begin{tikzpicture}[spy using outlines={}]
            \node {\includegraphics[width=\ww,frame]{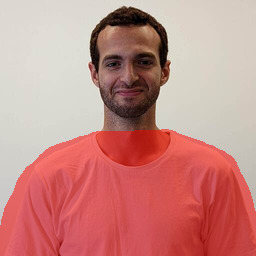}};
        \end{tikzpicture} &

        \begin{tikzpicture}[spy using outlines={circle,red,magnification=2.5,size=1cm, connect spies}]
            \node {\includegraphics[width=\ww,frame]{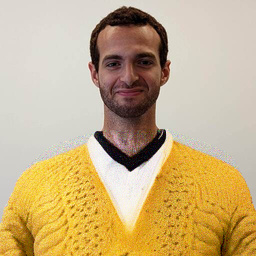}};
            \spy on (-0.8,-0.3) in node [left] at (-0.6,1.1);
        \end{tikzpicture} &

        \begin{tikzpicture}[spy using outlines={circle,red,magnification=2.5,size=1cm, connect spies}]
            \node {\includegraphics[width=\ww,frame]{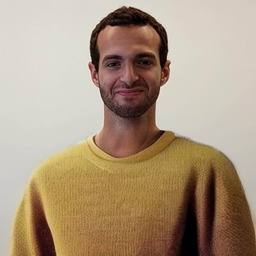}};
            \spy on (-0.8,-0.3) in node [left] at (-0.6,1.1);
        \end{tikzpicture}
        \\

        &&
        (a) Input image &
        (b) Input mask &
        (c) Blended Diffusion &
        (d) Ours
    \end{tabular}
    \vspace{-2.5mm}
    \caption{\textbf{Noise artifacts:} Given the input image (a) and mask (b) with some guiding text, Blended Diffusion produces noticeable pixel-level noise artifacts (c), in contrast to our method (d).}
    \label{fig:noise_artifact_comparison_extended}
\end{figure*}

%% file: figures/comparisons/baselines_extended/fig.tex
\begin{figure*}[h]
    \centering
    \setlength{\tabcolsep}{0.5pt}
    \renewcommand{\arraystretch}{0.5}
    \setlength{\ww}{0.33\columnwidth}
  
    \begin{tabular}{cccccccc}
        \rotatebox{90}{\phantom{A}} &
        \rotatebox{90}{\phantom{AA} Input + mask} &
        \includegraphics[width=\ww,frame]{figures/comparisons/baselines_extended/assets/corgi/mask_overlay.jpg} &
        \includegraphics[width=\ww,frame]{figures/comparisons/baselines_extended/assets/cucumber/mask_overlay.jpg} &
        \includegraphics[width=\ww,frame]{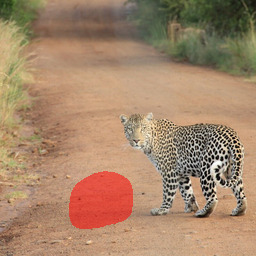} &
        \includegraphics[width=\ww,frame]{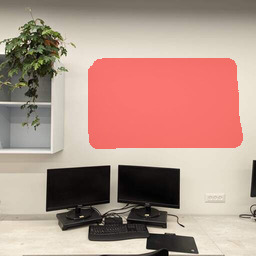} &
        \includegraphics[width=\ww,frame]{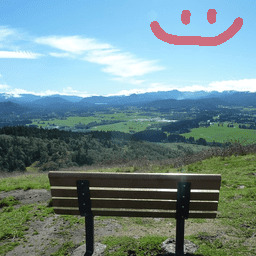} &
        \includegraphics[width=\ww,frame]{figures/comparisons/baselines_extended/assets/yellow_sweater/mask_overlay.jpg}
        \\
        
        \rotatebox{90}{\phantom{AAA}Local CLIP-} &
        \rotatebox{90}{\phantom{AA}guided diffusion} &
        \includegraphics[width=\ww,frame]{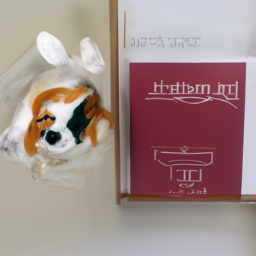} &
        \includegraphics[width=\ww,frame]{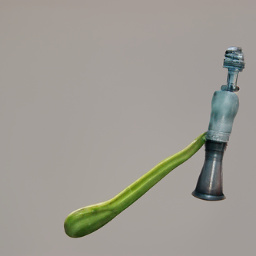} &
        \includegraphics[width=\ww,frame]{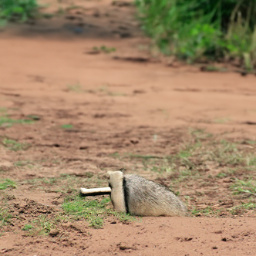} &
        \includegraphics[width=\ww,frame]{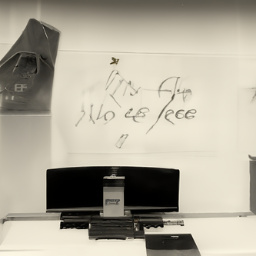} &
        \includegraphics[width=\ww,frame]{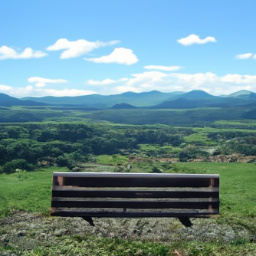} &
        \includegraphics[width=\ww,frame]{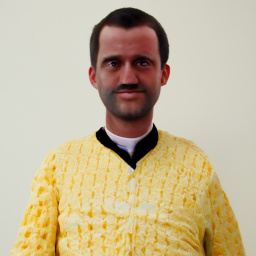}
        \\
        
        \rotatebox{90}{\phantom{A}} &
        \rotatebox{90}{\phantom{AA}{$\textit{PaintByWord++}$}} &
        \includegraphics[width=\ww,frame]{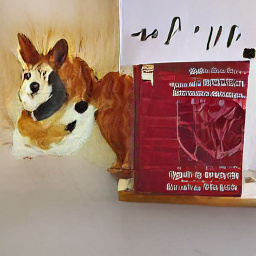} &
        \includegraphics[width=\ww,frame]{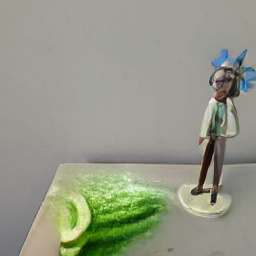} &
        \includegraphics[width=\ww,frame]{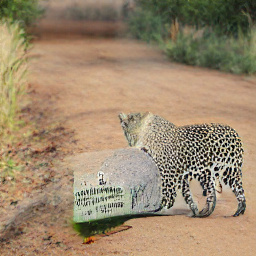} &
        \includegraphics[width=\ww,frame]{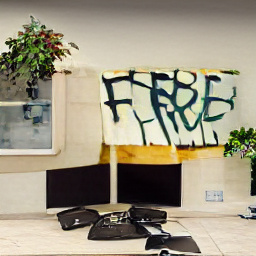} &
        \includegraphics[width=\ww,frame]{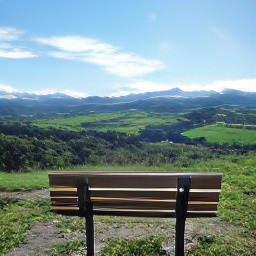} &
        \includegraphics[width=\ww,frame]{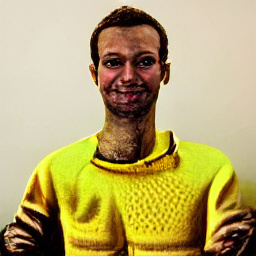}
        \\
        
        \rotatebox{90}{\phantom{A}} &
        \rotatebox{90}{\phantom{A}{Blended Diffusion}} &
        \includegraphics[width=\ww,frame]{figures/comparisons/baselines_extended/assets/corgi/blended_diffusion.jpg} &
        \includegraphics[width=\ww,frame]{figures/comparisons/baselines_extended/assets/cucumber/blended_diffusion.jpg} &
        \includegraphics[width=\ww,frame]{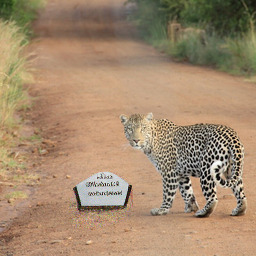} &
        \includegraphics[width=\ww,frame]{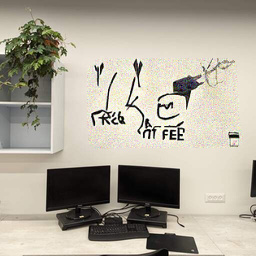} &
        \includegraphics[width=\ww,frame]{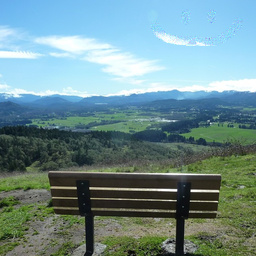} &
        \includegraphics[width=\ww,frame]{figures/comparisons/baselines_extended/assets/yellow_sweater/blended_diffusion.jpg}
        \\

        \rotatebox{90}{\phantom{A}} &
        \rotatebox{90}{\phantom{AAa}{GLIDE-filtered}} &
        \includegraphics[width=\ww,frame]{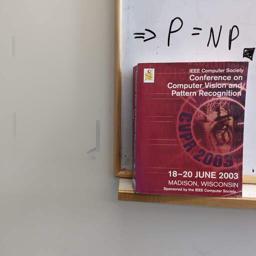} &
        \includegraphics[width=\ww,frame]{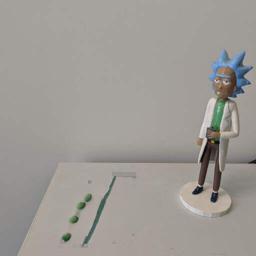} &
        \includegraphics[width=\ww,frame]{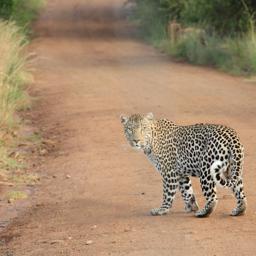} &
        \includegraphics[width=\ww,frame]{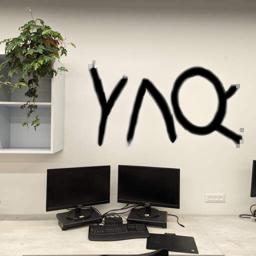} &
        \includegraphics[width=\ww,frame]{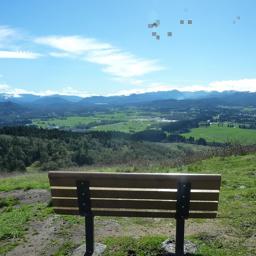} &
        \includegraphics[width=\ww,frame]{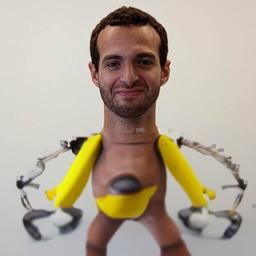}
        \\

        \rotatebox{90}{\phantom{A}} &
        \rotatebox{90}{\phantom{AAAa}{\DALLE~2}} &
        \includegraphics[width=\ww,frame]{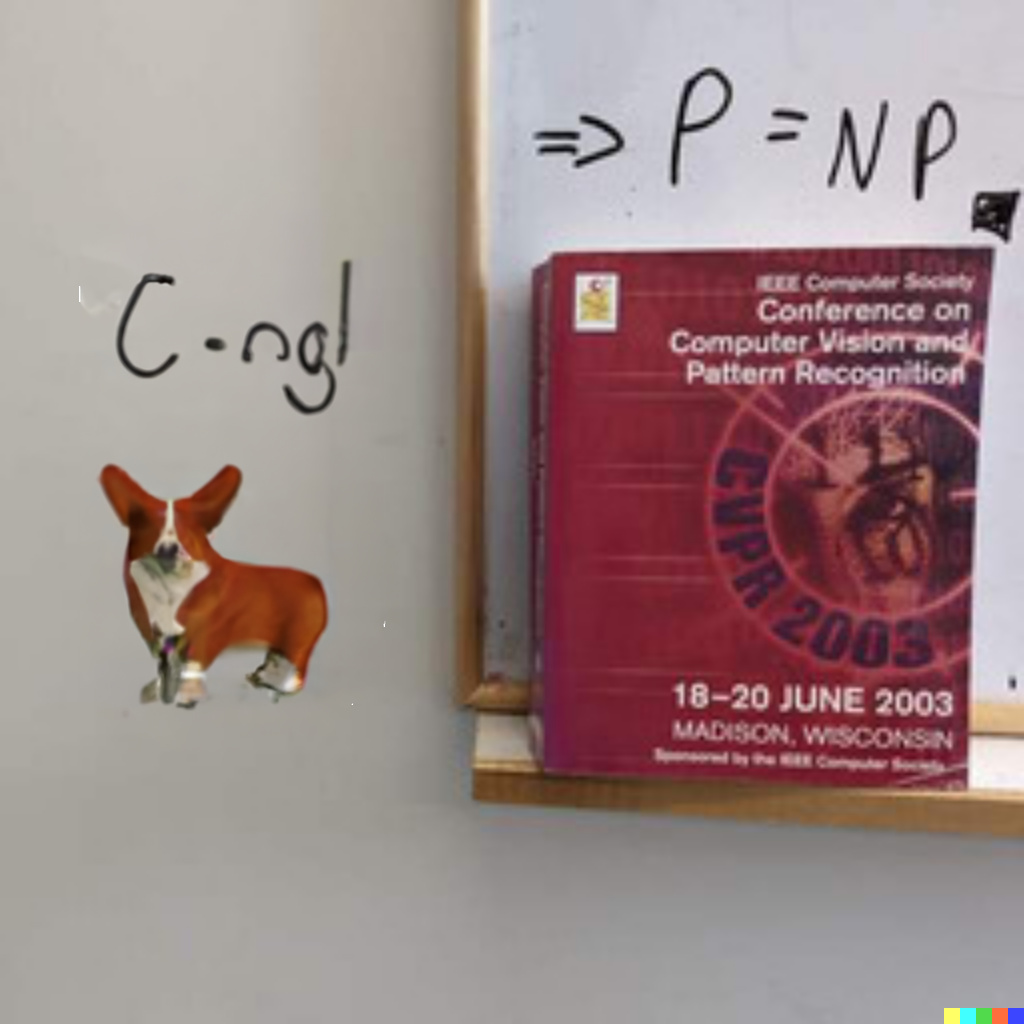} &
        \includegraphics[width=\ww,frame]{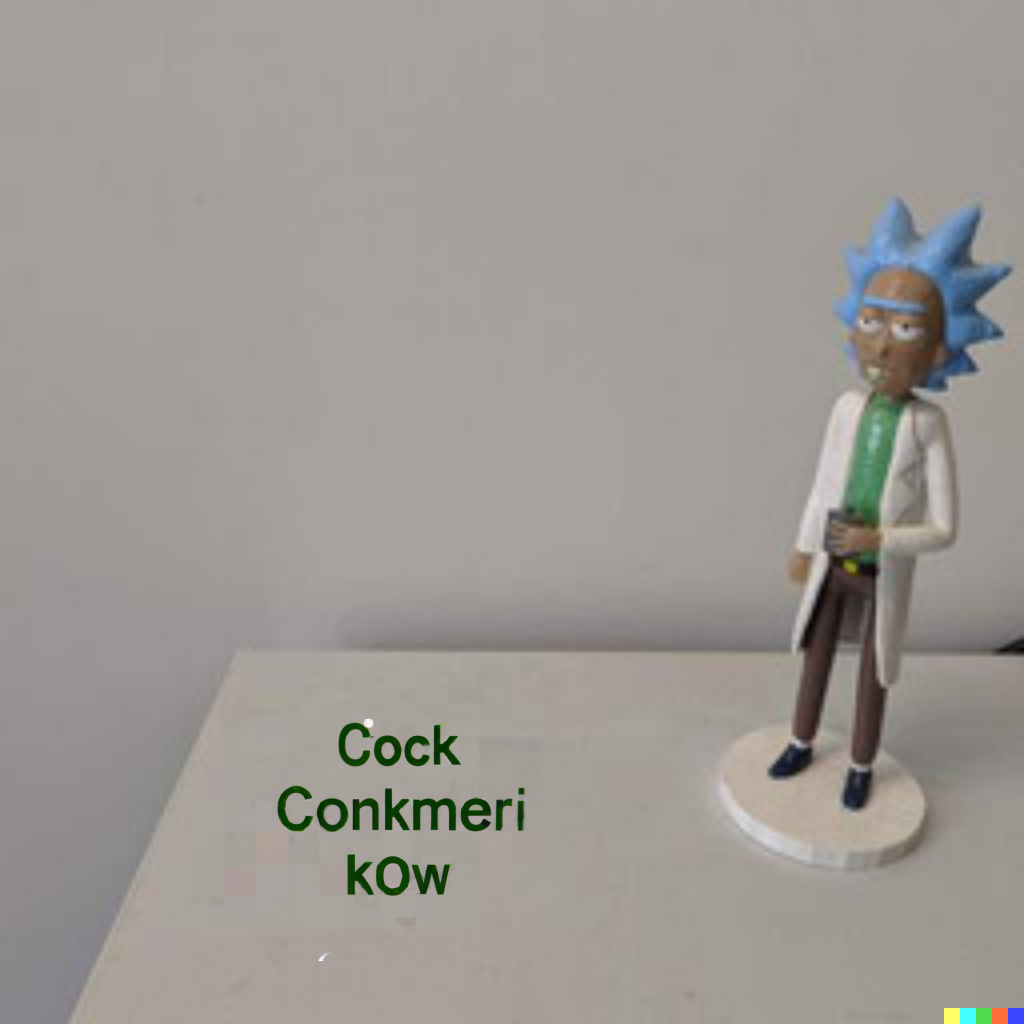} &
        \includegraphics[width=\ww,frame]{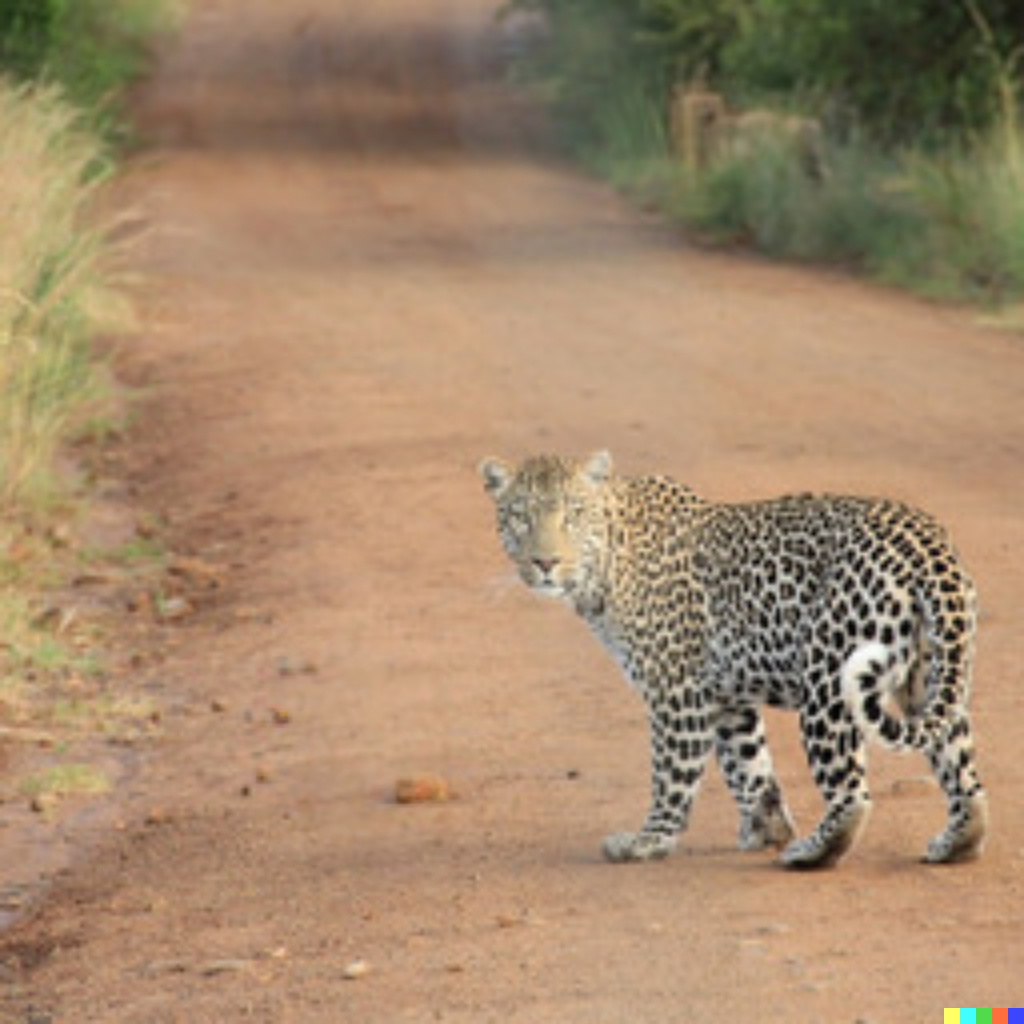} &
        \includegraphics[width=\ww,frame]{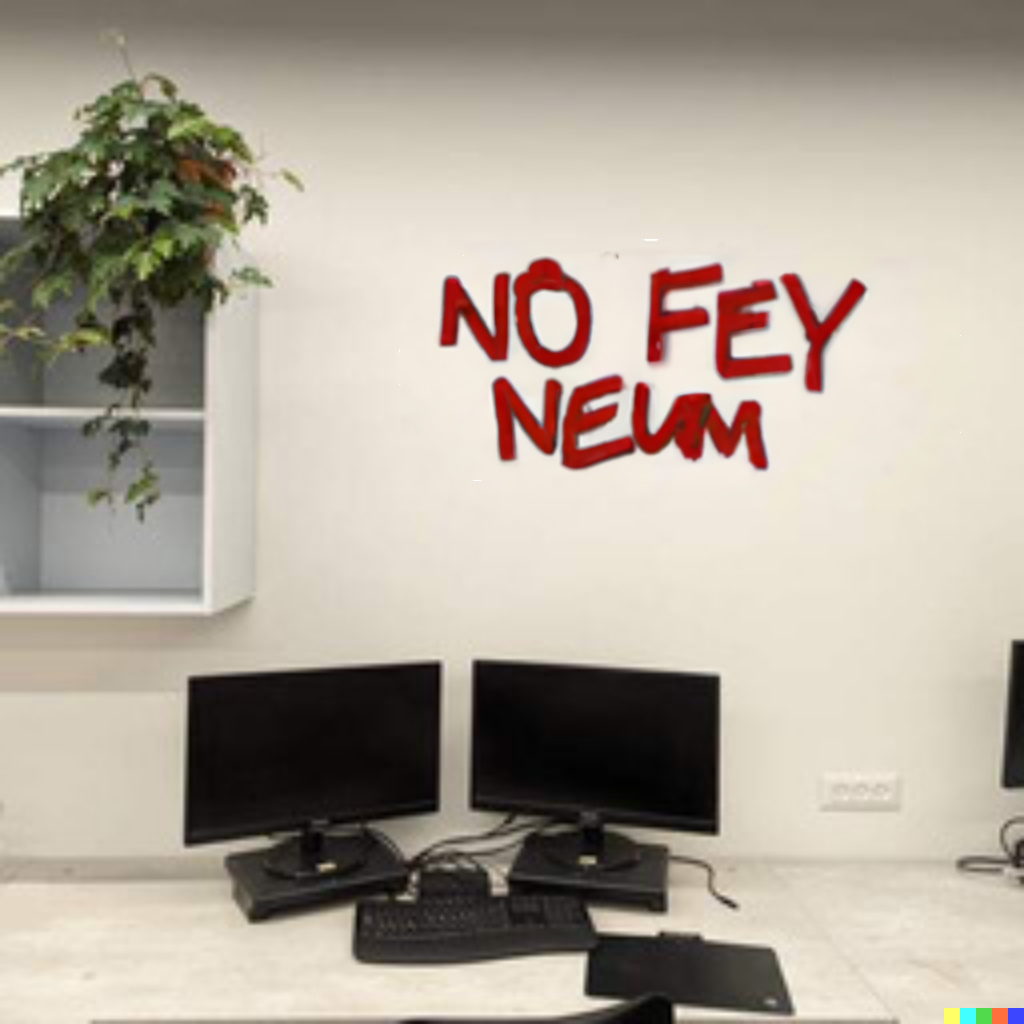} &
        \includegraphics[width=\ww,frame]{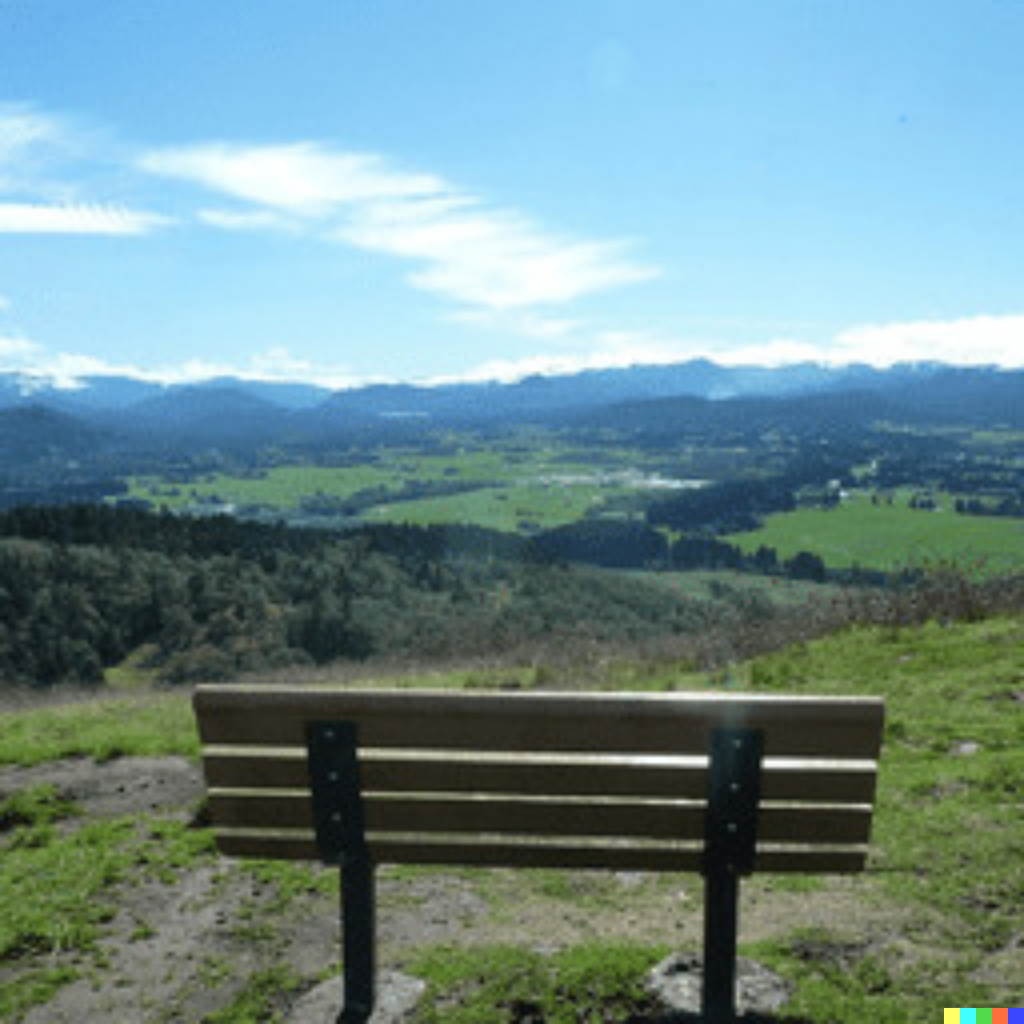} &
        \includegraphics[width=\ww,frame]{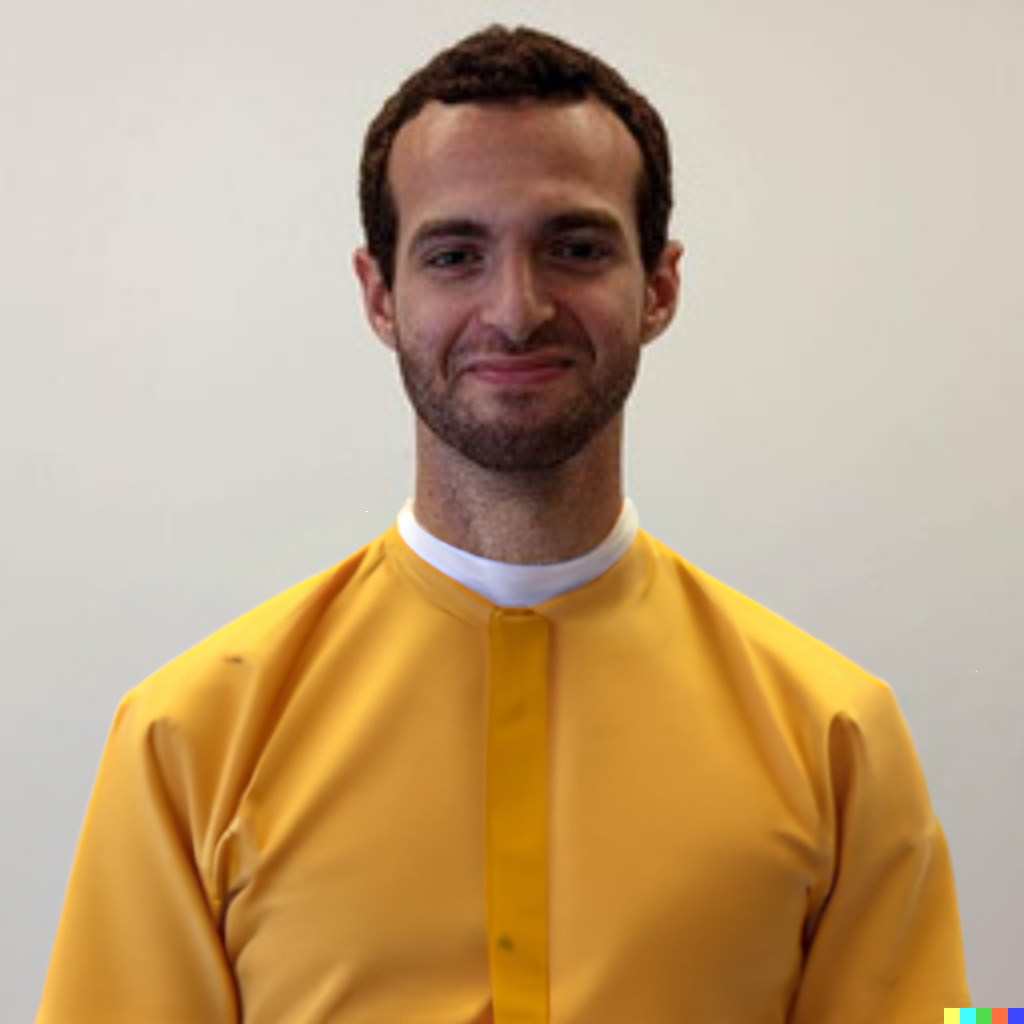}
        \\

        \rotatebox{90}{\phantom{A}} &
        \rotatebox{90}{\phantom{AAAAA}{Ours}} &
        \includegraphics[width=\ww,frame]{figures/comparisons/baselines_extended/assets/corgi/ours.jpg} &
        \includegraphics[width=\ww,frame]{figures/comparisons/baselines_extended/assets/cucumber/ours.jpg} &
        \includegraphics[width=\ww,frame]{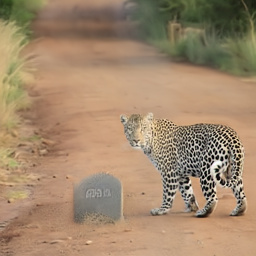} &
        \includegraphics[width=\ww,frame]{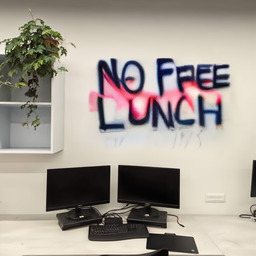} &
        \includegraphics[width=\ww,frame]{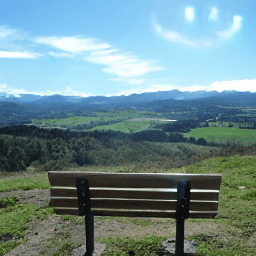} &
        \includegraphics[width=\ww,frame]{figures/comparisons/baselines_extended/assets/yellow_sweater/ours.jpg}
        \\
        
        &&
        ``corgi painting'' &
        ``cucumber'' &
        ``gravestone'' &
        ``graffiti with the text &
        ``white clouds'' &
        ``a man with a
        \\
        
        &&&&&
        `No Free Lunch' ''
        &&
        yellow sweater''
        \\
    \end{tabular}
    \caption{\textbf{Comparison to baselines:} A comparison with Local CLIP-guided diffusion \cite{clip_guided_diffusion}, $\textit{PaintByWord++}$ \cite{bau2021paint, crowson2022vqgan}, Blended Diffusion \cite{avrahami2022blended}, GLIDE-filtered \cite{nichol2021glide} and \DALLE~2 \cite{ramesh2022hierarchical}.}
    \label{fig:baselines_comparison_extended}
\end{figure*}

%% file: figures/reconstruction_loss_lambada_ablation/fig.tex
\begin{figure*}[t]
    \centering
    \setlength{\tabcolsep}{0.5pt}
    \renewcommand{\arraystretch}{0.5}
    \setlength{\ww}{0.4\columnwidth}
  
    \begin{tabular}{ccccc}
	
    \rotatebox{90}{\phantom{AAAAA}{``red hair''}}  &
    \includegraphics[width=\ww,frame]{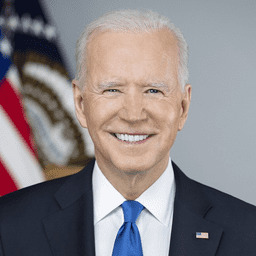} &
    \includegraphics[width=\ww,frame]{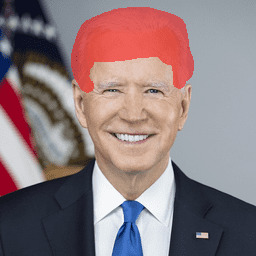} &
    \includegraphics[width=\ww,frame]{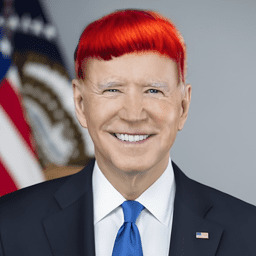} &
    \includegraphics[width=\ww,frame]{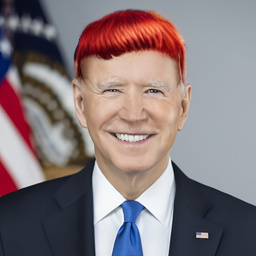}
    \\

    \rotatebox{90}{\phantom{AAA}{``colorful balls''}}  &
    \includegraphics[width=\ww,frame]{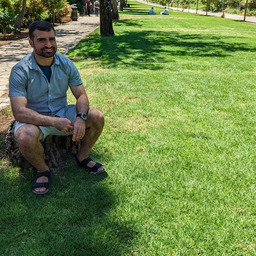} &
    \includegraphics[width=\ww,frame]{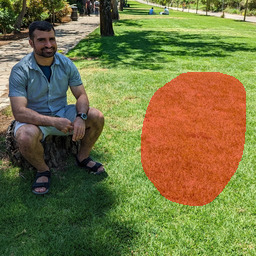} &
    \includegraphics[width=\ww,frame]{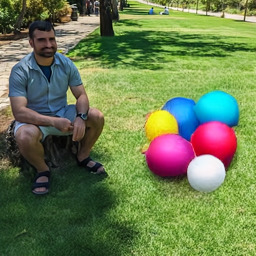} &
    \includegraphics[width=\ww,frame]{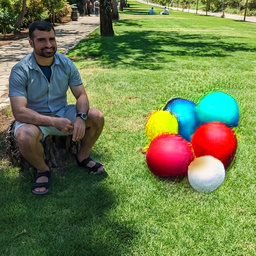}
    \\

    \rotatebox{90}{\phantom{AAA}{``purple stones''}}  &
    \includegraphics[width=\ww,frame]{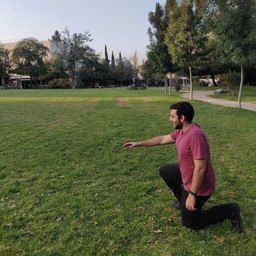} &
    \includegraphics[width=\ww,frame]{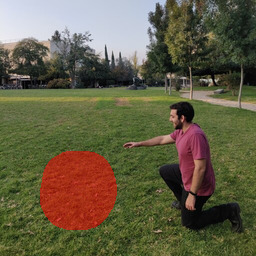} &
    \includegraphics[width=\ww,frame]{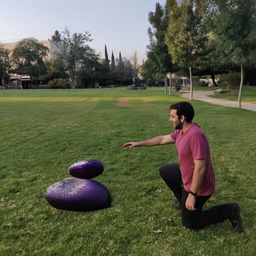} &
    \includegraphics[width=\ww,frame]{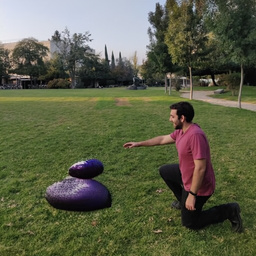}
    \\

    &
    &
	input + mask & 
    with foreground term & 
    without foreground term
    \\
\end{tabular}
\caption{\textbf{Reconstruction loss ablation:} Removing the foreground term in \Cref{eqn:weights_optimization_appdx} results in slightly less vivid colors.}
\label{fig:reconstruction_loss_lambda_ablation}
\end{figure*}

%% file: appendices/implementation_details.tex
\section{Implementation Details}
\label{sec:implemenation_details}

For all the experiments reported in this paper, the pretrained models that we have used are:
\begin{itemize}
    \item Text-to-image Latent Diffusion model published by Rombach~et~al.~\shortcite{rombach2021high}.
    \item CLIP model with ViT-B/16 backbone for the Vision Transformer \cite{dosovitskiy2020image}, as released by Radford et~al.~\shortcite{radford2021learning}.
    \item Blended Diffusion model from Avrahami~et~al.~\shortcite{avrahami2022blended}.
    \item GLIDE-filtered model from Nichol~et~al.~\shortcite{nichol2021glide}.
\end{itemize}

All these methods were released under MIT license and were implemented using PyTorch \cite{paszke2019pytorch}. 

In addition, we used the online demo of \DALLE~2 \cite{dalle2_demo} which enables the user to manually edit a real image using its interface. Nevertheless, the usage of the system is free for only a limited number of credit tokens, and the model is not available. Hence, we could not calculate our precision and diversity metrics on this model.

All the input images in this paper are real images that were released freely under a Creative Commons license or from our private collection.

In the reconstruction methods described in
\Cref{sec:handling_innacurate_reconstruction}
we used the following:
\begin{itemize}
    \item For Poisson image blending \cite{perez2003poisson} we used the OpenCV \cite{bradski2000opencv} implementation.
    \item For latent optimization and weights optimization we used Adam optimizer \cite{kingma2014adam} with a learning rate of 0.0001 for 75 optimization steps per image.
\end{itemize}

For the progressive mask shrinking described in
\Cref{sec:handling_thin_masks}
we used the following scheme: we dilate the downsampled mask $m_{\textit{latent}}$ with kernels of ones with sizes $3 \times 3$, $5 \times 5$ and $7 \times 7$, then we divide the diffusion process into four parts with the same number of steps in each part, with the first part using the most dilated mask, and the last part using the original mask.

\subsection{Precision \& Diversity Metrics}
\label{sec:precision_and_diversity_supp}
As described in
\Cref{sec:results}
we calculated precision and diversity metrics in order to compare our method against the baselines quantitatively. As was shown by Nichol~et~al.~\shortcite{nichol2021glide}, using CLIP model as an evaluator for text correspondence of images that were edited with models that use CLIP's gradients for generation, is not correlated with human evaluation, because these models are susceptible to adversarial examples. Hence, because some of our baselines are using CLIP, we had to look for an alternative evaluation model. We opted to use a pre-trained ImageNet classifier, EfficentNet \cite{tan2019efficientnet}, as our backbone.  

We took 50 random images from the web and local collection; next, for each image, we generated a random rectangular mask with dimensions that are in the range $[\frac{dim}{5}, \frac{dim}{2}]$ where $dim$ is the corresponding image dimension. Then, for each of the resulting image-mask pairs, we sample a random class from ImageNet classes and use the corresponding text label of that class as an input to our model. For each of the baseline models, we generate predictions of the recommended batch size. An example of an input and its predictions by the various baselines can be seen in \Cref{fig:experiment_results_example}.

\input{figures/experiment_results/fig.tex}

To calculate the precision for each model, we go over all its predictions, mask them using the input mask, and feed the masked results to the ImageNet classifier. Because ImageNet contains many classes with semantically close meaning (e.g., several different species of dogs), we considered prediction as a good prediction if the ground-truth class label (the label of the class that was fed to the generative model) is in the top-5 predictions of the classification model. We calculate the average accuracy at the batch level for each input. In addition, we calculate the precision only on the top result that was ranked by the CLIP model as described in
\Cref{sec:predictions_ranking}
Both of these metrics are reported in \Cref{tab:metrics_comparison}

In order to calculate the diversity at the batch level, for each input triplet, we take only the images that were classified correctly by the classifier (because only these images are of interest to the end-user). We then mask the images using the corresponding masks, in order to isolate the diversity of the foreground and then calculate the pairwise LPIPS \cite{zhang2018unreasonable} distance and take the average across all the predictions.

\subsection{User Study}
As described in
\Cref{sec:results}
we conducted a user study in order to assess the visual quality of the results and how well they match the guiding text, using the Amazon Mechanical Turk platform (AMT) \cite{amt_website}. 
We used the 50 random predictions that were used to evaluate our method quantitatively, as described in \Cref{sec:precision_and_diversity_supp}. We presented each human evaluator with two images --- one produced by our method and the other one by a baseline, and asked them to rate which of the two images has (1) better visual quality by asking ``Which of the following images has better visual quality?'' and (2) better matches the text prompt by asking ``Which of the following images matches the label X more closely?'' (replacing X with the text prompt). We used the majority vote of the raters for each question. The human raters could also indicate for each question that neither of the images is better than the other (``Equal quality'' for the image quality/``Equally match'' for the text matching), in which case we split the points between both of the models equally. We collected five ratings per question, resulting in 250 ratings per task (visual quality/text match). The time allotted per image-pair task was one hour, to allow the raters to properly evaluate the results without time pressure. 

We included in our user study only the freely available models that could be used with the random predictions, hence, the study does not include the GLIDE-full \cite{nichol2021glide} and \DALLE~2 \cite{ramesh2022hierarchical} models, which are unavailable. A binomial statistical significance test, reported in \Cref{tab:statistical_analysis}, suggests that these results are statistically significant.

\input{tables/statistical_analysis.tex}

\subsection{Ranking Effectiveness}

\input{figures/ranking_effectiveness/fig.tex}
\input{figures/baselines_ranking/blended_diffusion/fig.tex}

As described in
Section 4.4 in the main paper,
we utilized the CLIP model in order to rank the predictions of our method. As demonstrated in \Cref{fig:ranking_effectiveness}, during our experiments we noticed that the top 20\% are constantly better than the bottom 20\%, but not at the granularity of a single image --- the first image is not always strictly better than the second.

In addition, \Cref{fig:raking_effectiveness_blended_diffusion} demonstrates the importance of the CLIP ranking for the Blended Diffusion baseline \cite{avrahami2022blended}. As we can see, the CLIP ranking is essential to this method. Hence, the ``full batch'' column in Table 2 on the main paper is the relevant information we should take into account when comparing the inference times of our method with those of the Blended Diffusion baseline.

%% file: figures/experiment_results/fig.tex
\begin{figure*}[h]
    \centering
    \setlength{\tabcolsep}{0.5pt}
    \renewcommand{\arraystretch}{0.5}
    \setlength{\ww}{0.29\columnwidth}
  
    \begin{tabular}{cccccccc}
        \rotatebox{90}{\phantom{AAA}``crane''} &
        \includegraphics[width=\ww,frame]{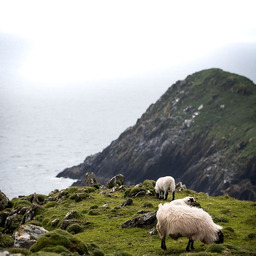} &
        \includegraphics[width=\ww,frame]{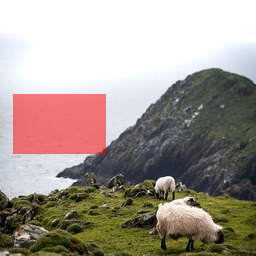} &
        \includegraphics[width=\ww,frame]{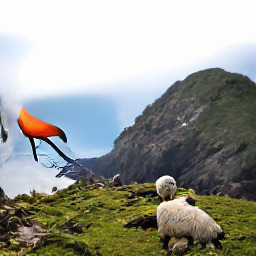} &
        \includegraphics[width=\ww,frame]{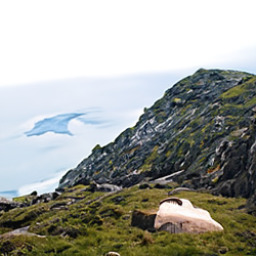} &
        \includegraphics[width=\ww,frame]{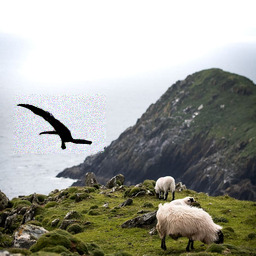} &
        \includegraphics[width=\ww,frame]{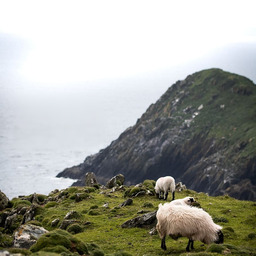} &
        \includegraphics[width=\ww,frame]{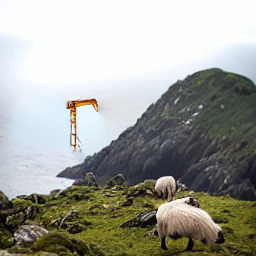}
        \\

        \rotatebox{90}{\phantom{AAAa}``iPod''} &
        \includegraphics[width=\ww,frame]{figures/experiment_results/assets/ipod/img.jpg} &
        \includegraphics[width=\ww,frame]{figures/experiment_results/assets/ipod/mask_overlay.jpg} &
        \includegraphics[width=\ww,frame]{figures/experiment_results/assets/ipod/pred_paint_by_word_pp.jpg} &
        \includegraphics[width=\ww,frame]{figures/experiment_results/assets/ipod/pred_local_clip_guided_diffusion.jpg} &
        \includegraphics[width=\ww,frame]{figures/experiment_results/assets/ipod/pred_blended_diffusion.jpg} &
        \includegraphics[width=\ww,frame]{figures/experiment_results/assets/ipod/pred_glide_filtered.jpg} &
        \includegraphics[width=\ww,frame]{figures/experiment_results/assets/ipod/pred_blended_latent_diffusion.jpg}
        \\

        \rotatebox{90}{\phantom{AAA}``hourglass''} &
        \includegraphics[width=\ww,frame]{figures/experiment_results/assets/hourglass/img.jpg} &
        \includegraphics[width=\ww,frame]{figures/experiment_results/assets/hourglass/mask_overlay.jpg} &
        \includegraphics[width=\ww,frame]{figures/experiment_results/assets/hourglass/pred_paint_by_word_pp.jpg} &
        \includegraphics[width=\ww,frame]{figures/experiment_results/assets/hourglass/pred_local_clip_guided_diffusion.jpg} &
        \includegraphics[width=\ww,frame]{figures/experiment_results/assets/hourglass/pred_blended_diffusion.jpg} &
        \includegraphics[width=\ww,frame]{figures/experiment_results/assets/hourglass/pred_glide_filtered.jpg} &
        \includegraphics[width=\ww,frame]{figures/experiment_results/assets/hourglass/pred_blended_latent_diffusion.jpg}
        \\

        \rotatebox{90}{\phantom{AAA}``flamingo''} &
        \includegraphics[width=\ww,frame]{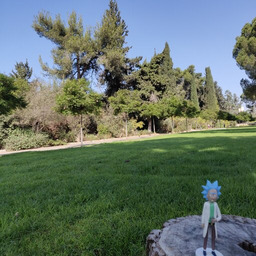} &
        \includegraphics[width=\ww,frame]{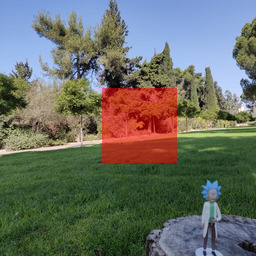} &
        \includegraphics[width=\ww,frame]{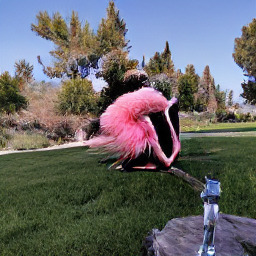} &
        \includegraphics[width=\ww,frame]{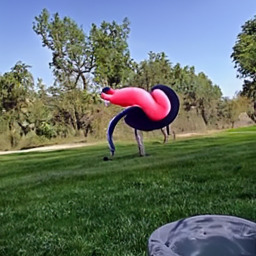} &
        \includegraphics[width=\ww,frame]{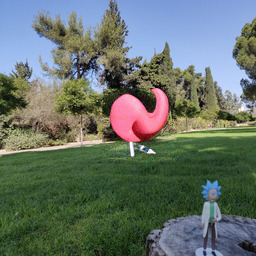} &
        \includegraphics[width=\ww,frame]{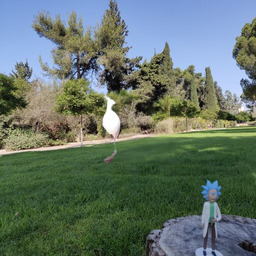} &
        \includegraphics[width=\ww,frame]{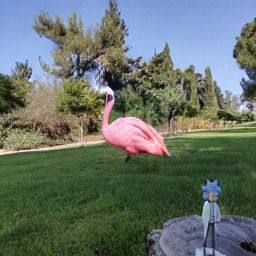}
        \\

        \rotatebox{90}{\phantom{AAa}``Chihuahua''} &
        \includegraphics[width=\ww,frame]{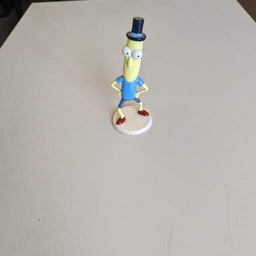} &
        \includegraphics[width=\ww,frame]{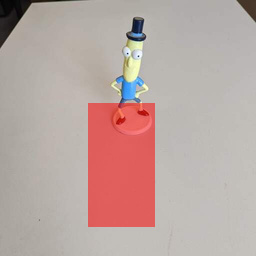} &
        \includegraphics[width=\ww,frame]{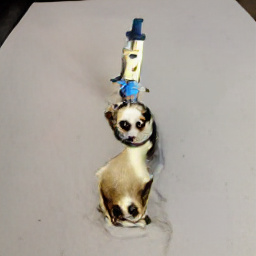} &
        \includegraphics[width=\ww,frame]{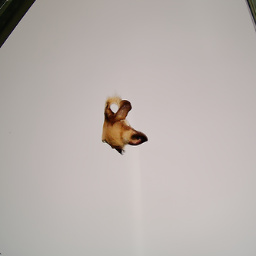} &
        \includegraphics[width=\ww,frame]{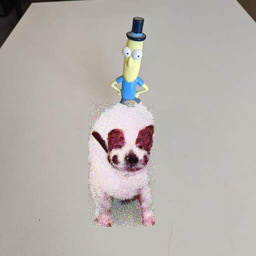} &
        \includegraphics[width=\ww,frame]{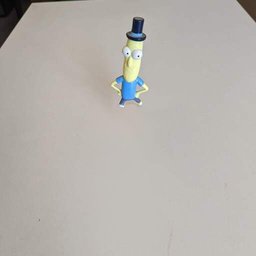} &
        \includegraphics[width=\ww,frame]{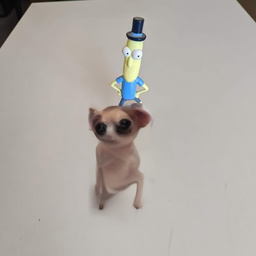}
        \\

        \rotatebox{90}{\phantom{AAa}``cockroach''} &
        \includegraphics[width=\ww,frame]{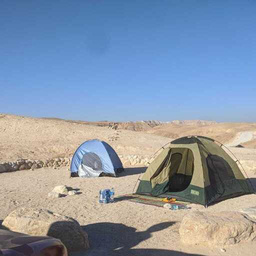} &
        \includegraphics[width=\ww,frame]{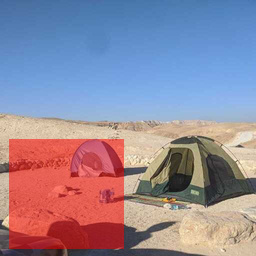} &
        \includegraphics[width=\ww,frame]{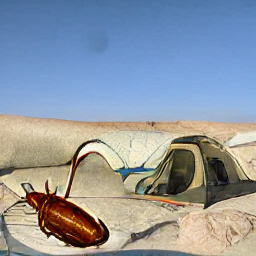} &
        \includegraphics[width=\ww,frame]{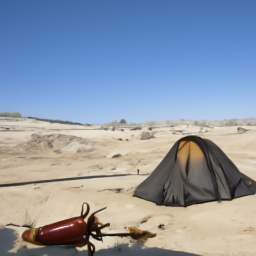} &
        \includegraphics[width=\ww,frame]{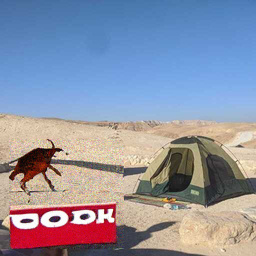} &
        \includegraphics[width=\ww,frame]{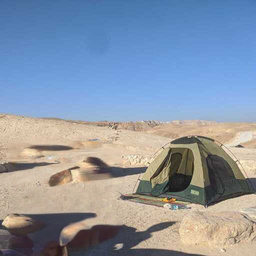} &
        \includegraphics[width=\ww,frame]{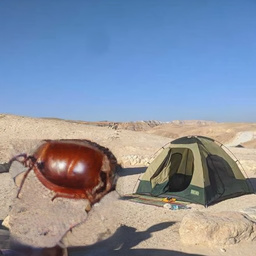}
        \\

        \rotatebox{90}{\phantom{AAA}``baloon''} &
        \includegraphics[width=\ww,frame]{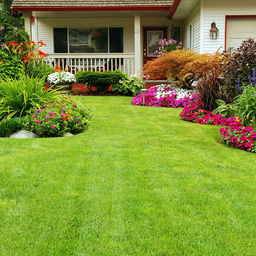} &
        \includegraphics[width=\ww,frame]{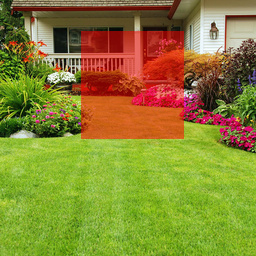} &
        \includegraphics[width=\ww,frame]{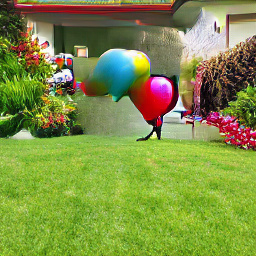} &
        \includegraphics[width=\ww,frame]{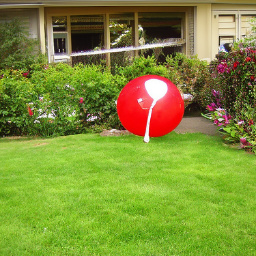} &
        \includegraphics[width=\ww,frame]{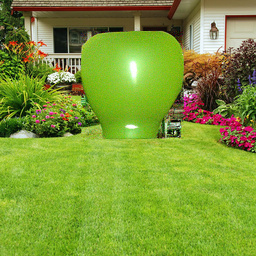} &
        \includegraphics[width=\ww,frame]{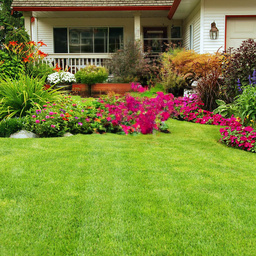} &
        \includegraphics[width=\ww,frame]{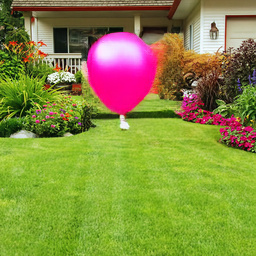}
        \\

        &
        Input image &
        Input mask &
        PaintByWord++ &
        Local CLIP-guided &
        Blended Diffusion &
        GLIDE-filtered &
        Ours
        \\

        &&&&
        diffusion
        \\
    \end{tabular}
    
    \caption{\textbf{Precision \& Diversity Experiment:} An example of a random image and mask, and the generated results, used in our quantitative evaluation.}
    \label{fig:experiment_results_example}
\end{figure*}

%% file: tables/statistical_analysis.tex
\begin{table}
    \centering
    \caption{\textbf{User study statistical analysis.} A binomial statistical test of the user study results suggests that our results are statistically significant (p-value < 5\%).}
    \begin{adjustbox}{width=1\columnwidth}
        \begin{tabular}{>{\columncolor[gray]{0.95}}lcc}
            \toprule
            
            \textbf{Method} & 
            Visualization Quality &
            Text Matching
            \\

            &
            p-value &
            p-value
            \\
            
            \midrule

            Blended Diffusion& 
            < 0.001 &
            0.043
            \\
            
            Local CLIP-guided diffusion & 
            < 0.001 &
            < 0.001
            \\
            
            PaintByWord++ & 
            < 0.001 &
            < 0.001
            \\

            GLIDE-filtered & 
            < 0.001 &
            < 0.001
            \\
            
            \bottomrule
        \end{tabular}
    \end{adjustbox}
    \label{tab:statistical_analysis}
\end{table}

%% file: figures/ranking_effectiveness/fig.tex
\begin{figure*}[h]
    \centering
    \setlength{\tabcolsep}{0.5pt}
    \renewcommand{\arraystretch}{0.5}
    \setlength{\ww}{0.33\columnwidth}
  
    \begin{tabular}{ccccccc}
        \rotatebox{90}{\phantom{a}}
        \rotatebox{90}{\phantom{AAA}``cat painting''} &
        \includegraphics[width=\ww,frame]{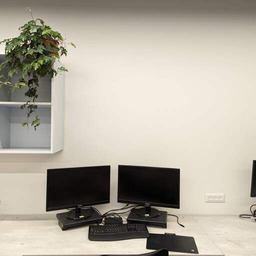} &
        \includegraphics[width=\ww,frame]{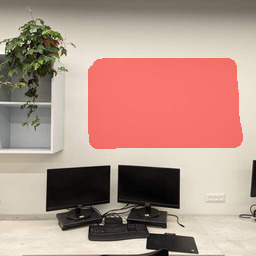} &
        \includegraphics[width=\ww,frame]{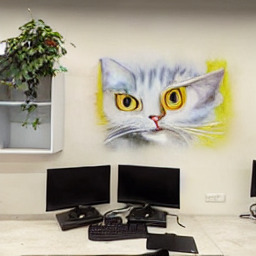} &
        \includegraphics[width=\ww,frame]{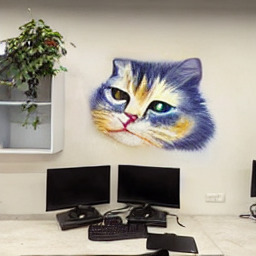} &
        \includegraphics[width=\ww,frame]{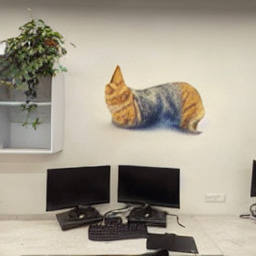} &
        \includegraphics[width=\ww,frame]{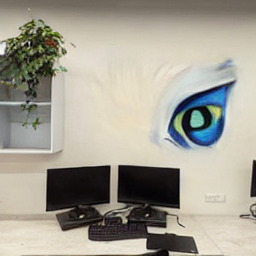}
        \\

        \rotatebox{90}{\phantom{a}}
        \rotatebox{90}{\phantom{AAAA}``Buddha''} &
        \includegraphics[width=\ww,frame]{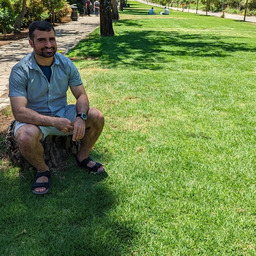} &
        \includegraphics[width=\ww,frame]{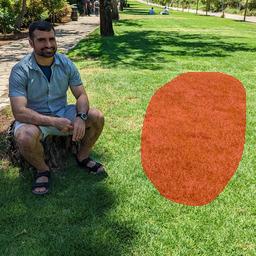} &
        \includegraphics[width=\ww,frame]{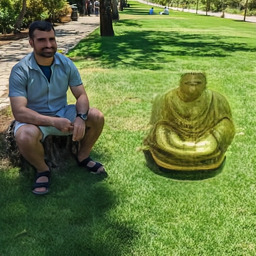} &
        \includegraphics[width=\ww,frame]{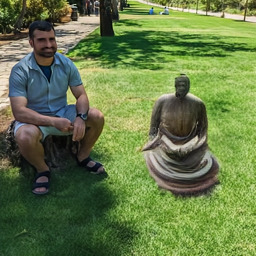} &
        \includegraphics[width=\ww,frame]{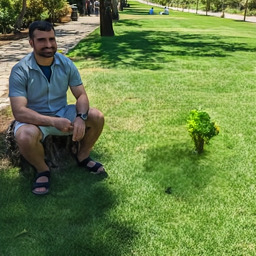} &
        \includegraphics[width=\ww,frame]{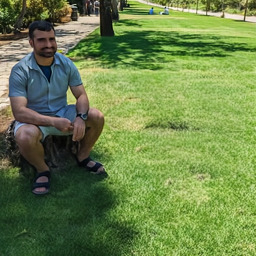}
        \\

        \rotatebox{90}{\phantom{AAa}a horror book}
        \rotatebox{90}{\phantom{AAa}named 'CVPR'} &
        \includegraphics[width=\ww,frame]{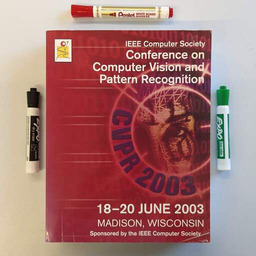} &
        \includegraphics[width=\ww,frame]{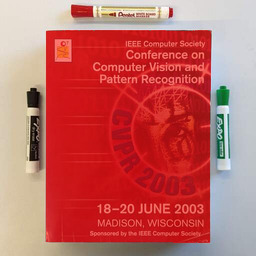} &
        \includegraphics[width=\ww,frame]{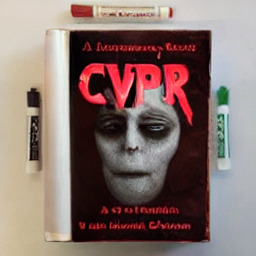} &
        \includegraphics[width=\ww,frame]{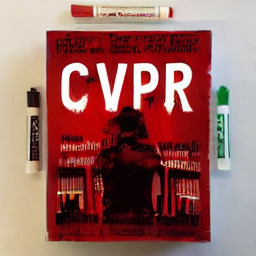} &
        \includegraphics[width=\ww,frame]{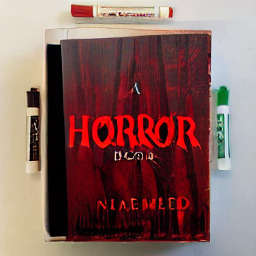} &
        \includegraphics[width=\ww,frame]{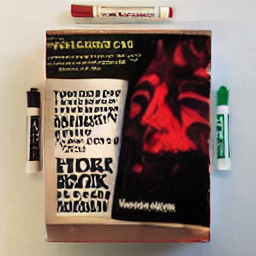}
        \\

        &
        Input image &
        Input mask &
        1st ranked result &
        2nd ranked result &
        23th ranked result &
        24th ranked result
        \\
    \end{tabular}
    
    \caption{\textbf{Ranking effectiveness:} We generate 24 prediction results and rank them using the CLIP \cite{radford2021learning} model. 
    The top 20\% of results are constantly better than the bottom 20\%, but not at the granularity of a single image --- the first image is not always strictly better than the second.}
    \label{fig:ranking_effectiveness}
\end{figure*}

%% file: figures/baselines_ranking/blended_diffusion/fig.tex
\begin{figure*}[h]
    \centering
    \setlength{\tabcolsep}{0.5pt}
    \renewcommand{\arraystretch}{0.5}
    \setlength{\ww}{0.5\columnwidth}
  
    \begin{tabular}{cccccc}
        \rotatebox{90}{\phantom{AAAAA} ``corgi painting''} &
        \includegraphics[width=\ww,frame]{figures/comparisons/baselines_extended/assets/corgi/img.jpg} &
        \includegraphics[width=\ww,frame]{figures/comparisons/baselines_extended/assets/corgi/mask_overlay.jpg} &
        \includegraphics[width=\ww,frame]{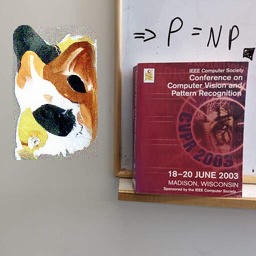} &
        \includegraphics[width=\ww,frame]{figures/comparisons/baselines_extended/assets/corgi/blended_diffusion.jpg}
        \\

        \rotatebox{90}{\phantom{a} ``a man with a yellow sweater''} &
        \includegraphics[width=\ww,frame]{figures/comparisons/baselines_extended/assets/yellow_sweater/img.jpg} &
        \includegraphics[width=\ww,frame]{figures/comparisons/baselines_extended/assets/yellow_sweater/mask_overlay.jpg} &
        \includegraphics[width=\ww,frame]{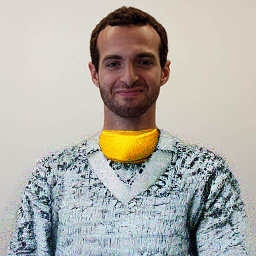} &
        \includegraphics[width=\ww,frame]{figures/comparisons/baselines_extended/assets/yellow_sweater/blended_diffusion.jpg}
        \\

        \rotatebox{90}{\phantom{AAAAAA} ``cucumber''} &
        \includegraphics[width=\ww,frame]{figures/comparisons/baselines_extended/assets/cucumber/img.jpg} &
        \includegraphics[width=\ww,frame]{figures/comparisons/baselines_extended/assets/cucumber/mask_overlay.jpg} &
        \includegraphics[width=\ww,frame]{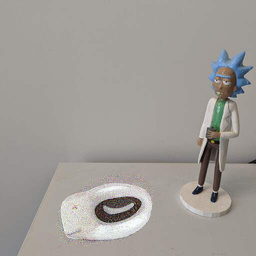} &
        \includegraphics[width=\ww,frame]{figures/comparisons/baselines_extended/assets/cucumber/blended_diffusion.jpg}
        \\

        \rotatebox{90}{\phantom{AAAAA} ``white clouds''} &
        \includegraphics[width=\ww,frame]{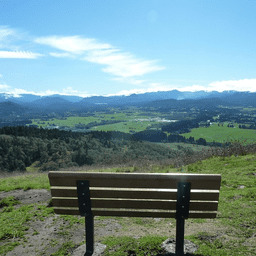} &
        \includegraphics[width=\ww,frame]{figures/comparisons/baselines_extended/assets/white_clouds/mask_overlay.jpg} &
        \includegraphics[width=\ww,frame]{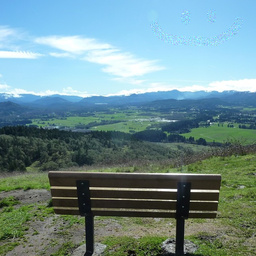} &
        \includegraphics[width=\ww,frame]{figures/comparisons/baselines_extended/assets/white_clouds/blended_diffusion.jpg}
        \\

        &
        Input image &
        Input mask &
        Without ranking &
        With ranking
        \\
    \end{tabular}
    
    \caption{\textbf{Ranking effectiveness in Blended Diffusion:} The CLIP ranking is a crucial part of Blended Diffusion \cite{avrahami2022blended}. When generating a single prediction result, the output rarely corresponds to the input text prompt.}
    \label{fig:raking_effectiveness_blended_diffusion}
\end{figure*}

%% file: appendices/sensitivity_analysis.tex
\section{Sensitivity Analysis}
We found that small input changes to our method may result in small output changes. In \Cref{fig:prompt_sensitivity} we demonstrate how small changes to the input prompt may result in small changes to the output. Furthermore, in \Cref{fig:mask_sensitivity} we demonstrate that small changes to the input mask (making it larger/smaller) may also change the output result. Lastly, in \Cref{fig:image_sensitivity} we performed small input changes: rotating the image by $5 \degree$ and performing a blurring by a Gaussian kernel with $\sigma=2$ standard deviation and kernel size $k=8$. As we can see, the outputs may change due to these input changes.

\input{figures/sensitivity_analysis/prompt_sensitivity/fig.tex}
\input{figures/sensitivity_analysis/mask_sensitivity/fig.tex}
\input{figures/sensitivity_analysis/image_sensitivity/fig.tex}

%% file: figures/sensitivity_analysis/prompt_sensitivity/fig.tex
\begin{figure*}[ht]
    \centering
    \setlength{\tabcolsep}{0.5pt}
    \renewcommand{\arraystretch}{0.5}
    \setlength{\ww}{0.33\columnwidth}
  
    \begin{tabular}{cccccc}
        \includegraphics[width=\ww,frame]{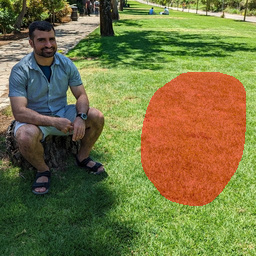} &
        \includegraphics[width=\ww,frame]{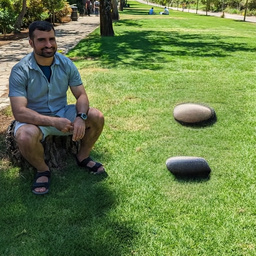} &
        \includegraphics[width=\ww,frame]{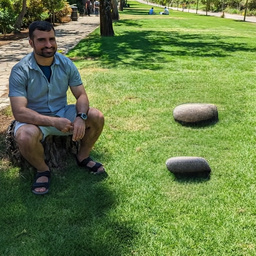} &
        \includegraphics[width=\ww,frame]{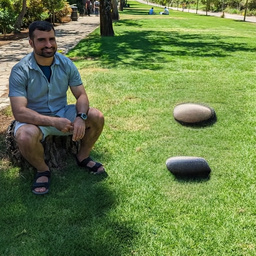} &
        \includegraphics[width=\ww,frame]{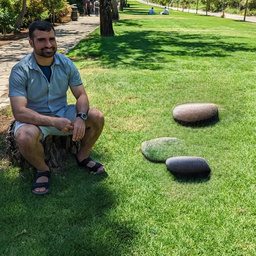} &
        \includegraphics[width=\ww,frame]{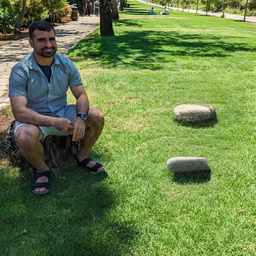}
        \\

        input + mask &
        ``stones'' &
        ``some stones'' &
        ``stones!'' &
        ``many stones'' &
        ``stone''
        \\
        \\

        \includegraphics[width=\ww,frame]{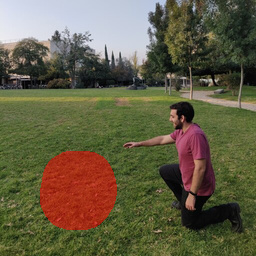} &
        \includegraphics[width=\ww,frame]{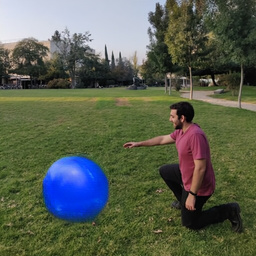} &
        \includegraphics[width=\ww,frame]{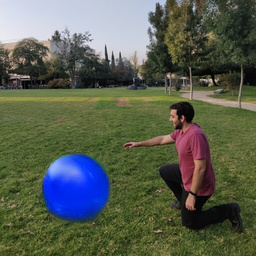} &
        \includegraphics[width=\ww,frame]{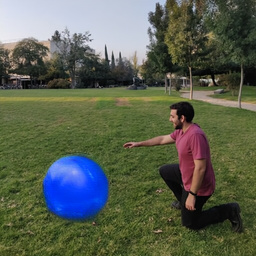} &
        \includegraphics[width=\ww,frame]{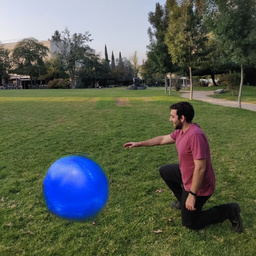} &
        \includegraphics[width=\ww,frame]{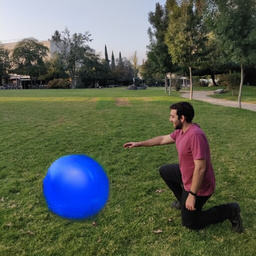}
        \\

        input + mask &
        ``a blue ball'' &
        ``a blue ball!'' &
        ``a blue ball?'' &
        ``a very blue ball'' &
        ``blue ball''
        \\
        \\

        \includegraphics[width=\ww,frame]{figures/sensitivity_analysis/prompt_sensitivity/assets/horses_beach/img_overlay.jpg} &
        \includegraphics[width=\ww,frame]{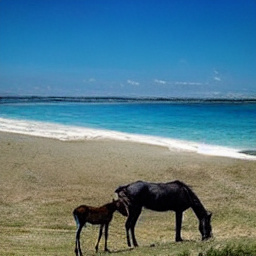} &
        \includegraphics[width=\ww,frame]{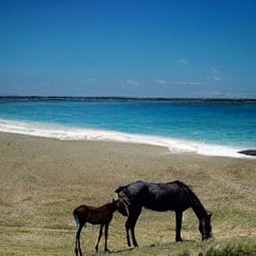} &
        \includegraphics[width=\ww,frame]{figures/sensitivity_analysis/prompt_sensitivity/assets/horses_beach/2_a_beach_ex.jpg} &
        \includegraphics[width=\ww,frame]{figures/sensitivity_analysis/prompt_sensitivity/assets/horses_beach/3_beaches.jpg} &
        \includegraphics[width=\ww,frame]{figures/sensitivity_analysis/prompt_sensitivity/assets/horses_beach/4_a_day_at_the_beach.jpg}
        \\

        input + mask &
        ``beach'' &
        ``a beach'' &
        ``a beach!'' &
        ``beaches'' &
        ``a day at the beach''

    \end{tabular}
    
    \caption{\textbf{Prompt sensitivity:} Our method is somewhat sensitive to the input prompt --- the results may change slightly for small input prompt changes.}
    \label{fig:prompt_sensitivity}
\end{figure*}

%% file: figures/sensitivity_analysis/mask_sensitivity/fig.tex
\begin{figure*}[ht]
    \centering
    \setlength{\tabcolsep}{0.5pt}
    \renewcommand{\arraystretch}{0.5}
    \setlength{\ww}{0.325\columnwidth}
  
    \begin{tabular}{ccccccc}
        \rotatebox{90}{\phantom{AAAA}``stones''} &
        \includegraphics[width=\ww,frame]{figures/sensitivity_analysis/mask_sensitivity/assets/qusay_stones/0_overlay.jpg} &
        \includegraphics[width=\ww,frame]{figures/sensitivity_analysis/mask_sensitivity/assets/qusay_stones/0_result.jpg} 
        \phantom{A}
        &
        \includegraphics[width=\ww,frame]{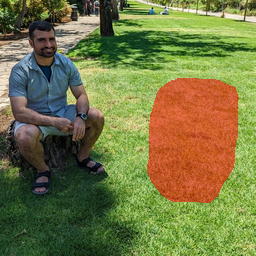} &
        \includegraphics[width=\ww,frame]{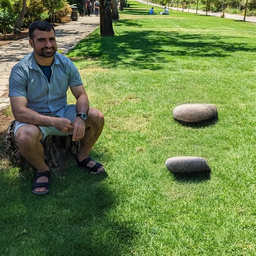} 
        \phantom{A}
        &
        \includegraphics[width=\ww,frame]{figures/sensitivity_analysis/mask_sensitivity/assets/qusay_stones/2_overlay.jpg} &
        \includegraphics[width=\ww,frame]{figures/sensitivity_analysis/mask_sensitivity/assets/qusay_stones/2_result.jpg}
        \\
        \\

        \rotatebox{90}{\phantom{AAA}``a blue ball''} &
        \includegraphics[width=\ww,frame]{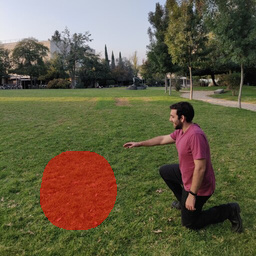} &
        \includegraphics[width=\ww,frame]{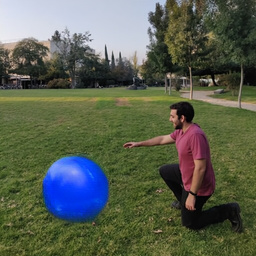} 
        \phantom{A}
        &
        \includegraphics[width=\ww,frame]{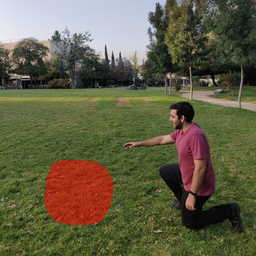} &
        \includegraphics[width=\ww,frame]{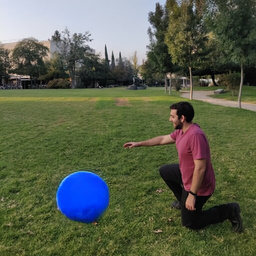} 
        \phantom{A}
        &
        \includegraphics[width=\ww,frame]{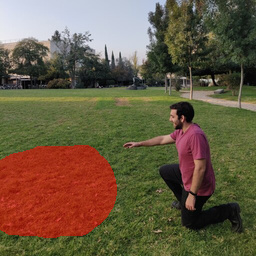} &
        \includegraphics[width=\ww,frame]{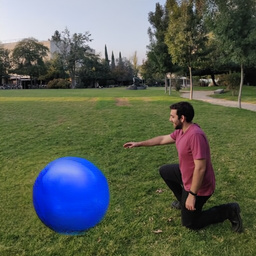}
        \\
        \\

        \rotatebox{90}{\phantom{AAAA}``a beach''} &
        \includegraphics[width=\ww,frame]{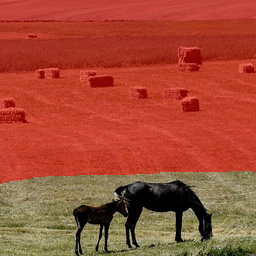} &
        \includegraphics[width=\ww,frame]{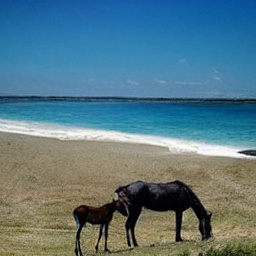} 
        \phantom{A}
        &
        \includegraphics[width=\ww,frame]{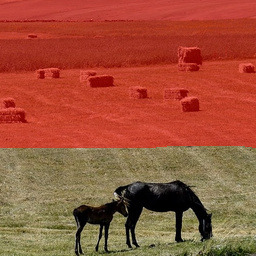} &
        \includegraphics[width=\ww,frame]{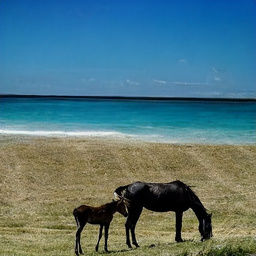} 
        \phantom{A}
        &
        \includegraphics[width=\ww,frame]{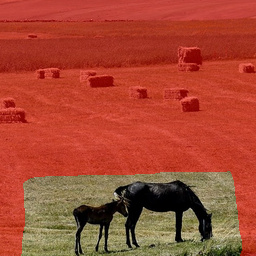} &
        \includegraphics[width=\ww,frame]{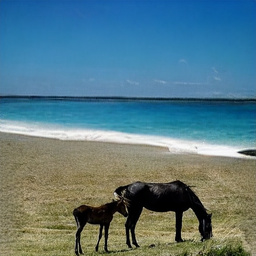}
        \\

        &
        input + mask &
        result &
        input + mask &
        result &
        input + mask &
        result
    \end{tabular}
    
    \caption{\textbf{Mask sensitivity:} Our method is somewhat sensitive to the input mask --- the results may change for small input mask changes.}
    \label{fig:mask_sensitivity}
\end{figure*}

%% file: figures/sensitivity_analysis/image_sensitivity/fig.tex
\begin{figure*}[ht]
    \centering
    \setlength{\tabcolsep}{0.5pt}
    \renewcommand{\arraystretch}{0.5}
    \setlength{\ww}{0.325\columnwidth}
  
    \begin{tabular}{ccccccc}
        \rotatebox{90}{\phantom{AAAA}``stones''} &
        \includegraphics[width=\ww,frame]{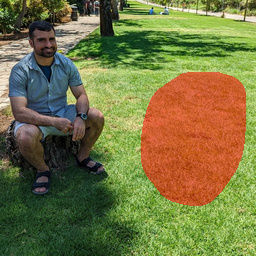} &
        \includegraphics[width=\ww,frame]{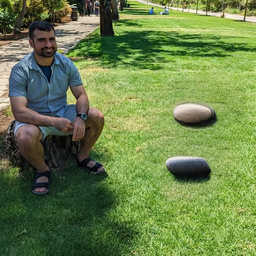} 
        \phantom{A}
        &
        \includegraphics[width=\ww,frame]{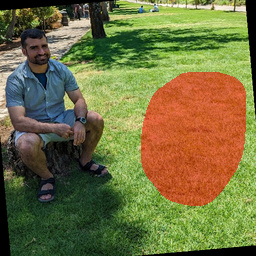} &
        \includegraphics[width=\ww,frame]{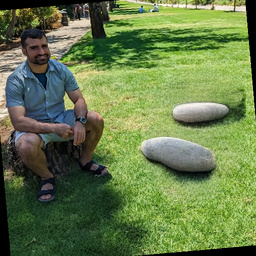} 
        \phantom{A}
        &
        \includegraphics[width=\ww,frame]{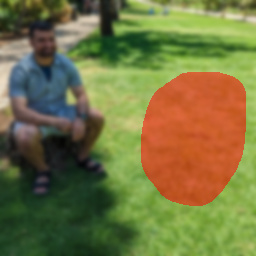} &
        \includegraphics[width=\ww,frame]{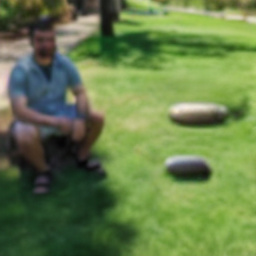} 
        \\
        \\

        \rotatebox{90}{\phantom{AAA}``a blue ball''} &
        \includegraphics[width=\ww,frame]{figures/sensitivity_analysis/image_sensitivity/assets/yossi_ball/0_result.jpg} &
        \includegraphics[width=\ww,frame]{figures/sensitivity_analysis/image_sensitivity/assets/yossi_ball/0_overlay.jpg} 
        \phantom{A}
        &
        \includegraphics[width=\ww,frame]{figures/sensitivity_analysis/image_sensitivity/assets/yossi_ball/1_overlay.jpg} &
        \includegraphics[width=\ww,frame]{figures/sensitivity_analysis/image_sensitivity/assets/yossi_ball/1_result.jpg} 
        \phantom{A}
        &
        \includegraphics[width=\ww,frame]{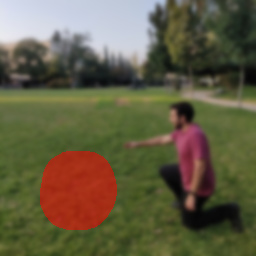} &
        \includegraphics[width=\ww,frame]{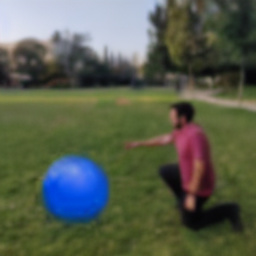}
        \\
        \\

        \rotatebox{90}{\phantom{AAAA}``a beach''} &
        \includegraphics[width=\ww,frame]{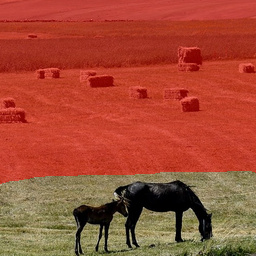} &
        \includegraphics[width=\ww,frame]{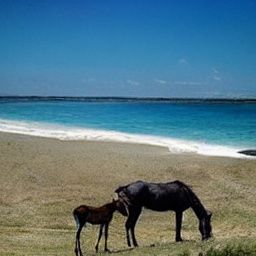} 
        \phantom{A}
        &
        \includegraphics[width=\ww,frame]{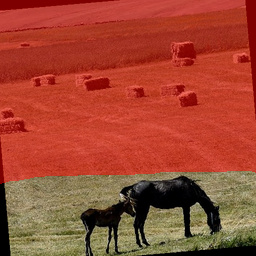} &
        \includegraphics[width=\ww,frame]{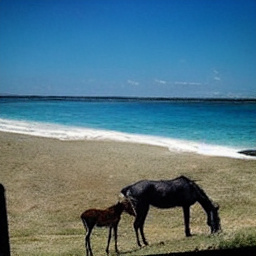} 
        \phantom{A}
        &
        \includegraphics[width=\ww,frame]{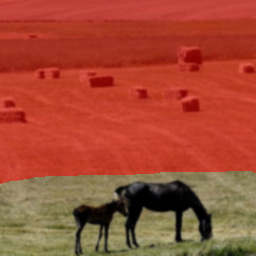} &
        \includegraphics[width=\ww,frame]{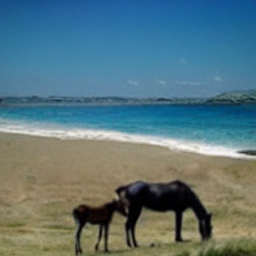}
        \\

        &
        input + mask &
        result &
        Rotated $5 \degree$ &
        result &
        Gaussian blur &
        result
    \end{tabular}
    
    \caption{\textbf{Image sensitivity:} Our method is somewhat sensitive to the input images --- the results may change for small input image changes such as rotation and blur.}
    \label{fig:image_sensitivity}
\end{figure*}

%% file: sections/societal_impact.tex
\section{Societal Impact}
Lowering the barrier for content manipulations is a mixed blessing: on the one hand, it democratizes content creation, enhances creativity, and enables new applications. On the other hand, it can be used in a nefarious manner for generating fake news, harassing, bullying, and causing bad psychological and sociological effects \cite{10.1145/3326601}. In addition, the LDM model was trained on LAION-400M dataset \cite{schuhmann2021laion} that consists of 400M text-image pairs that were collected from the internet. This dataset is non-curated, and as such may contain discomforting and disturbing content that may be repeated by the model. Moreover, it was shown \cite{nichol2021glide} that text-to-image generative models may inherit some of the biases in the training data, hence editing images guided by a text prompt may also suffer from this problem.

We strongly believe that despite these drawbacks, producing better content creation methods will produce a net positive to society.